\documentclass{article}

 \usepackage[preprint]{neurips_2026}

\usepackage[utf8]{inputenc}
\usepackage[T1]{fontenc}
\usepackage{hyperref}
\usepackage{url}
\usepackage{booktabs}
\usepackage{amsfonts}
\usepackage{nicefrac}
\usepackage{microtype}
\usepackage[table]{xcolor}
\usepackage{xspace}
\usepackage{multirow}
\usepackage{adjustbox}
\usepackage{makecell}
\definecolor{mctaHeader}{HTML}{173B63}
\definecolor{mctaHeaderLite}{HTML}{DCE8F5}
\definecolor{mctaStripe}{HTML}{F7FAFD}
\definecolor{mctaBest}{HTML}{CDECCF}
\definecolor{mctaWarn}{HTML}{FCE8C3}
\definecolor{mctaBad}{HTML}{F8D7DA}
\definecolor{mctaGreen}{HTML}{1B7837}
\definecolor{mctaRed}{HTML}{B2182B}
\definecolor{mctaGray}{HTML}{6B7280}
\newcommand{\tightarray}{\setlength{\tabcolsep}{4pt}\renewcommand{\arraystretch}{1.16}}
\newcommand{\theadc}[1]{\multicolumn{1}{c}{\textcolor{white}{\textbf{#1}}}}

\usepackage{booktabs}
\usepackage[table]{xcolor}
\usepackage{tabularx}
\usepackage{longtable}
\usepackage{array}
\usepackage{multirow}
\usepackage{threeparttable}
\usepackage{adjustbox}
\usepackage{makecell}
\usepackage{pdflscape}
\usepackage{siunitx}
\usepackage{caption}
\usepackage{amssymb}
\usepackage{rotating}

\usepackage{amssymb}
\usepackage{array}
\usepackage{graphicx}
\usepackage[normalem]{ulem}
\useunder{\uline}{\ul}{}
\usepackage{multicol}
\usepackage{tabularx}
\usepackage{enumitem}
\usepackage{pifont}
\usepackage{colortbl}
\usepackage{subcaption}
\usepackage{tikz}
\usepackage{geometry}
\usepackage{subcaption}
\usepackage{wrapfig}
\usepackage[most]{tcolorbox}
\usepackage{listings}
\usepackage{setspace}
\definecolor{darkgreen}{RGB}{0,120,0}   % darker green
\definecolor{darkgreenbg}{RGB}{200,235,200} % darker background shade
\newcommand{\cmark}{\textcolor{green!70!black}{\ding{51}}}
\newcommand{\xmark}{\textcolor{red}{\ding{55}}}
\usepackage[table]{xcolor}
\usepackage{tcolorbox}
\usepackage{caption}
\usepackage{enumitem}
\usepackage{xcolor}
\usepackage{graphicx}
\usepackage{tabularx}
\usepackage{amsmath}
\tcbuselibrary{breakable, skins}

\newtcolorbox{stagebox}[2][]{
    enhanced,
    breakable,
    colback=white,
    colframe=black!70,
    boxrule=0.8pt,
    arc=4pt,
    left=6pt,
    right=6pt,
    top=6pt,
    bottom=6pt,
    title={#2},
    fonttitle=\bfseries,
    coltitle=black,
    #1
}
\definecolor{headercolor}{RGB}{230,230,230}
\usepackage{hyperref}

\hypersetup{
    colorlinks=true,
    citecolor=green,   % citations
    linkcolor=blue,    % refs (tables, figs, sections)
    urlcolor=red       % URLs
}
\definecolor{openai}{RGB}{235,245,255}
\definecolor{anthropic}{RGB}{255,242,230}
\definecolor{xai}{RGB}{255,240,240}
\definecolor{google}{RGB}{236,248,236}
\definecolor{qwen}{RGB}{245,238,255}
\definecolor{deepseek}{RGB}{255,238,242}
\definecolor{meta}{RGB}{238,245,252}
\definecolor{mistral}{RGB}{252,247,232}
\definecolor{yi}{RGB}{238,252,252}
\definecolor{medical}{RGB}{245,245,245}
\definecolor{others}{RGB}{230,250,235}
\title{MedCTA: A Benchmark for Clinical Tool Agents}

\author{
\textbf{Tajamul Ashraf}$^{1}$,
\textbf{Hyewon Jeong}$^{2}$,
\textbf{Fida Mohammad Thoker}\thanks{Correspondence: \texttt{fida.thoker@kaust.edu.sa}} $^{1}$,
\textbf{Bernard Ghanem}$^{1}$\\[4pt]
$^{1}$King Abdullah University of Science and Technology (KAUST), Saudi Arabia\\
$^{2}$Massachusetts Institute of Technology (MIT), USA\\[4pt]
} 
\begin{document}

\def\llms{\texttt{LLMs}\xspace}
\def\lmms{\texttt{LMMs}\xspace}
\def\vlms{\texttt{VLMs}\xspace}
\def\ours{\texttt{MedCTA}\xspace}
\def\lmms{\texttt{LMMs}\xspace}
\def\lmm{\texttt{LMM}\xspace}
\def\llm{\texttt{LLM}\xspace}
\def\api{\texttt{APIs}\xspace}
\def\glue{\texttt{GLUE}\xspace}
\def\langchain{\texttt{LangChain}\xspace}
\def\autogpt{\texttt{AutoGPT}\xspace}
\def\babyagi{\texttt{BabyAGI}\xspace}
\def\webgpt{\texttt{WebGPT}\xspace}
\def\webshop{\texttt{WebShop}\xspace}
\def\webcpm{\texttt{WebCPM}\xspace}
\def\restgpt{\texttt{RestGPT}\xspace}
\def\appagent{\texttt{AppAgent}\xspace}
\def\visualgpt{\texttt{Visual ChatGPT}\xspace}
\def\mmreact{\texttt{MM-ReAct}\xspace}
\def\mllm{\texttt{MLLMTool}\xspace}
\def\llava{\texttt{LLaVA-Plus}\xspace}
\def\datacopilot{\texttt{DataCopilot}\xspace}
\def\hugginggpt{\texttt{HuggingGPT}\xspace}
\def\modelscope{\texttt{ModelScopeAgent}\xspace}
\def\avatar{\texttt{Avatar}\xspace}
\def\rest{\texttt{REST}\xspace}
\def\llamav{\texttt{LlamaV-01}\xspace}
\def\clevr{\texttt{CLEVR}\xspace}
\def\strategyqa{\texttt{StrategyQA}\xspace}
\def\scienceqa{\texttt{ScienceQA}\xspace}
\def\mathvista{\texttt{MathVista}\xspace}
\def\sharegpt{\texttt{ShareGPT-4o}\xspace}
\def\toolbench{\texttt{ToolBench}\xspace}
\def\apibench{\texttt{APIBench}\xspace}
\def\apibank{\texttt{API-Bank}\xspace}
\def\mnms{\texttt{m\&m’s}\xspace}
\def\gaia{\texttt{GAIA}\xspace}
\def\gta{\texttt{GTA}\xspace}
\def\osworld{\texttt{Osworld}\xspace}
\def\toolqa{\texttt{ToolQA}\xspace}
\def\gentopia{\texttt{Gentopia}\xspace}
\def\gorilla{\texttt{Gorilla}\xspace}
\def\agentbench{\texttt{AgentBench}\xspace}
\def\mlgym{\texttt{MLGym}\xspace}
\def\benchmark{{Agent-X}\xspace}
\def\llamav{\texttt{LLaMA-V}\xspace}
\def\llava{\texttt{LLaVA}\xspace}
\def\llavaplus{\texttt{LLaVA-plus}\xspace}
\def\mlllmtool{\texttt{MLLM-Tool}\xspace}
\def\deder{\texttt{DEDER}\xspace}
\def\lumos{\texttt{Lumos}\xspace}
\def\taskbench{\texttt{TASKBENCH}\xspace}
\def\visualagentbench{\texttt{VisualAgentBench}\xspace}
\def\apibank{\texttt{APIBank}\xspace}
\def\toolalpaca{\texttt{ToolAlpaca}\xspace}
\def\toolbench{\texttt{ToolBench}\xspace}
\def\anytool{\texttt{AnyTool}\xspace}
\def\agentohana{\texttt{agentohana}\xspace}
\def\apigen{\texttt{APIGen}\xspace}
\def\agentinstruct{\texttt{AgentInstruct}\xspace}
\def\osworld{\texttt{OSWorld}\xspace}
\def\mmina{\texttt{MMInA}\xspace}
\def\gaia{\texttt{GAIA}\xspace}
\def\gta{\texttt{GTA}\xspace}
\def\mm{\texttt{m\&m'}\xspace}
\def\agentbench{\texttt{AgentBench}\xspace}
\def\agentgym{\texttt{AgentGym}\xspace}

\newcommand{\fm}[1]{\textcolor{blue}{FM: #1}}

\definecolor{headerBlue}{HTML}{1F3A5F}
\definecolor{stripeA}{HTML}{F4F7FB}
\definecolor{stripeB}{HTML}{FFFFFF}
\definecolor{gainGreen}{HTML}{1B9E77}
\definecolor{lossRed}{HTML}{D62728}
\definecolor{highlightRow}{HTML}{E8F2FF}
\definecolor{headerPurple}{HTML}{4A3F75}
\definecolor{stripeLight}{HTML}{F5F3FA}
\definecolor{gainBlue}{HTML}{1F77B4}
\definecolor{gainTeal}{HTML}{1B9E77}
\definecolor{midOrange}{HTML}{E67E22}
\definecolor{alertRed}{HTML}{C0392B}
\definecolor{headerblue}{RGB}{220,235,250}   % light blue header
\definecolor{rowblue}{RGB}{245,250,255}      % very light blue rows
\definecolor{rowgreen}{RGB}{240,250,240}     % light green highlight
\definecolor{rowred}{RGB}{255,240,240}       % light red highlight
\definecolor{goldbg}{RGB}{255,248,220}
\definecolor{gptbg}{RGB}{245,250,255}
\definecolor{qwenbg}{RGB}{250,245,255}
\definecolor{failred}{RGB}{210,35,35}
\definecolor{successgreen}{RGB}{34,130,70}
\definecolor{toolblue}{RGB}{35,120,200}
\definecolor{softgray}{RGB}{245,247,250}

\maketitle

\begin{abstract}
To make clinically grounded decisions, medical AI agents are expected to go beyond simple recognition and be capable of tool retrieval, evidence acquisition, and integration. Existing benchmarks largely evaluate isolated perception or single-turn question answering, and therefore provide limited visibility into failures of planning, tool recruitment, and rollout reliability. We introduce \ours, a benchmark for evaluating \emph{medical tool agents} on clinician-validated, step-implicit tasks grounded in realistic multimodal clinical inputs, including radiology images, pathology slides, and reports. \ours comprises \textbf{107} real-world clinical tasks with clinician-verified executable trajectories over \textbf{5} deployed tools, and supports process-aware evaluation of tool selection, argument validity, execution stability, trajectory fidelity, and outcome quality. We benchmark \textbf{18} open- and closed-source multimodal models and find that even frontier systems remain brittle in multi-step clinical tool use: autonomous rollouts are dominated by protocol failures, premature stopping, and incorrect tool recruitment, while gold-standard tool routing yields large but still incomplete gains. These results show that strong backbone perception does not translate into reliable agentic behavior in clinical settings. \ours provides a rigorous testbed for auditing, diagnosing, and advancing trustworthy medical AI agents. The dataset and evaluation suite are available at \href{https://ivul-kaust.github.io/MedCTA/}{https://ivul-kaust.github.io/MedCTA/}.
\end{abstract}

% \vspace{-0.30cm}
\section{Why Clinical Tool Agents?}
% \vspace{-0.30cm}

Multimodal agentic systems are pushing \llms beyond single-turn prediction toward \emph{interactive problem solving}, where a model plans, invokes tools, and updates its beliefs over multiple steps \cite{xie2024large, liu2024llava, GTA, agentlego, 2023opencompass}. This paradigm has enabled strong assistants for web interaction and document understanding \cite{nakano2021webgpt, sun2025docagent}, visual analysis \cite{suris2023vipergpt, lu2023chameleon}, and software engineering \cite{jimenez2023swebench, wu2023autogen, yang2024sweagent}. Yet recent agentic benchmarks show that strong final answers can coexist with brittle multi-step behavior: models often fail in tool routing, trajectory maintenance, or reasoning over intermediate evidence \cite{ashraf2025agent, GTA, gaia}. As models become more capable, the limiting factor is increasingly not raw perception, but reliable execution under tool interaction.

These challenges are especially acute in medicine. Clinical decision making is inherently iterative and multimodal: it requires interpreting images, extracting text, measuring findings, aggregating evidence across steps, consulting clinical knowledge, and producing structured conclusions \cite{shen2017deep, chow2016review, cabral2024clinical, sun2022lesion}. Accordingly, agentic methods are beginning to appear in healthcare \cite{duncan2000medical, xu2025comprehensive, kim2024mdagents, kim2025tiered, kim2025behaviorsft, schmidgall2024agentclinic, jiang2025medagentbench, li2024mmedagent}. But existing evaluations~
\cite{hu2024omnimedvqa, jiang2025medagentbench, he2020pathvqa} still emphasize dialogue outcomes, EHR navigation, or perception accuracy, with limited support for \emph{multimodal, tool-executable, and traceable} clinical workflows (Table~\ref{tab:medical_comparison}). This gap matters in medical settings as a final answer is not enough: users also need process-level accountability about what evidence was used, what was measured, and how the conclusion was reached.

To address this gap, we introduce \textbf{\ours}, a trajectory-aware benchmark and executable evaluation platform for \emph{medical tool agents}, inspired by agentic benchmarks like GAIA~\cite{gaia}, GTA~\cite{GTA}, and Agent-X \cite{ashraf2025agent}. \ours contains clinician-validated, step-implicit clinical queries grounded in authentic multimodal medical inputs, including radiology images, pathology slides, figures, reports, and scanned clinical content. Each task must be solved by autonomously selecting from deployed tools for perception, operation, reasoning, and reporting, without being told the tool sequence. We pair every task with a clinician-verified reference trajectory, enabling fine-grained evaluation of tool selection, argument validity, intermediate evidence consistency, and final task success.

%Across \textbf{18} models and \textbf{107} clinician-verified tasks, \ours reveals a substantial gap between apparent medical competence and reliable medical agency. The best official outcome score is only \textbf{31.54}, the strongest open model reaches \textbf{27.80}, and under stricter rollout diagnostics no model achieves non-zero strict trajectory success. For frontier models, oracle tool routing raises diagnostic answer accuracy by \textbf{20\%}, showing that much of the current loss comes from controller failures rather than absent medical knowledge. The remaining errors are dominated by protocol instability, premature stopping, wrong tool recruitment, and failures in localized evidence grounding. These findings position \ours not just as a leaderboard benchmark, but as a diagnostic instrument for auditing and improving clinically grounded agents. Our contributions are:
Across \textbf{18} models and \textbf{107} tasks, \ours reveals a gap between medical competence and reliable medical agency. The best outcome accuracy in the agentic (tool-using) setting is only \textbf{31.54\%}, while the strongest open model reaches \textbf{27.80\%}. Under stricter rollout diagnostics, no model achieves non-zero strict trajectory success. In contrast, when provided with gold-standard next-tool routing, while still generating arguments and final answers, diagnostic answer accuracy improves by up to \textbf{+35\%}. This gap indicates that most performance degradation arises from controller failures rather than a lack of underlying clinical reasoning. The remaining errors are dominated by protocol instability, premature stopping, wrong tool recruitment, and failures in localized evidence grounding. These findings position \ours not just as a leaderboard benchmark, but as a diagnostic instrument for auditing and improving clinically grounded agents, helping to steer the field toward more reliable controller design and robust tool use. To summarize:

\begin{enumerate}[label=\textbullet, leftmargin=12pt, topsep=0pt, itemsep=1pt, partopsep=1pt, parsep=1pt]
\item We introduce \ours, a clinician-validated benchmark for \emph{medical tool agents} built around realistic multimodal clinical tasks with step-implicit queries and executable tool use.
\item We curate \textbf{107} real-world tasks with clinician-verified trajectories across \textbf{5} deployed tools, and define process-aware metrics for tool routing, trajectory fidelity, evidence use, and outcomes.
\item We benchmark \textbf{18} frontier models and show that current systems remain brittle on multi-step clinical tool use, with large controller gaps, severe under-calling, and persistent failures in long-horizon localized grounding.
\end{enumerate}
\begin{table*}[tb]
\centering
\caption{
Comparison with representative medical \texttt{VQA}, medical reasoning, and medical
agent benchmarks. \ours is the only benchmark in this comparison that combines step-implicit clinical queries with executable, clinician-verified tool trajectories across multimodal medical inputs.
}

\resizebox{\textwidth}{!}{

\rowcolors{3}{gray!10}{white}

\begin{tabular}{l|c c c c c c}
\toprule

\rowcolor{blue!15}
\textbf{Benchmark} 
& \textbf{\# Modalities}
& \textbf{Agentic} 
& \textbf{Real-world} 
& \textbf{Real-world} 
& \textbf{Multimodal} 
& \textbf{Deep} \\

\rowcolor{blue!15}
&  
& \textbf{Tasks} 
& \textbf{Assistive Tools}
& \textbf{Clinical Queries} 
& \textbf{Inputs} 
& \textbf{Reasoning} \\

\midrule

VQA-RAD~\cite{lau2018dataset} 
& 3
& \xmark 
& \xmark
& \xmark 
& \cmark 
& \xmark \\

SLAKE~\cite{liu2021slake}
& 3
& \xmark 
& \xmark
& \xmark 
& \cmark 
& \xmark \\

Path-VQA~\cite{he2020pathvqa} 
& 2
& \xmark 
& \xmark
& \xmark 
& \cmark 
& \xmark \\

VQA-Med~\cite{ben2019vqa}
& 5
& \xmark 
& \xmark
& \xmark 
& \cmark 
& \xmark \\

PMC-VQA~\cite{zhang2024pmc} 
& 2
& \xmark 
& \xmark
& \xmark 
& \cmark 
& \xmark \\

MMMU-Med~\cite{yue2023mmmu} 
& 2
& \xmark 
& \xmark
& \cmark 
& \cmark 
& \cmark \\

MedQA~\cite{jin2021disease}
& 1
& \xmark 
& \xmark
& \cmark 
& \xmark 
& \cmark \\

PubMedQA~\cite{jin2019pubmedqa}
& 1
& \xmark 
& \xmark
& \cmark 
& \xmark 
& \cmark \\

MedAgentBench~\cite{jiang2025medagentbench} 
& 1
& \cmark 
& \xmark
& \cmark 
& \xmark 
& \cmark \\

OmniMedVQA~\cite{hu2024omnimedvqa}
& 12
& \xmark 
& \xmark
& \xmark 
& \cmark 
& \xmark \\

\rowcolor{green!15}
\textbf{MedCTA (Ours)} 
& \textbf{16}
& \cmark 
& \cmark
& \cmark 
& \cmark 
& \cmark \\

\bottomrule
\end{tabular}
}

\label{tab:medical_comparison}
\end{table*}
\vspace{-0.8em}

 % \vspace{-0.10cm}
\section{Related Work}
 % \vspace{-0.10cm}

\noindent\textbf{Tool-augmented and trajectory-aware agents.}
General-purpose agent systems have shown that large language models can act as controllers over external tools and environments, as in LangChain~\cite{langchain}, Auto-GPT~\cite{autogpt}, and BabyAGI~\cite{babyagi}. Subsequent work extended this paradigm to web interaction with WebGPT~\cite{nakano2021webgpt}, WebShop~\cite{webshop}, and WebCPM~\cite{webcpm}; API and service orchestration with RESTGPT~\cite{restgpt} and APIBank~\cite{apibank}; operating-system and app control with OSWorld~\cite{osworld} and AppAgent~\cite{appagent}; and multimodal tool use with HuggingGPT~\cite{hugginggpt} and MSAgent~\cite{msagent}. Vision-centric agents such as VisualGPT~\cite{visualgpt}, MM-REACT~\cite{mmreact}, MLLM~\cite{mllm}, and LLaVA-Plus~\cite{liu2024llava} further demonstrate structured reasoning over visual inputs. These works establish the promise of tool-augmented reasoning, but they are largely open-domain and are typically evaluated by end-task success. They do not target safety-critical clinical workflows, medically grounded tool ecosystems, or trajectory-level auditing of tool choice, argument validity, and intermediate evidence consistency.

%\medskip
\noindent\textbf{Multimodal medical benchmarks and models.}
Progress in medical multimodal learning has been driven primarily by perception- and QA-centric benchmarks such as VQA-RAD~\cite{lau2018dataset}, SLAKE~\cite{liu2021slake}, Path-VQA~\cite{he2020pathvqa}, and VQA-Med~\cite{ben2019vqa}. More recent resources broaden coverage and reasoning scope, including PMC-VQA~\cite{zhang2024pmc}, MMMU-Med~\cite{yue2023mmmu}, and OmniMedVQA~\cite{hu2024omnimedvqa}. These datasets have supported increasingly strong medical vision-language models, including BioViL~\cite{biovil}, MedCLIP~\cite{medclip}, LLaVA-Med~\cite{llava-med}, Med-Flamingo~\cite{medflamingo}, RadFM~\cite{radfm}, Lingshu~\cite{xu2025lingshu}, Fleming-VL~\cite{shu2025fleming}, MedGemma~\cite{sellergren2025medgemma}, and MedMO~\cite{deria2026medmo}. However, these benchmarks and models still mostly evaluate static understanding: classification, report generation, or single-turn question answering. They say little about whether a model can decide \emph{when} to use a tool, \emph{which} tool to use, \emph{how} to structure the call, and \emph{how} to integrate intermediate evidence into a faithful multi-step clinical decision.

%\medskip
\noindent\textbf{Medical agents and interactive clinical evaluation.}
Recent work has begun moving toward more interactive medical settings. Multi-agent and coordination-oriented systems include MDAgents~\cite{kim2024mdagents} and TAO~\cite{kim2025tiered}. Simulated clinical-interaction benchmarks include AgentClinic~\cite{schmidgall2024agentclinic} and AI Hospital~\cite{fan2025aihospital}. Longitudinal record-centric environments are represented by Virtual EHR / MedAgentBench~\cite{jiang2025virtualehr}, while proactive clarification and uncertainty-aware interaction are emphasized in BehaviorSFT~\cite{kim2025behaviorsft} and MediQ~\cite{li2024mediq}. Large-scale clinical decision evaluation is further explored in CliBench~\cite{ma2024clibench}. Additional closely related efforts include modality-specific reasoning agents such as MedRAX~\cite{medrax}, contamination-free evaluation frameworks like LiveMedBench~\cite{yan2026livemedbench}, HealthAdminBench~\cite{bedi2026healthadminbench} and multimodal agentic diagnosis systems such as MedAgent-Pro~\cite{wang2025medagent}. Strong generalist and evaluation-oriented baselines include MedVersa~\cite{zhou2024medversa} and MedQA-CS~\cite{yao2024medqa}, while tool-centric and adaptive agent frameworks such as MedAgentGYM~\cite{xu2025medagentgym} further explore multi-step orchestration and self-evolving behaviors. In this work, we introduce \ours, the first agentic benchmark specifically designed to evaluate tool-augmented \llm models on realistic clinical tasks with executable tool chains and trajectory-level reasoning.
\begin{figure}[t]
    \includegraphics[width=\textwidth]{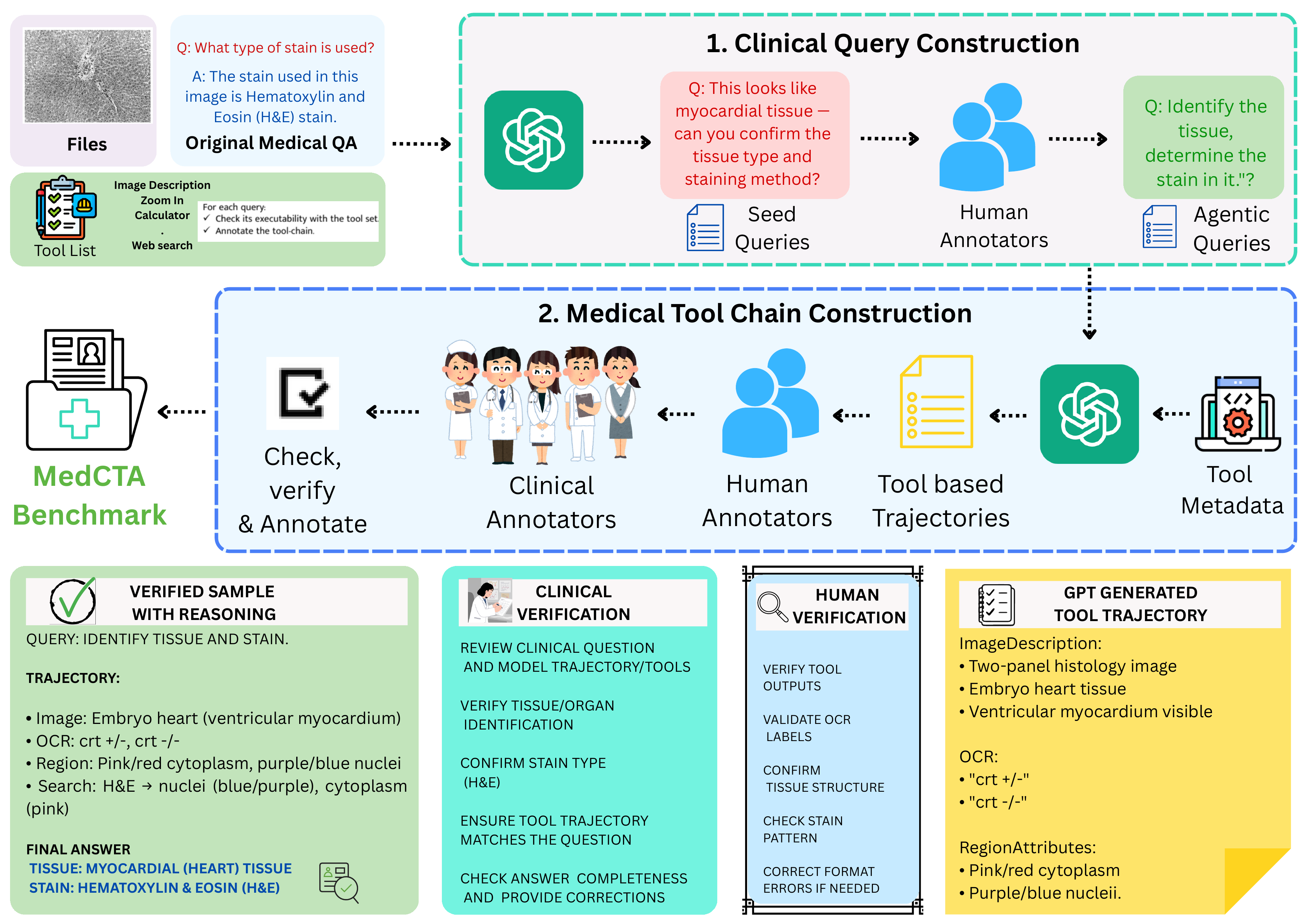}
    \caption{\textbf{MedCTA curation pipeline.}
\ding{202} \textit{Query construction:} Annotators expand expert exemplars into structured, executable, and tool-aware clinical queries, followed by clinical verification for correctness and relevance.
\ding{203} \textit{Tool chain construction:} GPT-generated tool trajectories are refined and corrected by annotators. Technical verification ensures executability and format compliance, while clinical experts validate medical correctness and reasoning soundness.}
    \label{fig:pipeline}
    % \vspace{-15pt}
\end{figure}

% \vspace{-0.25cm}
\section{MedCTA Benchmark}
% \vspace{-0.25cm}
In this section, we describe the design and content of the \ours benchmark. We introduce the formulation of a dataset sample in Section~\ref{sec:dform}. The construction procedures for clinical queries and medical tool chains are presented in Sections~\ref{sec:qcons} and~\ref{sec:tcons}, respectively. 
The overall dataset construction pipeline is shown in Figure~\ref{fig:pipeline} with corresponding dataset statistics in Section~\ref{sec:dstat}. 

% \vspace{-0.25cm}
\subsection{Dataset Formulation}
% \vspace{-0.25cm}
\label{sec:dform}

Let $\mathcal{D}=\{\tau_k\}_{k=1}^{N}$ denote a library of executable tools. We formulate each task in \ours as a structured tuple
$(\mathcal{X}, \mathcal{Q}, \mathcal{U}, \pi, \mathcal{A})$ that specifies (i) the clinical context, (ii) an objective to be solved, (iii) the tools required, (iv) the tool-mediated interaction trace, and (v) the final clinical outcome. This design follows the general agentic benchmark template of defining a context, a query that necessitates multi-step tool use, a tool subset, and a stepwise trace of tool calls. %:contentReference[oaicite:0]{index=0}

\noindent\textbf{Clinical context.} $\mathcal{X}$ is a collection of clinical images, including one or two images (e.g., CT, MRI, X-ray, ultrasound, pathological images) and/or scanned documents. The context may be heterogeneous and multimodal, reflecting realistic clinical setups where evidence is distributed across files.

\noindent\textbf{Agentic query.} $\mathcal{Q}$ is a clinically grounded request conditioned on $\mathcal{X}$. While $\mathcal{Q}$ describes a plausible medical goal, solving it requires decomposition into multiple subproblems and strategic invocation of tools from $\mathcal{D}$. Importantly, the query does not reveal which tools are needed, nor their execution order, and the agent must plan to execute the query. %The agent controller to solve the given query.
%:contentReference[oaicite:1]{index=1}

\noindent\textbf{Tool subset.} $\mathcal{U}\subseteq \mathcal{D}$ denotes the (hidden) subset of tools that are sufficient to resolve the query:
$\mathcal{U}=\bigcup_{i=1}^{m}\{\tau_i\}$, where $\tau_i$ is the tool used at step $i$.

\noindent\textbf{Reference tool chain (interaction trace).} We provide a reference tool-assisted reasoning trace
$\pi=\{s_i\}_{i=1}^{m}$, where each step is a triplet $s_i=(\tau_i,\alpha_i,\rho_i)$ consisting of the selected tool, its input arguments, and the returned observation (tool output), respectively. This representation enables fine-grained evaluation of whether the agent makes coherent intermediate decisions and uses tools effectively throughout the chain. %:contentReference[oaicite:2]{index=2}

\noindent\textbf{Final outcome.} $\mathcal{A}$ is the final clinical conclusion produced after completing the tool-mediated process (e.g., diagnosis, quantified assessment, classification, or a finding). For queries in which the medically correct response is not unique, we include multiple acceptable reference answers to reflect valid clinical variability. For queries with a single ground-truth target, $\mathcal{A}$ is unique.

% \noindent\textbf{Tool and query taxonomy.} In \ours, $\mathcal{D}$ contains X executable multimodal tools. We further organize queries $\mathcal{Q}$ into three families: (1) diagnostic reasoning, (2) objective clinical assessment, and (3) image synthesis (Appendix~\ref{appendix:query_type}). Diagnostic reasoning tasks admit descriptive, potentially non-unique conclusions and thus include multiple reference answers; objective assessment tasks yield a uniquely determined outcome; and for image synthesis tasks we do not directly score the generated image, hence we set $\mathcal{A}=\emptyset$.

\begin{table}[t]
\centering
\renewcommand{\arraystretch}{1}
\caption{
\textbf{Query transformation in \ours.} 
\textcolor{blue}{Blue} highlights perception-only prompts. 
\textcolor{red}{Red} highlights procedural/tool-leaking phrasing. 
\textcolor{darkgreen}{Green} highlights the final clinically grounded design.
}
\label{tab:query_comparison}
\resizebox{\textwidth}{!}{

\begin{tabular}{@{}>{\centering\arraybackslash}m{2.5cm}|
>{\centering\arraybackslash}m{4.5cm}|
>{\centering\arraybackslash}m{8cm}|
>{\centering\arraybackslash}m{8cm}@{}}

\toprule

\textbf{Domain} &
\textbf{Original Dataset Query} &
\textbf{LMM-based Agentic Query (Tool-Explicit)} &
\textbf{Final Query (Human + Clinician Verified)} \\

\midrule

Radiology (CT) 
& What type of \textcolor{blue}{thrombosis} is shown in the image? 
& Using the CT image, \textcolor{red}{localize} the portal vein and superior mesenteric vein regions and \textcolor{red}{verify} whether thrombosis is present in each; then determine the type of thrombosis. 
& \textcolor{darkgreen}{Based on the CT image}, what type of venous thrombosis is present? \\

\midrule

Radiology (Mass Measurement) 
& What is the \textcolor{blue}{size} of the mass? 
& Using the abdominal CT image, first \textcolor{red}{check for measurement labels via OCR}, then \textcolor{red}{estimate the mass dimensions} and summarize the size. 
& Are any measurement labels visible in the image? Also, find the mass size and identify its abdominal quadrant. \\

\midrule

Histology (Organ Comparison) 
& What organs are being compared in the image? 
& \textcolor{red}{Summarize the multi-panel figure}, \textcolor{red}{detect panel titles}, and verify organ labels before identifying which organs are compared. 
& Identify the organs shown. \\

\midrule

Histology (Tissue Identification) 
& What \textcolor{blue}{organ} is being analyzed in the image? 
& \textcolor{red}{Localize a representative tissue region with a bounding box} and describe its morphology to determine which organ is analyzed. 
& Identify the tissue and determine the stain used. \\

\midrule

Developmental Pathology 
& What is the main \textcolor{blue}{difference} between the normal and Gpc3 -/ mouse kidneys at E12.0? 
& Compare E12.0 kidney sections labeled Gpc3 +/+ and Gpc3 -/. \textcolor{red}{Confirm labels} and \textcolor{red}{inspect ureteric bud branching differences using tools}. 
& Compare Gpc3 +/+ and Gpc3 -/ embryonic kidneys across E12.0, E13.5, and E16.5, focusing on ureteric bud branching, kidney size, and cortical–medullary architecture. \\

\midrule

\rowcolor{green!10}
\textbf{Design Principle} 
& \textcolor{blue}{Single-step perception question.} 
& \textcolor{red}{Tool-explicit, step-decomposed, instruction-driven phrasing.} 
& \textcolor{darkgreen}{Clinically grounded, goal-oriented query without explicit tool or procedural references.} \\

\bottomrule
\end{tabular}
}
\vspace{-0.4 cm}
\end{table}

% \vspace{-0.25cm}
\subsection{Clinical Query Construction}
% \vspace{-0.25cm}
\label{sec:qcons}

A central goal of \ours is to move beyond \emph{perception-only} medical \texttt{VQA} and evaluate \emph{agentic} clinical problem solving. The model must interpret evidence in $\mathcal{X}$, infer which tools are useful, and execute a coherent multi-step workflow without being told the tool sequence. To build $(\mathcal{X},\mathcal{Q},\mathcal{U})$, we use a semi-automated, human-in-the-loop pipeline that converts shallow dataset-style questions into realistic, tool-requiring clinical objectives while explicitly removing procedural hints from the final query (Table~\ref{tab:query_comparison}). This follows the core principle used in recent agentic vision benchmarks~\cite{ashraf2025agent}: tool use should be necessary, but implicit. Queries are first drafted with \texttt{LMM} assistance, then refined by humans for realism and clarity, while tool usage remains latent in the final task statement.

%\paragraph{Stage 1: Seed collection (perception-level prompts).}
\noindent\textbf{Stage 1: Seed collection (perception-level prompts).}
We start from existing medical \texttt{VQA} datasets and curated clinical education resources, extracting image--question pairs $(\mathcal{X}, q)$ that target localized findings, attributes, or recognition (e.g., abnormality presence or organ identification). These seeds provide diverse medical imagery and grounded intent, but they are usually solvable in a single step and therefore do not evaluate planning or tool-mediated reasoning.

\begin{figure*}[t]
    \centering
    \includegraphics[width=\textwidth]{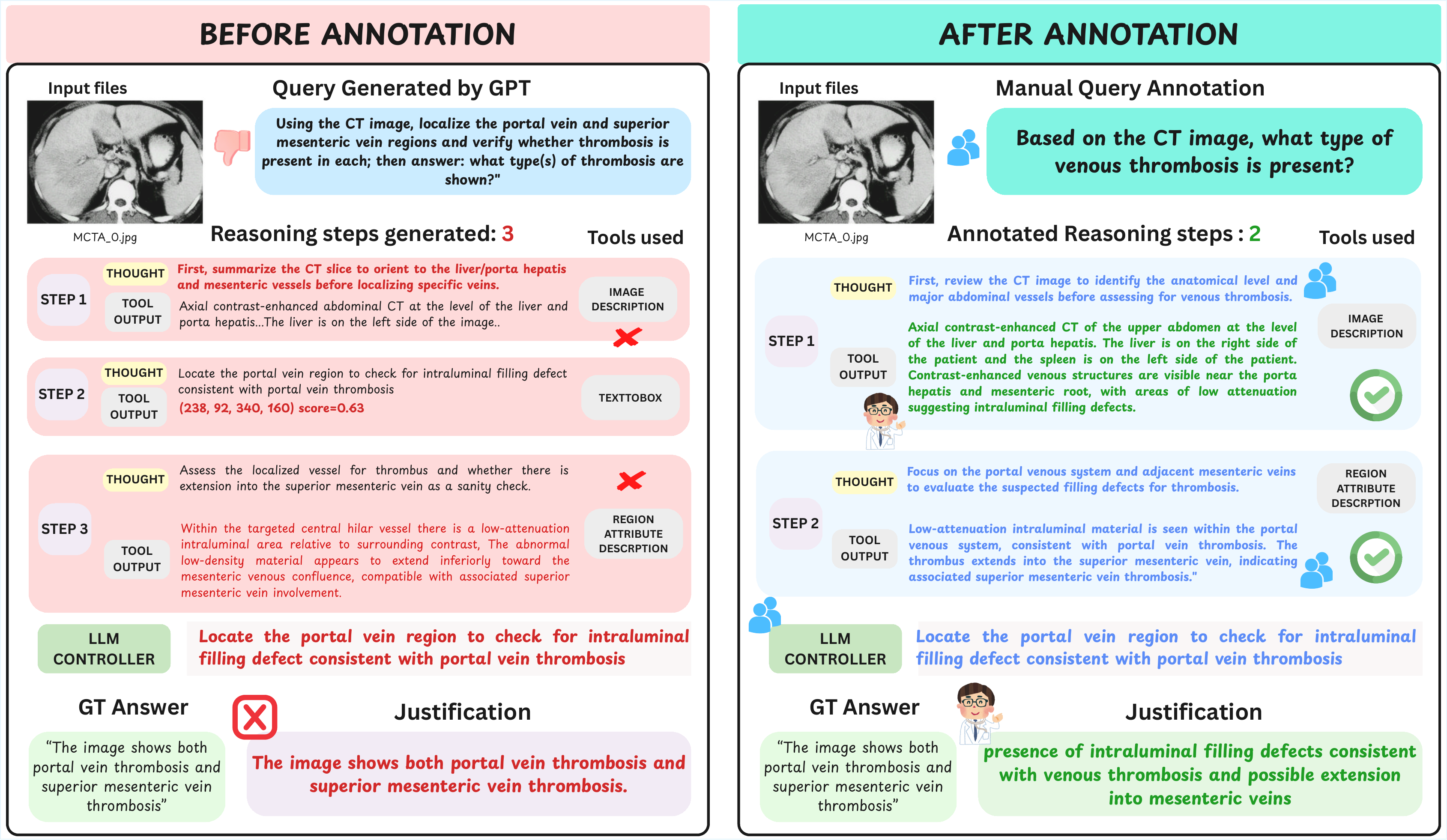}
   \caption{\textbf{Comparison of GPT-generated vs. human-annotated trajectories}. \textcolor{blue}{Blue}: technical fixes (tool compliance, stability, redundancy removal); \textcolor{green}{Green}: clinical corrections (sound reasoning, evidence grounding). The annotated version is clinically coherent, avoiding unnecessary steps while reaching the same diagnosis.}
    \label{fig:full_traj_comparison}
    % \vspace{-0.6cm}
\end{figure*}

%\paragraph{Stage 2: LMM drafting (agentic lift).}
\noindent\textbf{Stage 2: LMM drafting (agentic lift).}
Given a seed pair, we use an \lmm to propose an \emph{agentic} reformulation of the question. The input includes the original question, the corresponding visual input $\mathcal{X}$, and the full tool library $\mathcal{D}$. The model is prompted to rewrite the question into a clinically meaningful objective that \emph{requires} tool use and multi-hop reasoning, producing an initial set of agentic queries. Table~\ref{tab:query_comparison} (column 3) shows examples of these \texttt{LMM}-generated drafts. In practice, such drafts often capture useful latent workflows, but may also over-specify steps, add unnecessary subgoals, or assume unrealistic clinical operations. 

%\paragraph{Stage 3: Human rewrite (goal-first, tool-agnostic).}
\noindent\textbf{Stage 3: Human rewrite (goal-first, tool-agnostic).}
To ensure the benchmark tests autonomous planning rather than instruction following, expert annotators rewrite the \texttt{LMM}-generated drafts into a \emph{goal-oriented clinical request}. This rewrite removes step-by-step phrasing and tool-suggestive wording (e.g., explicit mention of reading labels, drawing boxes, or extracting text), while preserving the intended reasoning depth. Concretely, the final query is revised so that 1) it is solvable through some subset $\mathcal{U}\subseteq \mathcal{D}$, but $\mathcal{U}$ and the invocation order are not disclosed; 2) tool usage remains \emph{latent}, i.e., the query states the clinical goal rather than the procedure; 3) the wording matches real clinical communication and avoids raw pipeline-style scaffolding.
% \begin{itemize}
%     \item it is solvable through some subset $\mathcal{U}\subseteq \mathcal{D}$, but $\mathcal{U}$ and the invocation order are not disclosed;
%     \item tool usage remains \emph{latent}, i.e., the query states the clinical goal rather than the procedure;
%     \item the wording matches real clinical communication and avoids raw pipeline-style scaffolding.
% \end{itemize}
This aligns the task with the benchmark principle that the agent must infer both \emph{what to do next} and \emph{which tools to use}, rather than being guided by explicit procedural instructions. 

%\paragraph{Stage 4: Clinical validation and final annotation.}
\noindent\textbf{Stage 4: Clinical validation and final annotation.}
Finally, clinician reviewers verify each query for medical correctness, coherence, and executability under the tool library. They remove ambiguous or unstable cases and annotate reference outcomes $\mathcal{A}$ where needed, including multiple acceptable answers when clinical descriptions are inherently non-unique. The resulting tasks therefore require multi-step clinical reasoning with implicit tool interaction, enabling robust evaluation of agent planning and decision making in medical settings.

\noindent\textbf{Outcome.}
For each sample $(\mathcal{X}, q)$, this pipeline transforms a single-hop perception query into a clinically grounded agentic query $\mathcal{Q}$ that is solvable by a latent tool subset $\mathcal{U}$ and evaluated against reference answer(s) $\mathcal{A}$. \lmms provide scalable drafts, humans enforce goal-first phrasing and remove tool leakage, and clinicians ensure correctness and reliability. The result is a benchmark of realistic, challenging tasks that diagnose agentic capability rather than prompt compliance. Table~\ref{tab:query_comparison} shows examples of this process.

\subsection{Medical Tool Chain Construction}
\label{sec:tcons}

Given a finalized task instance from previous step $(\mathcal{X},\mathcal{Q},\mathcal{U},\mathcal{A})$ (Sec.~\ref{sec:dform}, Sec.~\ref{sec:qcons}), we construct an executable reference trajectory $\pi$ that operationalizes \emph{how} the answer $\mathcal{A}$ can be reached using tools in $\mathcal{U}\subseteq\mathcal{D}$, while remaining faithful to clinical reasoning. Importantly, $\pi$ is \emph{not} revealed to the agent at test time; rather, it serves as a structured reference for evaluating tool selection, argument validity, intermediate evidence consistency, and final outcome correctness. We emphasize two principles: \textbf{minimality} (avoid redundant steps) and \textbf{stability} (avoid brittle dependencies).

\noindent\textbf{Stage 1: LMM-Based Trajectory Drafting (Prompt-Guided Initialization).}
We first generate a candidate tool trajectory using an \lmm conditioned on the finalized query $\mathcal{Q}$, tool library $\mathcal{U}$, and answer $\mathcal{A}$. This step provides a scalable initial decomposition of the clinical objective into intermediate tool calls. However, \llm-generated trajectories frequently exhibit over-decomposition (e.g., unnecessary localization steps), unstable intermediate dependencies (e.g., bounding-box reliance), or loosely justified reasoning. For example, an \lmm draft may insert a \texttt{TextToBox} localization step before evaluating venous filling defects, thereby creating an avoidable failure point without improving diagnostic fidelity (Figure~\ref{fig:full_traj_comparison}).

\noindent\textbf{Stage 2: Technical Annotation (Executability and Minimality Enforcement).}
Next, technical annotators refine the \lmm-generated candidate trajectories to ensure strict tool-schema compliance and execution stability. Each step $s_i=(\tau_i,\alpha_i,\rho_i)$ is validated to confirm:
(i) correct tool selection,
(ii) valid and stable argument formatting,
(iii) coherent interpretation of tool outputs,
and (iv) absence of redundant or brittle steps.
Unnecessary operations (e.g., intermediate bounding box extraction when region-level evaluation suffices) are removed. The refined chain is then executed and logged in structured JSON format to ensure deterministic reproducibility. 

\noindent\textbf{Stage 3: Clinical Verification (Medical Soundness and Workflow Alignment).}
Clinician reviewers evaluate the technically validated trajectory to ensure medical correctness and realistic workflow alignment. They verify that intermediate outputs correspond to clinically meaningful findings (e.g., identification of intraluminal filling defects), that essential reasoning steps are not missing, and that the final conclusion $\mathcal{A}$ is supported by evidence. Clinicians may revise reasoning chains if they detect unsafe assumptions, incomplete diagnostic justification, or medically implausible tool usage.

\noindent\textbf{Stage 4: Quality Control and Finalization.}
Only trajectories that pass both technical and clinical review are included in \ours. This dual-layer validation ensures that reference trajectories are executable, minimal, clinically grounded, and robust to tool instability. The curated trajectories are typically shorter, more stable, and more medically coherent than naive \lmm-generated drafts, while preserving the multi-step structure required for trajectory-aware evaluation (Figure~\ref{fig:full_traj_comparison}). 

Appendix~\ref{datacard} presents the data card, Appendix~\ref{tool_details} details the executable tools, Appendix~\ref{app:clinician_verification} covers clinician verification and agreement, Appendix~\ref{app: mcta_instruction} describes the annotation protocol, Appendix~\ref{mcta_pipeline} explain the  stepwise construction of executable clinical tool
trajectories, and Appendix~\ref{mcta_prompts} provides \lmm-based generation prompts.

%See Appendix~\ref{app: mcta_instruction} for the complete annotation protocol and more detailed examples.
%More details and prompts are provided in the Appendix \S~\ref{mcta_prompts}.

%Inter annotator agreement and annotator guidelines are provided in the Appendix \S~\ref{app:clinician_verification} and \S~\ref{app: mcta_instruction} respectively.

% \vspace{-0.25cm}
\subsection{Dataset Statistics}
% \vspace{-0.2/5cm}
\label{sec:dstat}
%\input{tables/data_statistics}
%\ours contains 107 clinician-validated tasks grounded in multimodal medical data, with 107 images and 5 executable tools spanning perception, reasoning, and structured interaction. The benchmark covers 34 anatomical regions and body systems, including the brain, chest, abdomen, extremities, pelvis, ocular structures, and cellular-level pathology, enabling broad cross-specialty reasoning. Tasks require 2--4 tool-execution steps, with an average of 3.1 steps per task, reflecting the compositional nature of the benchmark. 
%Figure~\ref{fig:mcta_modality} 
%The modality distribution is led by CT (20.0\%), report-based inputs (20.0\%), histopathology (18.3\%), X-ray (9.6\%), MRI (8.7\%), fundus photography (7.0\%), and gross pathology (5.2\%), with additional coverage of microscopy/TEM (3.5\%), dermoscopy (3.5\%), PET (0.9\%), mammography (0.9\%), and general pathology (0.9\%). Detailed breakdown of body regions, tool descriptions, and modalities are in the Appendix.
With a total of 321 human annotation hours,
\ours contains 107 clinician-validated tasks grounded in multi-modal medical data, including 107 images and 5 executable tools spanning perception, reasoning, and structured interaction. The benchmark covers 34 anatomical regions and body systems, such as the brain, chest, abdomen, extremities, pelvis, ocular structures, and cellular-level pathology, enabling broad cross-specialty reasoning. Tasks require 2–4 tool-execution steps (average 3.1), reflecting the benchmark’s compositional nature.
\ours also exhibits diverse modality coverage, led by CT (20.0\%), report-based inputs (20.0\%), histopathology (18.3\%), X-ray (9.6\%), MRI (8.7\%), fundus photography (7.0\%), and gross pathology (5.2\%), with additional representation from microscopy/TEM (3.5\%), dermoscopy (3.5\%), PET (0.9\%), mammography (0.9\%), and general pathology (0.9\%) etc. More details are in the Appendix~\ref{app:metrics_details}.

% \vspace{-.35cm}
\section{Experiments}
\label{sec:experiments}

\begin{table}[t]
\centering
\caption{
Evaluation metrics are grouped into
step-by-step tool-use fidelity, clinical reasoning quality, and outcome
accuracy. All scores are reported as percentages in $[0,100]$ unless otherwise
specified. More detailed definitions are provided in Appendix \S~\ref{app:metrics_details}.
}
\label{tab:metrics}
\small

\resizebox{\textwidth}{!}{%
\begin{tabular}{m{5.5cm} m{11.5cm}}
\toprule

\rowcolor{blue!15}
\textbf{Metric (Symbol)} &
\textbf{Description} \\
\midrule

\rowcolor{gray!15}
\multicolumn{2}{c}{\textbf{Step-by-Step Mode}} \\
\midrule

\texttt{InstAcc} ($\mathbf{I_{acc}}$)
&
\textit{Whether the model follows the required agent protocol and output format at each step.} \\

\texttt{ToolAcc} ($\mathbf{T_{acc}}$)
&
\textit{Whether the model selects the correct tool for step $n{+}1$ given the first $n$ reference steps.} \\

\texttt{ArgAcc} ($\mathbf{A_{acc}}$)
&
\textit{Whether the predicted tool arguments are syntactically valid and clinically appropriate.} \\

\texttt{SummAcc} ($\mathbf{S_{acc}}$)
&
\textit{Whether the model correctly summarizes intermediate evidence into the expected conclusion.} \\

\midrule

\rowcolor{gray!15}
\multicolumn{2}{c}{\textbf{Clinical Reasoning Mode}} \\
\midrule

\texttt{Clinical Faithfulness} ($\mathbf{F_{acc}}$)
&
\textit{Logical consistency of intermediate reasoning with clinical evidence and workflow.} \\

\texttt{Context Integration Score} ($\mathbf{C_s}$)
&
\textit{Effective integration of multimodal clinical context (images, reports, structured data).} \\

\texttt{Semantic Completeness} ($\mathbf{S_{comp}}$)
&
\textit{Coverage of all clinically necessary findings required for a valid conclusion.} \\

\midrule

\rowcolor{gray!15}
\multicolumn{2}{c}{\textbf{Outcome Mode}} \\
\midrule

\texttt{Goal Accuracy} ($\mathbf{G_{acc}}$)
&
\textit{Final answer accuracy for diagnostic and interpretive clinical queries.} \\

\bottomrule
\end{tabular}%
}
\end{table}

%
% \vspace{-.20cm}
\subsection{Benchmark Evaluation}
% \vspace{-.20cm}
\label{sec:setup}

\textbf{Setup.}
We evaluate \ours with \textbf{18} large language models, covering both proprietary APIs and open-source checkpoints from the GPT~\cite{gpt4, gpt5_system_card, openai2025gptoss120bgptoss20bmodel}, Claude~\cite{claude3}, Gemini~\cite{gemini}, Qwen~\cite{qwen}, LLaMA~\cite{llama3}, DeepSeek~\cite{deepseek-vl}, Mistral~\cite{mistral}, and Phi~\cite{phi} families, on \textbf{107} clinician-verified tasks with \textbf{1926} autonomous rollouts in total. We use OpenCompass~\cite{2023opencompass} as the evaluation harness and Lagent~\cite{lagent} as the agent framework, adopting ReAct~\cite{react}-style prompting and an Agentlego~\cite{agentlego} tool interface. In this autonomous setting, the model is fully responsible for the entire control loop: it must decide \emph{which tool} to call, produce \emph{valid arguments}, interpret tool outputs, decide \emph{when to stop}, and synthesize the final answer. See Appendix \S\ref{tool_details} for the executable tool library and Appendix \S\ref{mcta_prompts} for prompts.

\textbf{Evaluation metrics.}
We report performance using three complementary metric groups (Table~\ref{tab:metrics}). \textsc{Step-by-Step} measures interaction fidelity (instruction following, tool selection, and argument validity), following the trajectory-level evaluation protocol of GTA~\cite{GTA}. \textsc{Clinical Reasoning} evaluates evidence usage and summary quality. \textsc{Outcome} evaluates task success via the official final outcome score $G_{acc}$. Together, these metrics disentangle controller competence from evidence-conditioned reasoning. More details are in Appendix \S~\ref{app:metrics_details}. To further validate the automated clinical-reasoning evaluation, we conducted an additional clinician audit on 36 benchmark tasks across four representative models, yielding 144 evaluated rollouts and 432 scalar clinical judgments over $F_{\mathrm{acc}}$, $C_s$, and $S_{\mathrm{comp}}$. The audit confirms the same qualitative conclusion as the automated judge: clinical reasoning remains weak even for strong models, with an overall human clinical-reasoning score of 28.5/100 and moderate model-level agreement with the automated evaluator on the aggregate clinical score (Spearman $\rho=0.60$; Appendix~\ref{app:human_eval}).

\begin{table*}[t]
\centering
\caption{\textbf{Main results of \ours.} $*$ refers to a closed-source model.}
\label{tab:main_result}

\renewcommand{\arraystretch}{1.25}

\resizebox{\textwidth}{!}{
\begin{tabular}{llcccccccc}
\toprule

\multirow{2}{*}{\textbf{Family}} &
\multirow{2}{*}{\textbf{Model}} &
\multicolumn{4}{c}{\textbf{Step-by-Step}} &
\multicolumn{3}{c}{\textbf{Clinical Reasoning}} &
\textbf{Outcome} \\

\cmidrule(lr){3-6}
\cmidrule(lr){7-9}
\cmidrule(lr){10-10}

& & Inst. & Tool. & Arg. & Summ.
& $F_{acc}$ & $C_s$ & $S_{comp}$ & $G_{acc}$ \\

\midrule

\rowcolor{openai}
 & GPT-5.4$^{*}$ & \underline{35.27} & \textbf{23.46} & \underline{12.61} & 35.51 & \underline{17.52} & 14.21 & 18.60 & \textbf{31.54} \\
\rowcolor{openai}
 & GPT-5.4-mini$^{*}$ & 5.36 & 6.74 & 3.23 & 0.93 & 16.47 & 10.56 & 17.29 & 28.31 \\
\rowcolor{openai}
OpenAI & GPT-5.4-nano~$^{*}$ & 33.93 & \underline{18.18} & 12.02 & 34.24 & \textbf{18.43} & 11.96 & 14.77 & 20.30 \\
\rowcolor{openai}
 & GPT-oss-20B & 1.79 & 0.00 & 0.00 & 0.00 & 1.31 & 0.56 & 1.68 & 3.18 \\

\midrule

\rowcolor{anthropic}
 & Claude-opus-4-6$^{*}$ & 24.78 & 8.80 & 0.59 & \underline{39.25} & 14.11 & \underline{14.86} & \textbf{23.83} & \underline{31.32} \\
\rowcolor{anthropic}
Anthropic & Claude-sonnet-4-6$^{*}$ & 23.66 & 4.99 & 0.00 & 33.64 & 12.77 & 12.90 & \underline{20.19} & 25.33 \\
\rowcolor{anthropic}
 & Claude-haiku-4-5$^{*}$ & 27.46 & 13.78 & 4.69 & \textbf{43.93} & 9.35 & 3.36 & 14.11 & 23.08 \\

\midrule

\rowcolor{google}
Google & Gemini-3-flash$^{*}$ & 3.35 & 17.30 & 0.00 & 5.61 & 11.31 & 8.60 & 15.98 & 25.87 \\
\rowcolor{google}
 & Gemini-3-flash-lite$^{*}$ & 2.90 & 8.21 & 0.00 & 1.87 & 10.75 & 6.82 & 14.58 & 23.64 \\

\midrule

\rowcolor{qwen}
Qwen & Qwen3.5-9B & \textbf{44.20} & 14.37 & \textbf{13.78} & 29.91 & 10.37 & \textbf{17.10} & 13.36 & 21.64 \\
\rowcolor{qwen}
 & Qwen3-8B & 33.93 & 10.56 & 7.04 & 32.71 & 8.50 & 10.09 & 11.50 & 27.80 \\

\midrule

\rowcolor{deepseek}
 & DeepSeek-R1-Distill-7B & 10.49 & 3.52 & 0.00 & 7.48 & 2.62 & 0.84 & 3.36 & 10.61 \\
\rowcolor{deepseek}
DeepSeek & Deepseek-llm-7b-chat & 11.61 & 6.45 & 0.00 & 4.67 & 4.30 & 2.62 & 4.02 & 11.00 \\
\rowcolor{deepseek}
 & DeepSeek-V2-Lite-Chat & 11.83 & 11.14 & 0.29 & 0.00 & 3.83 & 3.55 & 6.54 & 6.96 \\

\midrule

\rowcolor{meta}
Meta & Llama-3.1-8B-Instruct & 23.66 & 7.92 & 0.00 & 6.54 & 7.94 & 5.42 & 11.21 & 18.94 \\
\rowcolor{meta}
 & Llama-3.2-3B-Instruct & 18.53 & 1.76 & 0.00 & 4.67 & 3.08 & 1.68 & 5.14 & 11.29 \\

\midrule

\rowcolor{mistral}
Mistral & Mistral-7B & 18.75 & 14.66 & 0.00 & 9.35 & 2.52 & 1.87 & 3.46 & 9.40 \\

\midrule

\rowcolor{yi}
Microsoft & Phi-4 & 20.09 & 6.45 & 0.00 & 14.02 & 6.36 & 3.36 & 6.17 & 10.65 \\

\bottomrule
\end{tabular}
}
\end{table*}
\begin{figure*}[t]
    \centering
    \begin{subfigure}[t]{0.24\linewidth}
        \centering
        \includegraphics[width=\linewidth]{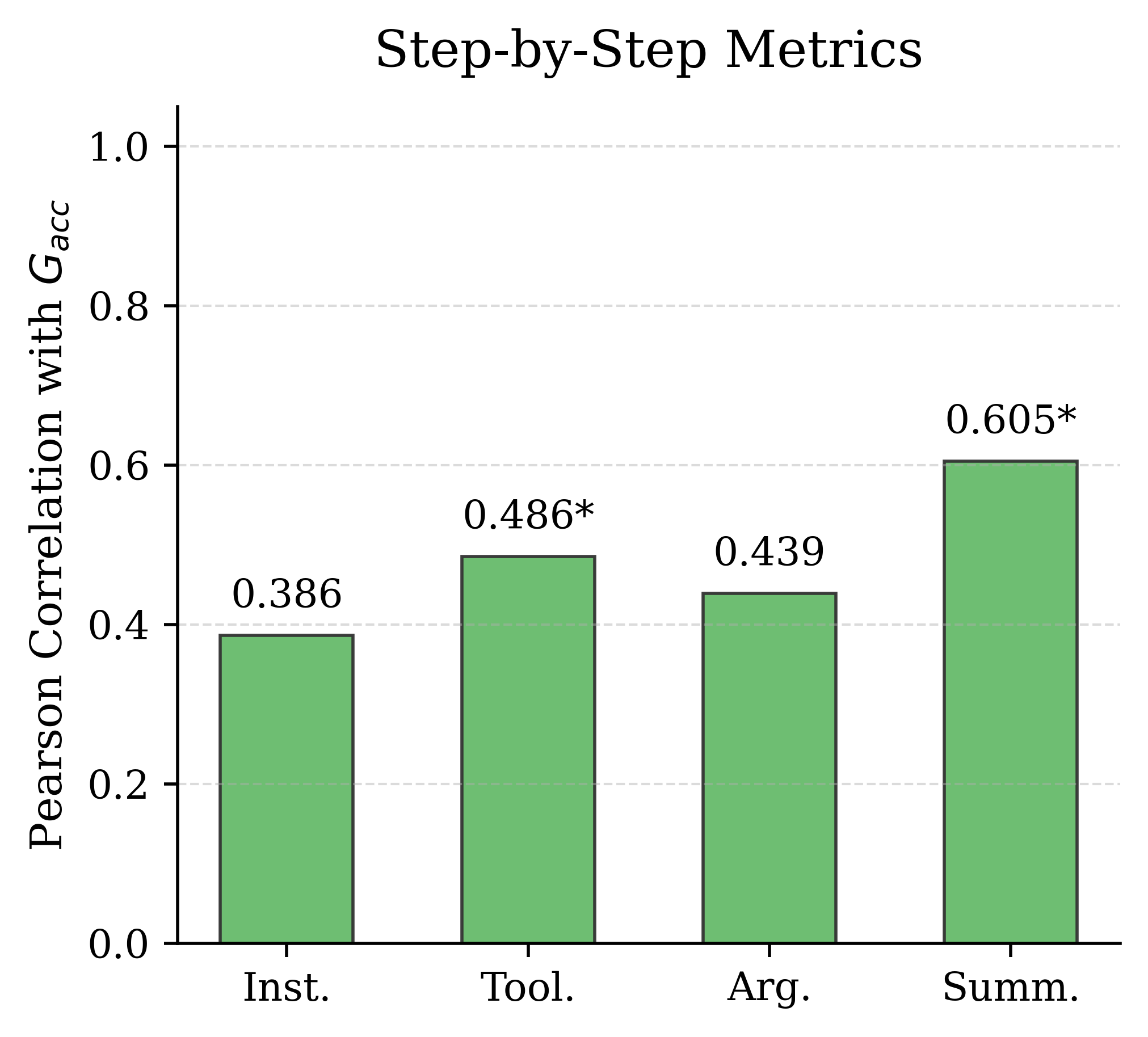}
        \caption{Step (Pearson)}
        \label{fig:step_corr}
    \end{subfigure}
    \hfill
    \begin{subfigure}[t]{0.24\linewidth}
        \centering
        \includegraphics[width=\linewidth]{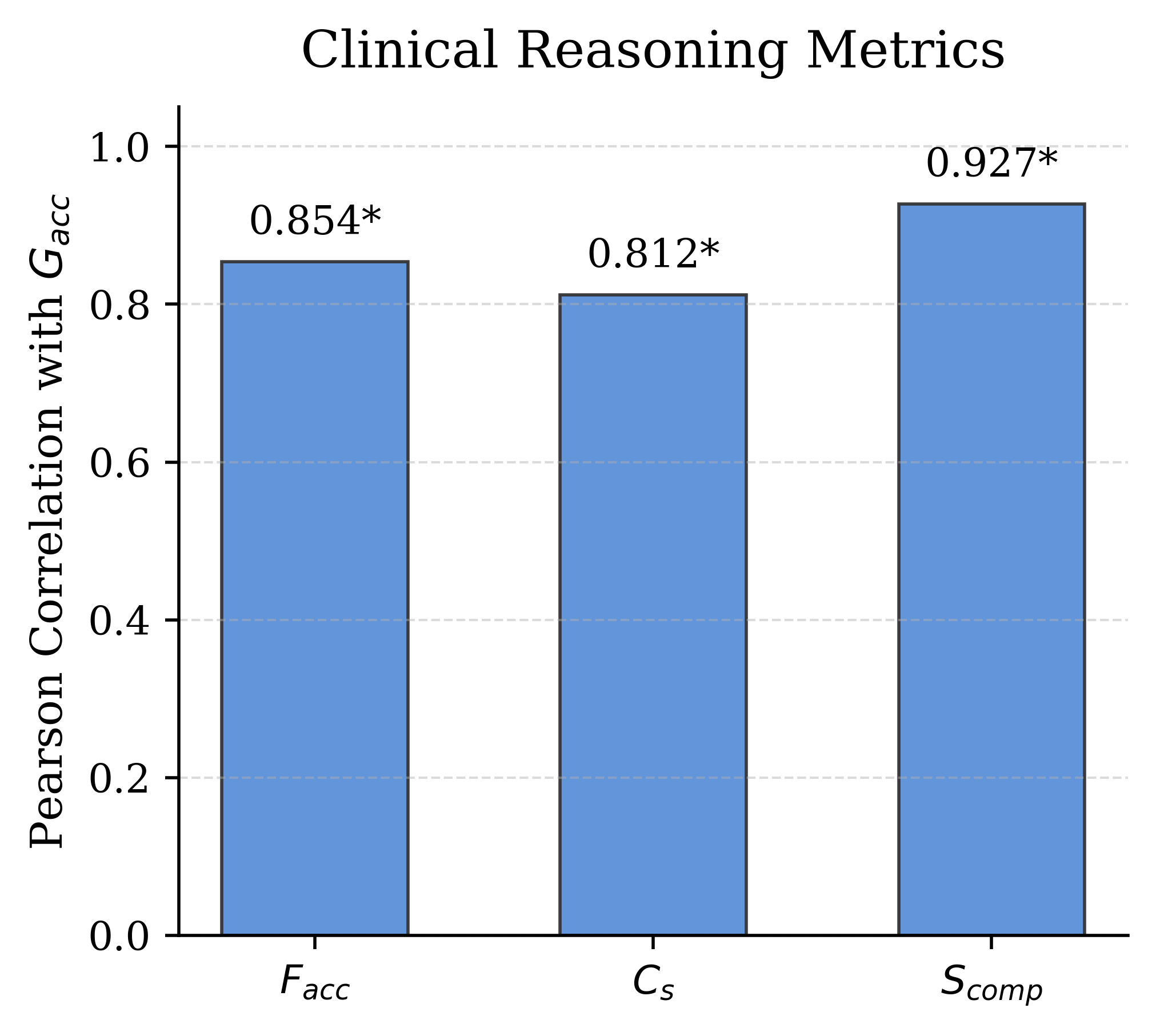}
        \caption{Reasoning (Pearson)}
        \label{fig:reason_corr}
    \end{subfigure}
    \hfill
    \begin{subfigure}[t]{0.24\linewidth}
        \centering
        \includegraphics[width=\linewidth]{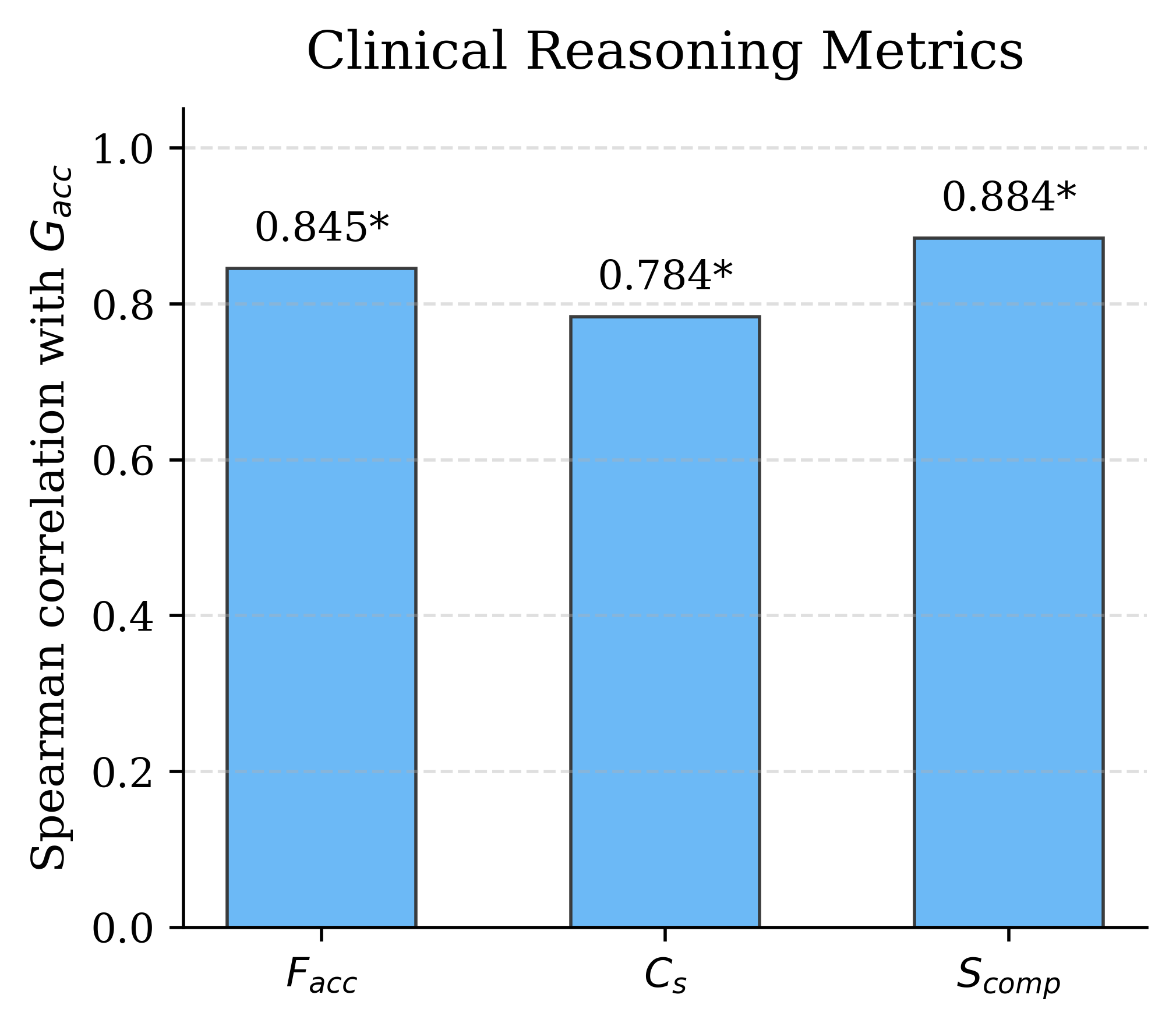}
        \caption{Reasoning (Spearman)}
        \label{fig:spearman_corr}
    \end{subfigure}
      \hfill
    \begin{subfigure}[t]{0.24\linewidth}
        \centering
        \includegraphics[width=\linewidth]{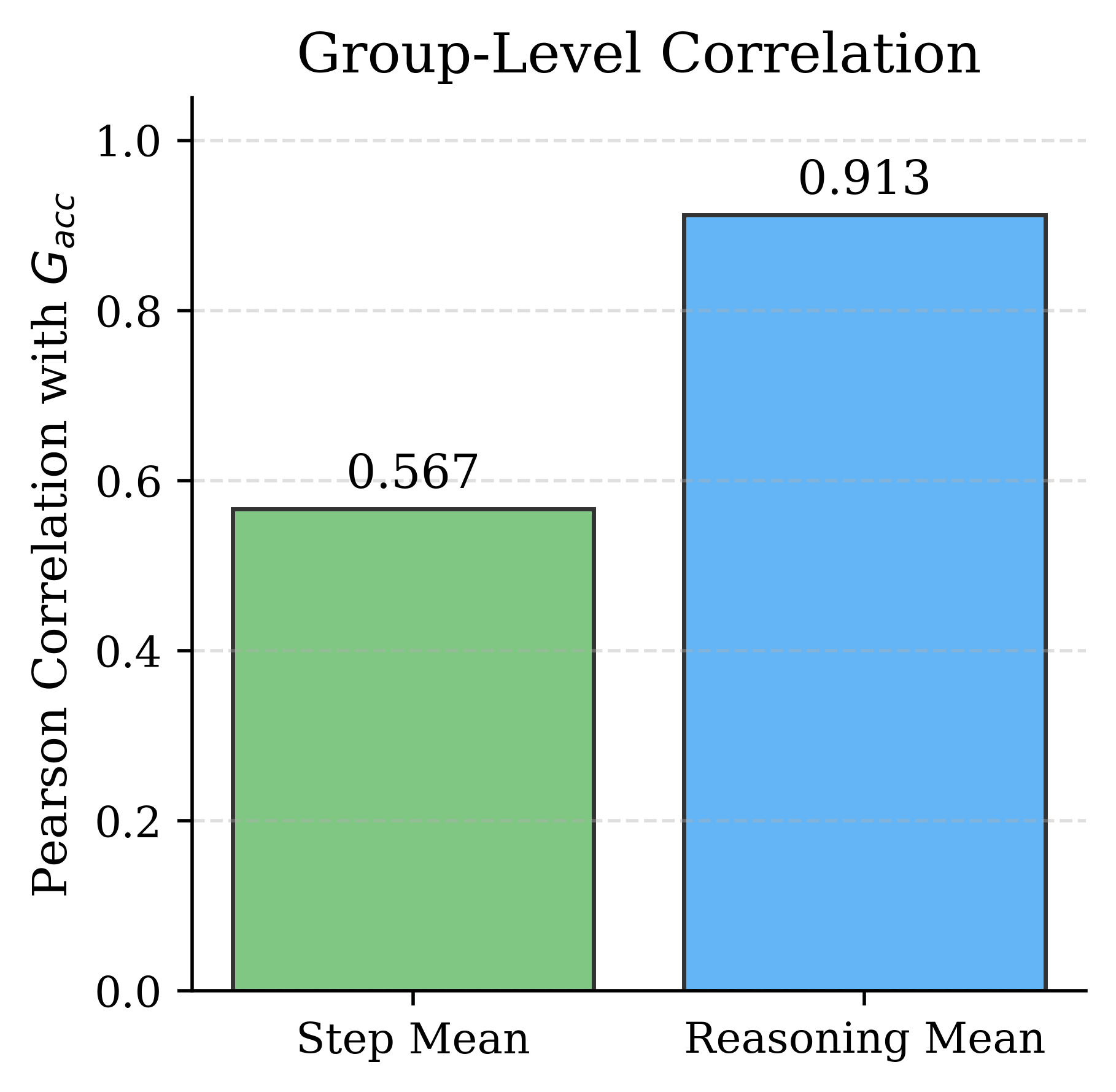}
        \caption{Group-level}
        \label{fig:group_corr}
    \end{subfigure}

    \caption{
  \textbf{Correlation with final accuracy ($G_{acc}$).}
Step-level metrics show moderate correlation, while clinical reasoning metrics (especially $S_{comp}$ and $F_{acc}$) strongly correlate with final performance. This indicates that controller instability prevents models from reaching stable reasoning.}
    \label{fig:correlation_combined}
    % \vspace{-0.5 cm}
\end{figure*}

\textbf{Results.}  Table~\ref{tab:main_result} shows that \ours remains far from saturated, with the best $G_{acc}$ only \textbf{31.54}, indicating substantial headroom. Closed models generally lead, but strong open models (e.g., Qwen3-8B at \textbf{27.80}) are competitive, suggesting controller stability can offset scale. Performance is multi-dimensional, no model dominates across tool use, instruction following, argument construction, and summarization, and these do not consistently translate to outcome. Notably, outcome can overestimate reliability, as some models achieve high $G_{acc}$ despite weak step-level fidelity, while others show strong interaction metrics without corresponding success. While scaling often helps, it is not monotonic, and interaction style (e.g., stopping and evidence synthesis) can matter as much as model size. Furthermore, Fig.~\ref{fig:correlation_combined} shows the relationship between step-level execution, clinical reasoning metrics, and final answer accuracy. 
The results indicate that reasoning-centric metrics, particularly $S_{comp}$ and $F_{acc}$, exhibit substantially stronger correlations with $G_{acc}$ than step-by-step signals, highlighting that final performance is primarily driven by reasoning quality rather than intermediate execution. Reasoning metrics correlate most with final answer quality, while rollout diagnostics show controller failures dominate trajectory collapse.

% \vspace{-0.25cm}
\subsection{Failure Analysis and Discussion}
\label{sec:analysis}
% \vspace{-0.25cm}
\begin{table*}[t]
\centering
\small
\setlength{\tabcolsep}{6pt}
\renewcommand{\arraystretch}{1.15}

\begin{minipage}{0.48\textwidth}
\centering
\captionof{table}{\textbf{Autonomous vs. Gold ($G_{acc}$}).}
\label{tab:auto_vs_oracle}
\rowcolors{2}{rowblue}{white}
\resizebox{\linewidth}{!}{
\begin{tabular}{lccc}
\toprule
\rowcolor{headerblue}
\textbf{Model} & \textbf{Auto} & \textbf{Gold} & $\boldsymbol{\Delta}$ \\
\midrule
GPT-5.4 & 31.54 & 49.50 & \cellcolor{rowgreen}\textbf{+17.96} \\
Claude-opus-4-6 & 31.32 & 66.40 & \cellcolor{rowgreen}\textbf{+35.08} \\
Qwen3.5-9B & 21.64 & 49.50 & \cellcolor{rowblue}+27.86 \\
\bottomrule
\end{tabular}
}
\end{minipage}
\hfill
\begin{minipage}{0.48\textwidth}
\centering
\captionof{table}{\textbf{Global rollout diagnostics.}}
\label{tab:global_diagnostics}
\rowcolors{2}{rowblue}{white}
\resizebox{\linewidth}{!}{
\begin{tabular}{lcl}
\toprule
\rowcolor{headerblue}
\textbf{Metric} & \textbf{Value} & \textbf{Insight} \\
\midrule
API error rate & \cellcolor{rowred}\textbf{64.2} & protocol instability \\
Under-call rate & \cellcolor{rowred}\textbf{99.2} & premature stopping \\
Protocol failure & \cellcolor{rowred}\textbf{58.3} & rollout breakdown \\
Tool-selection failure & 41.6 & incorrect actions \\
\bottomrule
\end{tabular}
}
\end{minipage}
\vspace{-0.5 cm}
\end{table*}
% The main benchmark results establish that \ours is challenging, but final outcome scores alone do not explain \emph{why} models fail. We therefore analyze failures through three complementary diagnostics: \textbf{gold-standard routing}, which supplies the reference next tool at each step while still requiring the model to generate arguments and synthesize the final answer; \textbf{rollout-level failure analysis}, which identifies where autonomous trajectories first break; and \textbf{backbone-only VLM evaluation}, which measures single-turn perception without tool use. In the gold-standard setting, the evaluator provides only the next reference tool identity and stopping boundary; it does \emph{not} provide reference arguments, tool outputs, schema repairs, clinical hints, or final answers. Thus, the comparison isolates tool-selection and stopping failures while preserving argument construction, evidence interpretation, and final clinical reasoning. Additional per-model, tool-wise, and horizon-wise ablations are provided in Appendix~\ref{app:diagnostics}.

\noindent\textbf{Gold-standard routing exposes a large controller gap.}
Table~\ref{tab:auto_vs_oracle} shows that replacing autonomous tool selection with a gold-standard tool trajectory produces large gains for representative frontier models. Claude-opus-4-6 rises from \textbf{31.32} to \textbf{66.40}, Qwen3.5-9B from \textbf{21.64} to \textbf{49.50}, and GPT-5.4 from \textbf{31.54} to \textbf{49.50}. This gap is the clearest evidence that current models possess substantial latent clinical reasoning ability, but fail to reliably access it under autonomous control. The bottleneck is therefore not simply medical knowledge or perception: it is the ability to maintain a valid tool-using loop, select the right evidence source, and stop only after sufficient evidence has been gathered. 

\noindent\textbf{Autonomous rollouts fail before clinical reasoning can fully begin.}
Table~\ref{tab:global_diagnostics} summarizes the dominant rollout errors: \textbf{64.2\%} of episodes encounter API/protocol errors, and \textbf{99.2\%} under-call relative to the reference trajectory. The first-failure breakdown is equally sharp: \textbf{58.3\%} of failures occur first at protocol/API handling and \textbf{41.6\%} at tool selection. Most models do not primarily fail after carefully reasoning over wrong evidence; they fail earlier, by leaving the clinical tool-use process before the necessary evidence has been acquired.

\noindent\textbf{Backbone perception is not the same as agent competence.} Table~\ref{tab:vlm_backbone} shows that
\begin{wraptable}{r}{0.48\textwidth}
% \vspace{-10pt}
\centering
\small
\setlength{\tabcolsep}{5pt}
\renewcommand{\arraystretch}{1.1}

\caption{\textbf{Backbone-only VLM performance.}}
\label{tab:vlm_backbone}

\rowcolors{2}{rowblue}{white}
\begin{tabular}{lcc}
\toprule
\rowcolor{headerblue}
\textbf{Model} & \textbf{Public} & $\boldsymbol{G_{acc}}$ \\
\midrule
GPT-5.4 &  & \cellcolor{rowgreen}\textbf{60.74} \\
Gemini-3-flash &  & 30.84 \\
Claude-haiku-4-5 &  & 47.65 \\
\midrule
Qwen3-VL-8B-Instruct & \checkmark & 40.18 \\
Llama-3.2-11B-Vision & \checkmark & 39.25 \\
DeepSeek-VL2 & \checkmark & \cellcolor{rowgreen}\textbf{48.59} \\
\midrule
MedGemma-4B-IT & \checkmark & \cellcolor{rowgreen}\textbf{45.79} \\
LLaVA-Med-v1.5-7B& \checkmark & 39.25 \\
MedMO-8B-Next & \checkmark & 27.10 \\
Fleming-VL-8B & \checkmark & 29.90 \\
\bottomrule
\end{tabular}

% \vspace{-12pt}
\end{wraptable}
 several backbone-only VLMs achieve strong zero-shot performance when tool interaction is removed: GPT-5.4 reaches \textbf{60.74}, DeepSeek-VL2 reaches \textbf{48.59}, Claude-haiku-4-5 reaches \textbf{47.65}, and MedGemma-4B-IT reaches \textbf{45.79}. These scores are substantially higher than autonomous tool-agent outcome scores in Table~\ref{tab:main_result}, revealing a key separation: recognizing medical content in a single turn is much easier than planning a multi-step clinical workflow. %\ours therefore evaluates a harder capability than static medical VQA. It requires tool arbitration, evidence accumulation, localized grounding, and calibrated stopping, which are precisely the behaviors that single-turn perception evaluation cannot measure. 
 In \ours, success requires selecting and sequencing tools, accumulating sufficient evidence, performing localized grounding when needed, and stopping only after the objective is justified. This capability cannot be inferred from static VQA-style evaluation, which is why \ours targets a strictly harder and more deployment-relevant setting.

%Moreover, comparing Table~\ref{} and Table~\ref{} shows that when LLM agents are given access to stronger tools that produce reliable intermediate evidence, multi-step tool-mediated execution can solve substantially more tasks than a single-turn setting. This highlights that performance in \ours is not only constrained by backbone perception, but also by the quality of the tool ecosystem and the agent’s ability to exploit meaningful trajectories.

 \begin{figure}[t]
\centering

\begin{tcolorbox}[
colback=orange!2,
colframe=orange!80!black,
title=\textbf{Qualitative Comparison: Sentinel-node metastasis detection (semantic drift)},
boxrule=0.5pt,
arc=1mm,
left=3pt,right=3pt,top=3pt,bottom=3pt
]

\small
\textbf{Query:} \emph{Identify the methods used to detect metastases in sentinel nodes.}

\vspace{2pt}

\begin{minipage}[t]{0.32\textwidth}
\textbf{Gold}
\begin{itemize}[leftmargin=1em,itemsep=1pt,topsep=1pt]
\item \texttt{ImageDescription} + \texttt{OCR} extract H\&E, CK-18, CEA, hTRT, MUC-1.
\item \texttt{Search + Region} maps to histologic + molecular assays.
\item \textbf{Answer:} H\&E staining + RT-PCR/Southern blot.
\end{itemize}
\end{minipage}\hfill
\begin{minipage}[t]{0.32\textwidth}
\textbf{GPT-5.4}
\begin{itemize}[leftmargin=1em,itemsep=1pt,topsep=1pt]
\item \textcolor{red!70!black}{\textbf{Skips tool reasoning}} and answers directly.
\item \textcolor{orange!80!black}{\textbf{Over-generalizes}} to generic ``IHC + RT-PCR''.
\item \textbf{Answer:} IHC + RT-PCR.
\item \textcolor{red!70!black}{\textbf{Failure:}} premature answer, incomplete grounding.
\end{itemize}
\end{minipage}\hfill
\begin{minipage}[t]{0.32\textwidth}
\textbf{Qwen3.5-9B}
\begin{itemize}[leftmargin=1em,itemsep=1pt,topsep=1pt]
\item \textcolor{red!70!black}{\textbf{API errors}} + weak initial grounding.
\item \textcolor{orange!80!black}{\textbf{Ignores OCR evidence}}.
\item \textbf{Answer:} unrelated clinical procedures.
\item \textcolor{red!70!black}{\textbf{Failure:}} semantic drift from visual evidence.
\end{itemize}
\end{minipage}

\vspace{2pt}
\textbf{Issues:} \textit{Errors arise from broken reasoning chains (GPT-5.4) and drift from extracted evidence (Qwen3.5-9B), showing that correct perception alone is insufficient without grounded execution.}

\end{tcolorbox}

\caption{\textbf{Qualitative failure analysis on sentinel-node metastasis detection.}
The gold trajectory grounds reasoning in extracted visual evidence via tool execution. GPT-5.4 fails due to premature answering, while Qwen3.5-9B exhibits semantic drift by ignoring OCR evidence and relying on unrelated priors. This shows that accurate perception alone does not ensure grounded reasoning.}

\label{fig:qualitative_failure_analysis}
\vspace{-0.5 cm}
\end{figure}

\noindent\textbf{Qualitative failures reveal evidence drift, not just wrong answers.}
Figure~\ref{fig:qualitative_failure_analysis} illustrates the same diagnostic story at trajectory level. In the sentinel-node metastasis example, the gold trajectory grounds the answer by extracting visible chart labels and mapping them to histologic and molecular assays. GPT-5.4 bypasses this evidence chain and answers prematurely with an incomplete generic method description, while Qwen3.5-9B encounters unstable execution and drifts toward unrelated clinical procedures despite recoverable OCR evidence. This example is representative of a broader pattern documented in Appendix~\ref{app:diagnostics}: failures often arise from broken evidence acquisition, semantic drift away from tool observations, or premature synthesis from prior knowledge. The central lesson is that correct perception alone is insufficient; reliable medical agents must remain evidence-obedient throughout the full trajectory.

\noindent\textbf{Takeaway.}
Together, these diagnostics show that current medical tool agents fail in two separable layers. The first and dominant layer is \emph{controller reliability}: protocol adherence, tool recruitment, and stop/continue calibration. The second layer, visible after gold-standard routing improves tool access, is \emph{grounded long-horizon clinical reasoning}: using localized evidence correctly across multiple steps. This decomposition is the main diagnostic value of \ours and identifies where future progress is most likely to matter. To mitigate evaluation bias, we use a structured LLM-as-judge framework with disentangled prompts that independently assess workflow, evidence usage, factual correctness, and completeness while explicitly ignoring other factors. All evaluations are grounded in clinician-validated reference trajectories, ensuring alignment with medically plausible reasoning. This conclusion is also supported by a clinician audit of model rollouts (Appendix~\ref{app:human_eval}): the best audited model reaches only 37.9/100 mean human clinical-reasoning score, and only 6.9\% of audited rollouts achieve a strong mean score ($\geq 0.7$), indicating that the observed failures are visible to clinical reviewers and are not merely artifacts of LLM-based judging.

\noindent\textbf{Additional Experiments.} We also include additional experiments showing a large gold-standard tool routing gap (Appendix~\ref{app:oracle}), early protocol and tool-use failures with under-calling (Appendix~\ref{app:failures}), and residual long-horizon grounding challenges after routing correction (Appendices~\ref{app:length},~\ref{app:tools}).

% \noindent \textbf{Takeaway.}
% Taken together, the main limitation of current medical tool agents is not perception, but \emph{fragile control}. Autonomous rollouts are dominated by protocol errors, wrong or missing tool use, and premature stopping. Oracle routing reveals substantial latent reasoning ability, but the remaining challenge is still hard and depends on sustained, grounded evidence acquisition over longer trajectories. This decomposition, controller reliability first, evidence-conditioned reasoning second, is the main diagnostic value of \ours.
% \vspace{-0.1cm}
\section{Conclusion}
% \vspace{-0.25cm}
We introduce \ours, a benchmark for evaluating medical tool agents beyond static medical VQA by measuring planning, tool selection, localized evidence acquisition, and end-to-end outcome quality on 107 clinician-verified multimodal tasks. Across 18 models, the benchmark shows that reliable clinical tool use remains unsolved: even strong systems achieve limited outcome accuracy, gold-standard tool routing exposes a large controller gap, and most failures arise from protocol instability, under-calling, and weak local grounding rather than final synthesis alone. These results position \ours not only as a challenging benchmark, but also as a diagnostic framework for developing more faithful, robust, and clinically grounded medical agents.

\textbf{Limitations.}
\ours is intentionally diagnostic, not exhaustive: it exposes key failures but does not cover the full diversity of clinical settings (e.g., institutional and regional diversity) or tools (e.g., imaging protocols). The library focuses on perception, OCR, retrieval, and calculation, with future extensions needed for segmentation, report/note or structured EHR retrieval, and clinical score calculators. Limited demographic metadata prevents strong fairness conclusions, and automated reasoning metrics do not replace expert review.

% \ours is intentionally diagnostic rather than exhaustive. It contains
% 107 clinician-validated tasks and 5 tools, which is sufficient to
% expose common controller failures but does not cover the full diversity of clinical specialties, institutions, patient populations, imaging protocols, or tool ecosystems. The tool library emphasizes perception, OCR, localized
% description, retrieval, and calculation; future versions should include
% segmentation, measurement, report extraction, structured EHR retrieval, and specialty-specific calculators. Because many public medical assets lack demographic metadata, \ours cannot support strong conclusions about fairness across patient groups. Our reasoning metrics use clinician-calibrated automatic judges, which improves scalability but does not replace full expert adjudication. Finally, oracle-routing experiments isolate controller failures but are not deployment interventions: real clinical systems would require stronger safeguards, uncertainty handling, human oversight, and regulatory validation.

\bibliographystyle{plain}
\bibliography{refs}

@String(CVPR  = {IEEE Conf. Comput. Vis. Pattern Recog.})

@String(ICCV  = {Int. Conf. Comput. Vis.})

@String(ICML  = {Int. Conf. Mach. Learn.})

@String(CVPR  = {CVPR})

@String(ICCV  = {ICCV})

@String(ICML  = {ICML})

@article{visualgpt,
  title={Visual chatgpt: Talking, drawing and editing with visual foundation models},
  author={Wu, Chenfei and Yin, Shengming and Qi, Weizhen and Wang, Xiaodong and Tang, Zecheng and Duan, Nan},
  journal={arXiv preprint arXiv:2303.04671},
  year={2023}
}

@article{mmreact,
  title={Mm-react: Prompting chatgpt for multimodal reasoning and action},
  author={Yang, Zhengyuan and Li, Linjie and Wang, Jianfeng and Lin, Kevin and Azarnasab, Ehsan and Ahmed, Faisal and Liu, Zicheng and Liu, Ce and Zeng, Michael and Wang, Lijuan},
  journal={arXiv preprint arXiv:2303.11381},
  year={2023}
}

@article{phi,
  title={Phi-4 technical report},
  author={Abdin, Marah and Aneja, Jyoti and Behl, Harkirat and Bubeck, S{\'e}bastien and Eldan, Ronen and Gunasekar, Suriya and Harrison, Michael and Hewett, Russell J and Javaheripi, Mojan and Kauffmann, Piero and others},
  journal={arXiv preprint arXiv:2412.08905},
  year={2024}
}

@article{wang2025medagent,
  title={Medagent-pro: Towards evidence-based multi-modal medical diagnosis via reasoning agentic workflow},
  author={Wang, Ziyue and Wu, Junde and Cai, Linghan and Low, Chang Han and Yang, Xihong and Li, Qiaxuan and Jin, Yueming},
  journal={arXiv preprint arXiv:2503.18968},
  year={2025}
}

@article{yan2026livemedbench,
  title={Livemedbench: A contamination-free medical benchmark for llms with automated rubric evaluation},
  author={Yan, Zhiling and Song, Dingjie and Fang, Zhe and Ji, Yisheng and Li, Xiang and Li, Quanzheng and Sun, Lichao},
  journal={arXiv preprint arXiv:2602.10367},
  year={2026}
}

@article{bedi2026healthadminbench,
  title={HealthAdminBench: Evaluating computer-use agents on healthcare administration tasks},
  author={Bedi, Suhana and Welch, Ryan and Steinberg, Ethan and Wornow, Michael and Kim, Taeil Matthew and Ahmed, Haroun and Sterling, Peter and Purohit, Bravim and Akram, Qurat and Acosta, Angelic and others},
  journal={arXiv preprint arXiv:2604.09937},
  year={2026}
}

@inproceedings{xu2025medagentgym,
  title={Medagentgym: Training llm agents for code-based medical reasoning at scale},
  author={Xu, Ran and Zhuang, Yuchen and Zhong, Yishan and Yu, Yue and Tang, Xiangru and Wu, Hang and Wang, May Dongmei and Ruan, Peifeng and Yang, Donghan and Wang, Tao and others},
  booktitle={The Second Workshop on GenAI for Health: Potential, Trust, and Policy Compliance},
  year={2025}
}

@misc{2023opencompass,
    title={OpenCompass: A Universal Evaluation Platform for Foundation Models},
    author={OpenCompass Contributors},
    howpublished = {\url{https://github.com/open-compass/opencompass}},
    year={2023}
}

@article{yao2024medqa,
  title={Medqa-cs: Benchmarking large language models clinical skills using an ai-sce framework},
  author={Yao, Zonghai and Zhang, Zihao and Tang, Chaolong and Bian, Xingyu and Zhao, Youxia and Yang, Zhichao and Wang, Junda and Zhou, Huixue and Jang, Won Seok and Ouyang, Feiyun and others},
  journal={arXiv preprint arXiv:2410.01553},
  year={2024}
}

@article{zhou2024medversa,
  title={MedVersa: a generalist foundation model for medical image interpretation},
  author={Zhou, Hong-Yu and Acosta, Juli{\'a}n Nicol{\'a}s and Adithan, Subathra and Datta, Suvrankar and Topol, Eric J and Rajpurkar, Pranav},
  journal={arXiv preprint arXiv:2405.07988},
  year={2024}
}

@article{medrax,
  title={Medrax: Medical reasoning agent for chest x-ray},
  author={Fallahpour, Adibvafa and Ma, Jun and Munim, Alif and Lyu, Hongwei and Wang, Bo},
  journal={arXiv preprint arXiv:2502.02673},
  year={2025}
}

@inproceedings{sun2022lesion,
  title={Lesion guided explainable few weak-shot medical report generation},
  author={Sun, Jinghan and Wei, Dong and Wang, Liansheng and Zheng, Yefeng},
  booktitle={International Conference on Medical Image Computing and Computer-Assisted Intervention},
  pages={615--625},
  year={2022},
  organization={Springer}
}

@article{gpt5_system_card,
  title={OpenAI GPT-5 System Card},
  author={Singh, Aaditya and Fry, Adam and Perelman, Adam and Tart, Adam and Ganesh, Adi and El-Kishky, Ahmed and McLaughlin, Aidan and Low, Aiden and Ostrow, AJ and Ananthram, Akhila and others},
  journal={arXiv preprint arXiv:2601.03267},
  year={2026}
}

@misc{openai2025gptoss120bgptoss20bmodel,
      title={gpt-oss-120b and gpt-oss-20b Model Card}, 
      author={OpenAI and : and Sandhini Agarwal and Lama Ahmad and Jason Ai and Sam Altman and Andy Applebaum and Edwin Arbus and Rahul K. Arora and Yu Bai and Bowen Baker and Haiming Bao and Boaz Barak and Ally Bennett and Tyler Bertao and Nivedita Brett and Eugene Brevdo and Greg Brockman and Sebastien Bubeck and Che Chang and Kai Chen and Mark Chen and Enoch Cheung and Aidan Clark and Dan Cook and Marat Dukhan and Casey Dvorak and Kevin Fives and Vlad Fomenko and Timur Garipov and Kristian Georgiev and Mia Glaese and Tarun Gogineni and Adam Goucher and Lukas Gross and Katia Gil Guzman and John Hallman and Jackie Hehir and Johannes Heidecke and Alec Helyar and Haitang Hu and Romain Huet and Jacob Huh and Saachi Jain and Zach Johnson and Chris Koch and Irina Kofman and Dominik Kundel and Jason Kwon and Volodymyr Kyrylov and Elaine Ya Le and Guillaume Leclerc and James Park Lennon and Scott Lessans and Mario Lezcano-Casado and Yuanzhi Li and Zhuohan Li and Ji Lin and Jordan Liss and Lily and Liu and Jiancheng Liu and Kevin Lu and Chris Lu and Zoran Martinovic and Lindsay McCallum and Josh McGrath and Scott McKinney and Aidan McLaughlin and Song Mei and Steve Mostovoy and Tong Mu and Gideon Myles and Alexander Neitz and Alex Nichol and Jakub Pachocki and Alex Paino and Dana Palmie and Ashley Pantuliano and Giambattista Parascandolo and Jongsoo Park and Leher Pathak and Carolina Paz and Ludovic Peran and Dmitry Pimenov and Michelle Pokrass and Elizabeth Proehl and Huida Qiu and Gaby Raila and Filippo Raso and Hongyu Ren and Kimmy Richardson and David Robinson and Bob Rotsted and Hadi Salman and Suvansh Sanjeev and Max Schwarzer and D. Sculley and Harshit Sikchi and Kendal Simon and Karan Singhal and Yang Song and Dane Stuckey and Zhiqing Sun and Philippe Tillet and Sam Toizer and Foivos Tsimpourlas and Nikhil Vyas and Eric Wallace and Xin Wang and Miles Wang and Olivia Watkins and Kevin Weil and Amy Wendling and Kevin Whinnery and Cedric Whitney and Hannah Wong and Lin Yang and Yu Yang and Michihiro Yasunaga and Kristen Ying and Wojciech Zaremba and Wenting Zhan and Cyril Zhang and Brian Zhang and Eddie Zhang and Shengjia Zhao},
      year={2025},
      eprint={2508.10925},
      archivePrefix={arXiv},
      primaryClass={cs.CL},
      url={https://arxiv.org/abs/2508.10925}, 
}

@article{cabral2024clinical,
  title={Clinical reasoning of a generative artificial intelligence model compared with physicians},
  author={Cabral, Stephanie and Restrepo, Daniel and Kanjee, Zahir and Wilson, Philip and Crowe, Byron and Abdulnour, Raja-Elie and Rodman, Adam},
  journal={JAMA internal medicine},
  volume={184},
  number={5},
  pages={581--583},
  year={2024},
  publisher={American Medical Association}
}

@article{chow2016review,
  title={Review of medical image quality assessment},
  author={Chow, Li Sze and Paramesran, Raveendran},
  journal={Biomedical signal processing and control},
  volume={27},
  pages={145--154},
  year={2016},
  publisher={Elsevier}
}

@article{shen2017deep,
  title={Deep learning in medical image analysis},
  author={Shen, Dinggang and Wu, Guorong and Suk, Heung-Il},
  journal={Annual review of biomedical engineering},
  volume={19},
  number={1},
  pages={221--248},
  year={2017},
  publisher={Annual Reviews}
}

@inproceedings{liu2021slake,
  title={Slake: A semantically-labeled knowledge-enhanced dataset for medical visual question answering},
  author={Liu, Bo and Zhan, Li-Ming and Xu, Li and Ma, Lin and Yang, Yan and Wu, Xiao-Ming},
  booktitle={2021 IEEE 18th international symposium on biomedical imaging (ISBI)},
  pages={1650--1654},
  year={2021},
  organization={IEEE}
}

@article{he2020pathvqa,
  title={Pathvqa: 30000+ questions for medical visual question answering},
  author={He, Xuehai and Zhang, Yichen and Mou, Luntian and Xing, Eric and Xie, Pengtao},
  journal={arXiv preprint arXiv:2003.10286},
  year={2020}
}

@inproceedings{ben2019vqa,
  title={Vqa-med: Overview of the medical visual question answering task at imageclef 2019},
  author={Ben Abacha, Asma and Hasan, Sadid A and Datla, Vivek V and Demner-Fushman, Dina and M{\"u}ller, Henning},
  booktitle={Proceedings of CLEF (Conference and Labs of the Evaluation Forum) 2019 Working Notes},
  year={2019},
  organization={9-12 September 2019}
}

@article{jiang2025medagentbench,
  title={MedAgentBench: a virtual EHR environment to benchmark medical LLM agents},
  author={Jiang, Yixing and Black, Kameron C and Geng, Gloria and Park, Danny and Zou, James and Ng, Andrew Y and Chen, Jonathan H},
  journal={Nejm Ai},
  volume={2},
  number={9},
  pages={AIdbp2500144},
  year={2025},
  publisher={Massachusetts Medical Society}
}

@inproceedings{hu2024omnimedvqa,
  title={Omnimedvqa: A new large-scale comprehensive evaluation benchmark for medical lvlm},
  author={Hu, Yutao and Li, Tianbin and Lu, Quanfeng and Shao, Wenqi and He, Junjun and Qiao, Yu and Luo, Ping},
  booktitle={Proceedings of the IEEE/CVF Conference on Computer Vision and Pattern Recognition},
  pages={22170--22183},
  year={2024}
  }

@inproceedings{jin2019pubmedqa,
  title={Pubmedqa: A dataset for biomedical research question answering},
  author={Jin, Qiao and Dhingra, Bhuwan and Liu, Zhengping and Cohen, William and Lu, Xinghua},
  booktitle={Proceedings of the 2019 conference on empirical methods in natural language processing and the 9th international joint conference on natural language processing (EMNLP-IJCNLP)},
  pages={2567--2577},
  year={2019}
}

@article{jin2021disease,
 title={What disease does this patient have? a large-scale open domain question answering dataset from medical exams},
 author={Jin, Di and Pan, Eileen and Oufattole, Nassim and Weng, Wei-Hung and Fang, Hanyi and Szolovits, Peter},
 journal={Applied Sciences},
 volume={11},
 number={14},
 pages={6421},
 year={2021}
}

@inproceedings{yue2023mmmu,
            title={MMMU: A Massive Multi-discipline Multimodal Understanding and Reasoning Benchmark for Expert AGI},
            author={Xiang Yue and Yuansheng Ni and Kai Zhang and Tianyu Zheng and Ruoqi Liu and Ge Zhang and Samuel Stevens and Dongfu Jiang and Weiming Ren and Yuxuan Sun and Cong Wei and Botao Yu and Ruibin Yuan and Renliang Sun and Ming Yin and Boyuan Zheng and Zhenzhu Yang and Yibo Liu and Wenhao Huang and Huan Sun and Yu Su and Wenhu Chen},
            booktitle={Proceedings of CVPR},
            year={2024},
          }

@article{zhang2024pmc,
  title={Pmc-vqa: Visual instruction tuning for medical visual question answering, 2024},
  author={Zhang, Xiaoman and Wu, Chaoyi and Zhao, Ziheng and Lin, Weixiong and Zhang, Ya and Wang, Yanfeng and Xie, Weidi},
  journal={URL https://arxiv. org/abs/2305.10415},
  volume={40},
  year={2024}
}

@article{lau2018dataset,
  title={A dataset of clinically generated visual questions and answers about radiology images},
  author={Lau, Jason J and Gayen, Soumya and Ben Abacha, Asma and Demner-Fushman, Dina},
  journal={Scientific data},
  volume={5},
  number={1},
  pages={180251},
  year={2018},
  publisher={Nature Publishing Group}
}

@inproceedings{liu2024llava,
  title={Llava-plus: Learning to use tools for creating multimodal agents},
  author={Liu, Shilong and Cheng, Hao and Liu, Haotian and Zhang, Hao and Li, Feng and Ren, Tianhe and Zou, Xueyan and Yang, Jianwei and Su, Hang and Zhu, Jun and others},
  booktitle={European conference on computer vision},
  pages={126--142},
  year={2024},
  organization={Springer}
}

@article{xie2024large,
  title={Large multimodal agents: A survey},
  author={Xie, Junlin and Chen, Zhihong and Zhang, Ruifei and Wan, Xiang and Li, Guanbin},
  journal={arXiv preprint arXiv:2402.15116},
  year={2024}
}

@article{gemini,
  title={Gemini: a family of highly capable multimodal models},
  author={Team, Gemini and Anil, Rohan and Borgeaud, Sebastian and Wu, Yonghui and Alayrac, Jean-Baptiste and Yu, Jiahui and Soricut, Radu and Schalkwyk, Johan and Dai, Andrew M and Hauth, Anja and others},
  journal={arXiv preprint arXiv:2312.11805},
  year={2023}
}

@article{toolbench,
  title={Toolllm: Facilitating large language models to master 16000+ real-world apis},
  author={Qin, Yujia and Liang, Shihao and Ye, Yining and Zhu, Kunlun and Yan, Lan and Lu, Yaxi and Lin, Yankai and Cong, Xin and Tang, Xiangru and Qian, Bill and others},
  journal={arXiv preprint arXiv:2307.16789},
  year={2023}
}

@article{agentbench,
  title={Agentbench: Evaluating llms as agents},
  author={Liu, Xiao and Yu, Hao and Zhang, Hanchen and Xu, Yifan and Lei, Xuanyu and Lai, Hanyu and Gu, Yu and Ding, Hangliang and Men, Kaiwen and Yang, Kejuan and others},
  journal={arXiv preprint arXiv:2308.03688},
  year={2023}
}

@article{llava,
  title={Visual instruction tuning},
  author={Liu, Haotian and Li, Chunyuan and Wu, Qingyang and Lee, Yong Jae},
  journal={Advances in Neural Information Processing Systems},
  volume={36},
  year={2024}
}

@article{apibench,
  title={Gorilla: Large language model connected with massive apis},
  author={Patil, Shishir G and Zhang, Tianjun and Wang, Xin and Gonzalez, Joseph E},
  journal={arXiv preprint arXiv:2305.15334},
  year={2023}
}

@article{gpt4,
  title={Gpt-4 technical report},
  author={Achiam, Josh and Adler, Steven and Agarwal, Sandhini and Ahmad, Lama and Akkaya, Ilge and Aleman, Florencia Leoni and Almeida, Diogo and Altenschmidt, Janko and Altman, Sam and Anadkat, Shyamal and others},
  journal={arXiv preprint arXiv:2303.08774},
  year={2023}
}

@inproceedings{react,
  title = {{ReAct}: Synergizing Reasoning and Acting in Language Models},
  author = {Yao, Shunyu and Zhao, Jeffrey and Yu, Dian and Du, Nan and Shafran, Izhak and Narasimhan, Karthik and Cao, Yuan},
  booktitle = {International Conference on Learning Representations},
  year = {2023},
  html = {https://arxiv.org/abs/2210.03629},
}

@article{claude3,
  title={The claude 3 model family: Opus, sonnet, haiku},
  author={Anthropic, AI},
  journal={Claude-3 Model Card},
  year={2024}
}

@article{llama3,
  title={Introducing Meta Llama 3: The most capable openly available LLM to date},
  author={Meta, AI},
  journal={Meta AI Blog (accessed 2024--04--20). There is no corresponding record for this reference},
  year={2024}
}

@article{qwen,
  title={Qwen technical report},
  author={Bai, Jinze and Bai, Shuai and Chu, Yunfei and Cui, Zeyu and Dang, Kai and Deng, Xiaodong and Fan, Yang and Ge, Wenbin and Han, Yu and Huang, Fei and others},
  journal={arXiv preprint arXiv:2309.16609},
  year={2023}
}

@article{mistral,
  title={Mistral 7B},
  author={Jiang, Albert Q and Sablayrolles, Alexandre and Mensch, Arthur and Bamford, Chris and Chaplot, Devendra Singh and Casas, Diego de las and Bressand, Florian and Lengyel, Gianna and Lample, Guillaume and Saulnier, Lucile and others},
  journal={arXiv preprint arXiv:2310.06825},
  year={2023}
}

@article{yi,
  title={Yi: Open foundation models by 01. ai},
  author={Young, Alex and Chen, Bei and Li, Chao and Huang, Chengen and Zhang, Ge and Zhang, Guanwei and Li, Heng and Zhu, Jiangcheng and Chen, Jianqun and Chang, Jing and others},
  journal={arXiv preprint arXiv:2403.04652},
  year={2024}
}

@article{deepseek,
  title={Deepseek llm: Scaling open-source language models with longtermism},
  author={Bi, Xiao and Chen, Deli and Chen, Guanting and Chen, Shanhuang and Dai, Damai and Deng, Chengqi and Ding, Honghui and Dong, Kai and Du, Qiushi and Fu, Zhe and others},
  journal={arXiv preprint arXiv:2401.02954},
  year={2024}
}

@article{deepseek-vl,
  title={DeepSeek-VL: towards real-world vision-language understanding},
  author={Lu, Haoyu and Liu, Wen and Zhang, Bo and Wang, Bingxuan and Dong, Kai and Liu, Bo and Sun, Jingxiang and Ren, Tongzheng and Li, Zhuoshu and Sun, Yaofeng and others},
  journal={arXiv preprint arXiv:2403.05525},
  year={2024}
}

@inproceedings{gaia,
  title={GAIA: a benchmark for General AI Assistants},
  author={Mialon, Gr{\'e}goire and Fourrier, Cl{\'e}mentine and Wolf, Thomas and LeCun, Yann and Scialom, Thomas},
  booktitle={The Twelfth International Conference on Learning Representations},
  year={2023}
}

@inproceedings{apibank,
  title={A comprehensive benchmark for tool-augmented LLMs},
  author={Li, Minghao and Zhao, Yingxiu and Yu, Bowen and Song, Feifan and Li, Hangyu and Yu, Haiyang and Li, Zhoujun and Huang, Fei and API-bank, Yongbin Li},
  booktitle={Proceedings of the 2023 Conference on Empirical Methods in Natural Language Processing},
  pages={3102--3116},
  year={2023}
}

@misc{langchain,
author = {Chase, Harrison},
month = oct,
title = {LangChain},
url = {https://github.com/langchain-ai/langchain},
year = {2022}
}

@misc{autogpt,
author = {Significant Gravitas},
license = {MIT},
title = {AutoGPT},
url = {https://github.com/Significant-Gravitas/AutoGPT},
year = {2023}
}

@misc{babyagi,
author = {Yohei Nakajima},
license = {MIT},
title = {BabyAGI},
url = {https://github.com/yoheinakajima/babyagi},
year = {2023}
}

@article{restgpt,
  title={Restgpt: Connecting large language models with real-world applications via restful apis},
  author={Song, Yifan and Xiong, Weimin and Zhu, Dawei and Li, Cheng and Wang, Ke and Tian, Ye and Li, Sujian},
  journal={arXiv preprint arXiv:2306.06624},
  year={2023}
}

@misc{datacopilot,
      title={Data-Copilot: Bridging Billions of Data and Humans with Autonomous Workflow}, 
      author={Wenqi Zhang and Yongliang Shen and Weiming Lu and Yueting Zhuang},
      year={2024},
      eprint={2306.07209},
      archivePrefix={arXiv},
      primaryClass={cs.CL}
}

@misc{appagent,
      title={AppAgent: Multimodal Agents as Smartphone Users}, 
      author={Chi Zhang and Zhao Yang and Jiaxuan Liu and Yucheng Han and Xin Chen and Zebiao Huang and Bin Fu and Gang Yu},
      year={2023},
      eprint={2312.13771},
      archivePrefix={arXiv},
      primaryClass={cs.CV}
}

@misc{lagent,
    title={{Lagent: InternLM} a lightweight open-source framework that allows users to efficiently build large language model(LLM)-based agents},
    author={Lagent Developer Team},
    howpublished = {\url{https://github.com/InternLM/lagent}},
    year={2023}
}

@article{mllm,
  title={Mllm-tool: A multimodal large language model for tool agent learning},
  author={Wang, C and Luo, W and Chen, Q and Mai, H and Guo, J and Dong, S and Xuan, XM and Li, Z and Ma, L and Gao, S},
  journal={arXiv preprint arXiv:2401.10727},
  volume={4},
  year={2024}
}

@inproceedings{webcpm,
  title={WebCPM: Interactive Web Search for Chinese Long-form Question Answering},
  author={Qin, Yujia and Cai, Zihan and Jin, Dian and Yan, Lan and Liang, Shihao and Zhu, Kunlun and Lin, Yankai and Han, Xu and Ding, Ning and Wang, Huadong and others},
  booktitle={Proceedings of the 61st Annual Meeting of the Association for Computational Linguistics (Volume 1: Long Papers)},
  pages={8968--8988},
  year={2023}
}

@article{webshop,
  title={Webshop: Towards scalable real-world web interaction with grounded language agents},
  author={Yao, Shunyu and Chen, Howard and Yang, John and Narasimhan, Karthik},
  journal={Advances in Neural Information Processing Systems},
  volume={35},
  pages={20744--20757},
  year={2022}
}

@article{hugginggpt,
  title={Hugginggpt: Solving ai tasks with chatgpt and its friends in hugging face},
  author={Shen, Yongliang and Song, Kaitao and Tan, Xu and Li, Dongsheng and Lu, Weiming and Zhuang, Yueting},
  journal={Advances in Neural Information Processing Systems},
  volume={36},
  year={2024}
}

@inproceedings{msagent,
  title={ModelScope-Agent: Building Your Customizable Agent System with Open-source Large Language Models},
  author={Li, Chenliang and Chen, He and Yan, Ming and Shen, Weizhou and Xu, Haiyang and Wu, Zhikai and Zhang, Zhicheng and Zhou, Wenmeng and Chen, Yingda and Cheng, Chen and others},
  booktitle={Proceedings of the 2023 Conference on Empirical Methods in Natural Language Processing: System Demonstrations},
  pages={566--578},
  year={2023}
}

@misc{agentlego,
    title={Enhance LLM agents with versatile tool APIs},
    author={AgentLego Developer Team},
    howpublished = {\url{https://github.com/InternLM/agentlego}},
    year={2023}
}

@article{osworld,
  title={Osworld: Benchmarking multimodal agents for open-ended tasks in real computer environments},
  author={Xie, Tianbao and Zhang, Danyang and Chen, Jixuan and Li, Xiaochuan and Zhao, Siheng and Cao, Ruisheng and Hua, Toh Jing and Cheng, Zhoujun and Shin, Dongchan and Lei, Fangyu and others},
  journal={arXiv preprint arXiv:2404.07972},
  year={2024}
}

@inproceedings{GTA,
 author = {Wang, Jize and Ma, Zerun and Li, Yining and Zhang, Songyang and Chen, Cailian and Chen, Kai and Le, Xinyi},
 booktitle = {Advances in Neural Information Processing Systems},
 doi = {10.52202/079017-2412},
 editor = {A. Globerson and L. Mackey and D. Belgrave and A. Fan and U. Paquet and J. Tomczak and C. Zhang},
 pages = {75749--75790},
 publisher = {Curran Associates, Inc.},
 title = {GTA: A Benchmark for General Tool Agents},
 url = {https://proceedings.neurips.cc/paper_files/paper/2024/file/8a75ee6d4b2eb0b777f549a32a5a5c28-Paper-Datasets_and_Benchmarks_Track.pdf},
 volume = {37},
 year = {2024}
}

@article{ashraf2025agent,
  title={Agent-x: Evaluating deep multimodal reasoning in vision-centric agentic tasks},
  author={Ashraf, Tajamul and Saqib, Amal and Ghani, Hanan and AlMahri, Muhra and Li, Yuhao and Ahsan, Noor and Nawaz, Umair and Lahoud, Jean and Cholakkal, Hisham and Shah, Mubarak and others},
  journal={arXiv preprint arXiv:2505.24876},
  year={2025}
}

@article{duncan2000medical,
  title   = {Medical Image Analysis: Progress over Two Decades and the Challenges Ahead},
  author  = {Duncan, James S. and Ayache, Nicholas},
  journal = {IEEE Transactions on Pattern Analysis and Machine Intelligence},
  volume  = {22},
  number  = {1},
  pages   = {85--106},
  year    = {2000},
  doi     = {10.1109/34.824822}
}

@article{xu2025comprehensive,
  title   = {A Comprehensive Survey of AI Agents in Healthcare},
  author  = {Xu, Gelei and Li, Xueyang and Chen, Yixiong and Duan, Yuying and Wu, Shuqing and Yu, Alexander and Chiu, Ching-Hao and Ni, Juntong and Tang, Ningzhi and Li, Toby Jia-Jun and others},
  journal = {TechRxiv},
  year    = {2025},
  doi     = {10.36227/techrxiv.176240542.22279040/v1},
  url     = {https://www.techrxiv.org/doi/full/10.36227/techrxiv.176240542.22279040/v1}
}

@misc{schmidgall2024agentclinic,
  title         = {AgentClinic: A Multimodal Agent Benchmark to Evaluate AI in Simulated Clinical Environments},
  author        = {Schmidgall, Samuel and Ziaei, Rojin and Harris, Carl and Reis, Eduardo and Jopling, Jeffrey and Moor, Michael},
  year          = {2024},
  eprint        = {2405.07960},
  archivePrefix = {arXiv},
  primaryClass  = {cs.HC},
  doi           = {10.48550/arXiv.2405.07960},
  url           = {https://arxiv.org/abs/2405.07960}
}

@article{jiang2025virtualehr,
  title   = {A Virtual {EHR} Environment to Benchmark Medical {LLM} Agents},
  author  = {Jiang, Yixing and Black, Kameron C. and Geng, Gloria and Park, Danny and Zou, James and Ng, Andrew Y. and Chen, Jonathan H.},
  journal = {NEJM AI},
  year    = {2025},
  doi     = {10.1056/AIdbp2500144},
  url     = {https://ai.nejm.org/doi/full/10.1056/AIdbp2500144}
}

@misc{li2024mediq,
  title         = {MediQ: Question-Asking {LLM}s and a Benchmark for Reliable Interactive Clinical Reasoning},
  author        = {Li, Shuyue Stella and Balachandran, Vidhisha and Feng, Shangbin and Ilgen, Jonathan S. and Pierson, Emma and Koh, Pang Wei and Tsvetkov, Yulia},
  year          = {2024},
  eprint        = {2406.00922},
  archivePrefix = {arXiv},
  primaryClass  = {cs.CL},
  doi           = {10.48550/arXiv.2406.00922},
  url           = {https://arxiv.org/abs/2406.00922}
}

@misc{ma2024clibench,
  title         = {CliBench: A Multifaceted and Multigranular Evaluation of Large Language Models in Clinical Decisions on Diagnoses, Procedures, Lab Tests Orders and Prescriptions},
  author        = {Ma, Mingyu Derek and Ye, Chenchen and others},
  year          = {2024},
  eprint        = {2406.09923},
  archivePrefix = {arXiv},
  primaryClass  = {cs.CL},
  doi           = {10.48550/arXiv.2406.09923},
  url           = {https://arxiv.org/abs/2406.09923}
}

@inproceedings{fan2025aihospital,
  title     = {AI Hospital: Benchmarking Large Language Models in a Multi-agent Medical Interaction Simulator},
  author    = {Fan, Zhihao and Wei, Lai and Tang, Jialong and Chen, Wei and Wang, Siyuan and Wei, Zhongyu and Xie, Jun and Huang, Fei and Zhou, Jingren},
  booktitle = {Proceedings of the 31st International Conference on Computational Linguistics},
  pages     = {10183--10213},
  year      = {2025},
  month     = jan,
  publisher = {Association for Computational Linguistics},
  url       = {https://aclanthology.org/2025.coling-main.680/}
}

@inproceedings{li2024mmedagent,
  title={Mmedagent: Learning to use medical tools with multi-modal agent},
  author={Li, Binxu and Yan, Tiankai and Pan, Yuanting and Luo, Jie and Ji, Ruiyang and Ding, Jiayuan and Xu, Zhe and Liu, Shilong and Dong, Haoyu and Lin, Zihao and others},
  booktitle={Findings of the Association for Computational Linguistics: EMNLP 2024},
  pages={8745--8760},
  year={2024}
}

@misc{nakano2021webgpt,
  title         = {WebGPT: Browser-assisted question-answering with human feedback},
  author        = {Nakano, Reiichiro and Hilton, Jacob and Balaji, Suchir and Wu, Jeff and Ouyang, Long and Kim, Christina and Hesse, Christopher and Jain, Shantanu and Kosaraju, Vineet and Saunders, William and Jiang, Xu and Cobbe, Karl and Eloundou, Tyna and Krueger, Gretchen and Button, Kevin and Knight, Matthew and Chess, Benjamin and Schulman, John},
  year          = {2021},
  eprint        = {2112.09332},
  archivePrefix = {arXiv},
  primaryClass  = {cs.CL},
  doi           = {10.48550/arXiv.2112.09332},
  url           = {https://arxiv.org/abs/2112.09332}
}

@inproceedings{suris2023vipergpt,
  title     = {ViperGPT: Visual Inference via Python Execution for Reasoning},
  author    = {Sur{\'i}s, D{\'i}dac and Menon, Sachit and Vondrick, Carl},
  booktitle = {Proceedings of the IEEE/CVF International Conference on Computer Vision (ICCV)},
  year      = {2023},
  url       = {https://openaccess.thecvf.com/content/ICCV2023/papers/Suris_ViperGPT_Visual_Inference_via_Python_Execution_for_Reasoning_ICCV_2023_paper.pdf}
}

@misc{jimenez2023swebench,
  title         = {SWE-bench: Can Language Models Resolve Real-World {G}it{H}ub Issues?},
  author        = {Jimenez, Carlos E. and Yang, John and Wettig, Alexander and Yao, Shunyu and Pei, Kexin and Press, Ofir and Narasimhan, Karthik},
  year          = {2023},
  eprint        = {2310.06770},
  archivePrefix = {arXiv},
  primaryClass  = {cs.CL},
  doi           = {10.48550/arXiv.2310.06770},
  url           = {https://arxiv.org/abs/2310.06770}
}

@inproceedings{sun2025docagent,
  title     = {DocAgent: An Agentic Framework for Multi-Modal Long-Context Document Understanding},
  author    = {Sun, Lin and others},
  booktitle = {Proceedings of the 2025 Conference on Empirical Methods in Natural Language Processing (EMNLP)},
  year      = {2025},
  url       = {https://aclanthology.org/2025.emnlp-main.893/}
}

@misc{lu2023chameleon,
  title         = {Chameleon: Plug-and-Play Compositional Reasoning with Large Language Models},
  author        = {Lu, Pan and Peng, Baolin and Cheng, Hao and Galley, Michel and Chang, Kai-Wei and Wu, Ying Nian and Zhu, Song-Chun and Gao, Jianfeng},
  year          = {2023},
  eprint        = {2304.09842},
  archivePrefix = {arXiv},
  primaryClass  = {cs.CL},
  doi           = {10.48550/arXiv.2304.09842},
  url           = {https://arxiv.org/abs/2304.09842}
}

@misc{yang2024sweagent,
  title         = {SWE-agent: Agent-Computer Interfaces Enable Automated Software Engineering},
  author        = {Yang, John and Jimenez, Carlos E. and Wettig, Alexander and Lieret, Kilian and Yao, Shunyu and Narasimhan, Karthik and Press, Ofir},
  year          = {2024},
  eprint        = {2405.15793},
  archivePrefix = {arXiv},
  primaryClass  = {cs.SE},
  doi           = {10.48550/arXiv.2405.15793},
  url           = {https://arxiv.org/abs/2405.15793}
}

@misc{wu2023autogen,
  title         = {AutoGen: Enabling Next-Gen LLM Applications via Multi-Agent Conversation Framework},
  author        = {Wu, Qingyun and Bansal, Gagan and Zhang, Jieyu and Wu, Yiran and Zhang, Shaokun and Zhu, Erkang and Li, Beibin and Jiang, Li and Zhang, Xiaoyun and Wang, Chi},
  year          = {2023},
  eprint        = {2308.08155},
  archivePrefix = {arXiv},
  primaryClass  = {cs.AI},
  doi           = {10.48550/arXiv.2308.08155},
  url           = {https://arxiv.org/abs/2308.08155}
}

@misc{biovil,
  title         = {Making the Most of Text Semantics to Improve Biomedical Vision--Language Processing},
  author        = {Boecking, Benedikt and Usuyama, Naoto and Bannur, Shruthi and Castro, Daniel C. and Schwaighofer, Anton and Hyland, Stephanie L. and Wetscherek, Maria T. and Naumann, Tristan and Nori, Harsha and Ahuja, Neeraj and others},
  year          = {2022},
  eprint        = {2204.09817},
  archivePrefix = {arXiv},
  primaryClass  = {cs.CV},
  doi           = {10.48550/arXiv.2204.09817},
  url           = {https://arxiv.org/abs/2204.09817}
}

@misc{medclip,
  title         = {MedCLIP: Contrastive Learning from Unpaired Medical Images and Text},
  author        = {Wang, Zifeng and Wu, Zhenbang and Agarwal, Dinesh and Sun, Jimeng},
  year          = {2022},
  eprint        = {2210.10163},
  archivePrefix = {arXiv},
  primaryClass  = {cs.CV},
  doi           = {10.48550/arXiv.2210.10163},
  url           = {https://arxiv.org/abs/2210.10163}
}

@misc{llava-med,
  title         = {LLaVA-Med: Training a Large Language-and-Vision Assistant for Biomedicine in One Day},
  author        = {Li, Chunyuan and Wong, Cliff and Zhang, Sheng and Usuyama, Naoto and Liu, Haotian and Yang, Jianwei and Naumann, Tristan and Poon, Hoifung and Gao, Jianfeng},
  year          = {2023},
  eprint        = {2306.00890},
  archivePrefix = {arXiv},
  primaryClass  = {cs.CV},
  doi           = {10.48550/arXiv.2306.00890},
  url           = {https://arxiv.org/abs/2306.00890}
}

@inproceedings{medflamingo,
  title     = {Med-Flamingo: a Multimodal Medical Few-shot Learner},
  author    = {Moor, Michael and Huang, Shraey B. and Wu, Yadong and Kapur, T. and others},
  booktitle = {Proceedings of the 40th International Conference on Machine Learning (ICML)},
  series    = {Proceedings of Machine Learning Research},
  volume    = {225},
  year      = {2023},
  url       = {https://proceedings.mlr.press/v225/moor23a.html}
}

@misc{radfm,
  title         = {Towards Generalist Foundation Model for Radiology by Leveraging Web-scale 2D\&3D Medical Data},
  author        = {Wu, Chaoyi and Zhang, Xiaoman and Zhang, Ya and Wang, Yanfeng and Xie, Weidi},
  year          = {2023},
  eprint        = {2308.02463},
  archivePrefix = {arXiv},
  primaryClass  = {cs.CV},
  doi           = {10.48550/arXiv.2308.02463},
  url           = {https://arxiv.org/abs/2308.02463}
}

@article{sellergren2025medgemma,
  title={MedGemma Technical Report},
  author={Sellergren, Andrew and Kazemzadeh, Sahar and Jaroensri, Tiam and Kiraly, Atilla and Traverse, Madeleine and Kohlberger, Timo and Xu, Shawn and Jamil, Fayaz and Hughes, Cían and Lau, Charles and others},
  journal={arXiv preprint arXiv:2507.05201},
  year={2025}
}

@article{shu2025fleming,
  title={Fleming-VL: Towards Universal Medical Visual Reasoning with Multimodal LLMs},
  author={Shu, Yan and Liu, Chi and Chen, Robin and Li, Derek and Dai, Bryan},
  journal={arXiv preprint arXiv:2511.00916},
  year={2025}
}

@article{xu2025lingshu,
  title={Lingshu: A generalist foundation model for unified multimodal medical understanding and reasoning},
  author={Xu, Weiwen and Chan, Hou Pong and Li, Long and Aljunied, Mahani and Yuan, Ruifeng and Wang, Jianyu and Xiao, Chenghao and Chen, Guizhen and Liu, Chaoqun and Li, Zhaodonghui and others},
  journal={arXiv preprint arXiv:2506.07044},
  year={2025}
}

@article{deria2026medmo,
  title={MedMO: Grounding and Understanding Multimodal Large Language Model for Medical Images},
  author={Deria, Ankan and Kumar, Komal and Dukre, Adinath Madhavrao and Segal, Eran and Khan, Salman and Razzak, Imran},
  journal={arXiv preprint arXiv:2602.06965},
  year={2026}
}

@article{kim2024mdagents,
  title={Mdagents: An adaptive collaboration of llms for medical decision-making},
  author={Kim, Yubin and Park, Chanwoo and Jeong, Hyewon and Chan, Yik S and Xu, Xuhai and McDuff, Daniel and Lee, Hyeonhoon and Ghassemi, Marzyeh and Breazeal, Cynthia and Park, Hae W},
  journal={Advances in Neural Information Processing Systems},
  volume={37},
  pages={79410--79452},
  year={2024}
}

@article{kim2025tiered,
  title={Tiered Agentic Oversight: A Hierarchical Multi-Agent System for Healthcare Safety},
  author={Kim, Yubin and Jeong, Hyewon and Park, Chanwoo and Park, Eugene and Zhang, Haipeng and Liu, Xin and Lee, Hyeonhoon and McDuff, Daniel and Ghassemi, Marzyeh and Breazeal, Cynthia and others},
  journal={arXiv preprint arXiv:2506.12482},
  year={2025}
}

@inproceedings{kim2025behaviorsft,
  title={BehaviorSFT: Behavioral Token Conditioning for Health Agents Across the Proactivity Spectrum},
  author={Kim, Yubin and Hu, Zhiyuan and Jeong, Hyewon and Park, Eugene W and Li, Shuyue Stella and Park, Chanwoo and Xiong, Shiyun and Lu, MingYu and Lee, Hyeonhoon and Liu, Xin and others},
  booktitle={Findings of the Association for Computational Linguistics: EMNLP 2025},
  pages={9789--9817},
  year={2025}
}

%%%%%%%%%%%%%%%%%%%%%%%%%%%%%%%%%%%%%%%%%%%%%%%%%%%%%%%%%%%%

% \newpage
% \input{checklist.tex}

\newpage

\newpage
\appendix
\begin{center}
\Large\textbf{Appendix for \ours: A benchmark for Clinical Tool Agents}
\end{center}

\vspace{1em}

\begin{table}[h]
\centering
\renewcommand{\arraystretch}{1.3}
\label{tab:appendix_contents}
\begin{tabular}{@{}p{4cm} p{9.5cm}@{}}
\toprule
\texttt{Section~\ref{datacard}} & \textbf{Datacard for MedCTA Benchmark} \\
\texttt{Section~\ref{tool_details}} & \textbf{Executable Tool Library} \\
\texttt{Section~\ref{app:clinician_verification}} & \textbf{Inter Annotator Agreement} \\
\texttt{Section~\ref{app: mcta_instruction}} & \textbf{Instruction for Annotators} \\
\texttt{Section~\ref{mcta_pipeline}} & \textbf{Stepwise Construction of Executable Clinical Tool Trajectories} \\
\texttt{Section~\ref{mcta_prompts}} & \textbf{Prompting Templates for Agent Trajectory Construction} \\
\texttt{Section~\ref{app:metrics_details}} & \textbf{Metrics and Implementation Details} \\

\texttt{Section~\ref{app:diagnostics}} & \textbf{Additional Experimental Results and Diagnostic Analyses} \\

\texttt{Section~\ref{mcta_examples}} & \textbf{MedCTA Task Examples} \\
\texttt{Section~\ref{app:judge_prompts}} & \textbf{LLM-as-Judge Prompts} \\

\bottomrule
\end{tabular}
\end{table}

\section{Datacard for MedCTA}
\label{datacard}

\subsection{Motivation}

\begin{itemize}

\item \textbf{For what purpose was the dataset created?}

MedCTA is designed to evaluate the \emph{multi-step clinical reasoning and tool-orchestration capabilities} of multimodal medical agents in realistic healthcare scenarios. Unlike traditional medical VQA benchmarks, it features clinician-validated, step-implicit clinical queries that require structured reasoning and implicit tool use. The benchmark integrates executable tools spanning perception, operation, clinical reasoning, and reporting, grounded in authentic multimodal clinical inputs. MedCTA bridges the gap between static medical QA and real-world tool-mediated clinical workflows.

\item \textbf{Who created the dataset?}

The dataset was created by the authors of this paper in collaboration with clinician annotators and technical researchers.

\item \textbf{Who funded the creation of the dataset?}

Funding details will be disclosed after the review process.

\end{itemize}

\subsection{Composition}

\begin{itemize}

\item \textbf{What do the instances represent?}

Each instance represents a clinically grounded task stored in JSON format. It includes a natural-language clinical query, multimodal medical inputs (e.g., CT, MRI, X-ray, pathology images, reports), a set of executable tools, and a reference tool trajectory. The trajectory consists of step-wise tool calls with arguments and outputs, ending with a clinically validated final answer.

\item \textbf{How many instances are there?}

MedCTA contains \textbf{107 clinician-validated tasks}, each grounded in real medical data and paired with executable tool trajectories.

\item \textbf{Is the dataset a sample or complete?}

The dataset is a purpose-built benchmark and contains all curated and validated instances constructed for MedCTA.

\item \textbf{What data does each instance include?}

Each instance includes a clinical query, multimodal inputs, tool descriptions, a reference execution trajectory, and a final validated answer.

\item \textbf{Is there a label or target?}

Yes. Each instance includes a reference trajectory and final clinical outcome. For ambiguous tasks, multiple acceptable answers may be provided.

\item \textbf{Is any information missing?}

No.

\item \textbf{Are relationships between instances explicit?}

No.

\item \textbf{Are there recommended splits?}

MedCTA is designed as an evaluation benchmark and does not define training/validation splits.

\item \textbf{Are there errors or noise?}

The dataset is constructed through a semi-automated pipeline and validated by clinicians. Minor annotation errors may still exist.

\item \textbf{Is the dataset self-contained?}

Yes. While images originate from public datasets, all queries, trajectories, and annotations are newly constructed and included.

\item \textbf{Does it contain confidential data?}

No. All data is publicly sourced and de-identified.

\item \textbf{Does it contain harmful content?}

No.

\end{itemize}

\subsection{Collection Process}

\begin{itemize}

\item \textbf{How was the data acquired?}

Clinical queries and draft trajectories were generated using large multimodal models (e.g., GPT-4o) and refined through human annotation and clinician validation. Medical images were sourced from publicly available academic datasets.

\item \textbf{What procedures were used?}

A semi-automated pipeline was used: LLM-based drafting, technical trajectory construction, and clinician validation. Tool executions were logged and verified using structured interfaces.

\item \textbf{Who was involved and how were they compensated?}

The dataset was constructed by researchers, graduate students, and clinician collaborators.

\item \textbf{Timeframe?}

Data collection and curation were performed during \textbf{2025–2026}.

\item \textbf{Ethical review?}

All inputs come from public datasets approved for research use. No identifiable patient data is included.

\end{itemize}

\subsection{Preprocessing / Cleaning / Labeling}

\begin{itemize}

\item \textbf{Was preprocessing done?}

Yes. Queries and trajectories were refined through technical verification and clinician validation. Redundant or unstable steps were removed, and tool arguments were standardized.

\item \textbf{Was raw data preserved?}

There is no separate raw trajectory dataset; trajectories are purpose-built. Medical images originate from existing datasets.

\item \textbf{Is preprocessing software available?}

Annotation and logging were performed using internal tools and standard research environments.

\end{itemize}

\subsection{Uses}

\begin{itemize}

\item \textbf{Has the dataset been used?}

No prior use beyond this work.

\item \textbf{Is there a repository?}

An anonymized repository is released with the submission

\item \textbf{Other potential uses?}

MedCTA can be used for evaluating multimodal reasoning, tool-augmented agents, trajectory-level evaluation, and safe clinical decision-support research.

\item \textbf{Any limitations affecting use?}

Evaluation assumes access to compatible tool interfaces and structured reasoning outputs.

\item \textbf{Potential negative impacts?}

Misuse may lead to overestimating model readiness for clinical deployment. The dataset is intended strictly for research.

\end{itemize}

\subsection{Distribution}

\begin{itemize}

\item \textbf{Will it be shared?}

Yes, for academic research use.

\item \textbf{How?}

Via a public GitHub repository.

\item \textbf{License?}

Academic research license.

\item \textbf{IP restrictions?}

No.

\end{itemize}

\subsection{Maintenance}

\begin{itemize}

\item \textbf{Who maintains it?}

The authors.

\item \textbf{Contact?}

Via the corresponding author.

\item \textbf{Updates?}

Yes, including corrections, new tasks, and tool improvements.

\item \textbf{External contributions?}

Supported via GitHub issues and pull requests.

\end{itemize}

\clearpage
\section{Executable Tool Library.}
\label{tool_details}

MedCTA integrates a compact library of executable multimodal tools that enable structured, step-wise reasoning over clinical inputs. As summarized in Table~\ref{tab:tool_desc}, the toolset consists of five core components spanning perception, grounding, retrieval, and computation: \texttt{OCR}, \texttt{ImageDescription}, \texttt{RegionAttributeDescription}, \texttt{GoogleSearch}, and \texttt{Calculator}.

The perception module includes \texttt{OCR} for extracting textual content from medical images and documents, and \texttt{ImageDescription} for generating holistic semantic summaries of visual inputs. For fine-grained spatial reasoning, \texttt{RegionAttributeDescription} enables localized attribute extraction conditioned on specified image regions, supporting tasks such as lesion characterization or anatomical detail analysis. External knowledge retrieval is handled by \texttt{GoogleSearch}, which provides access to up-to-date medical and general-domain information when required for reasoning beyond the input context. Finally, \texttt{Calculator} supports symbolic and numerical computation, enabling quantitative reasoning (e.g., dosage calculation, measurement comparison).

Each tool is defined by a standardized input–output interface, allowing tool invocations to be executed deterministically and logged in structured JSON format. This explicit interface design facilitates fine-grained evaluation along multiple dimensions, including tool selection accuracy, argument validity, intermediate evidence consistency, and overall task correctness.

Importantly, within the MedCTA evaluation framework, tool usage is mediated through a Lagent~\cite{lagent}-based agent wrapper integrated with OpenCompass~\cite{2023opencompass}. Depending on the evaluation setting, tools are either (i) exposed as a static tool schema via \texttt{tool\_meta} for step-by-step evaluation without execution (\texttt{every\_with\_gt}), or (ii) accessed through a deployed tool server via \texttt{tool\_server} for end-to-end execution (\texttt{every}). While the reference tool chain $\pi$ specifies a sufficient set of tools to solve each task, the available subset $\mathcal{U} \subseteq \mathcal{D}$ and the invocation order remain hidden from the agent at test time. This design requires the agent to perform autonomous planning, implicit tool selection, and multi-step reasoning under partial observability.
\begin{table}[t]
\centering
\renewcommand{\arraystretch}{1.2}
\caption{Detailed Definitions of Tools in \ours}
\label{tab:tool_desc}
\resizebox{\textwidth}{!}{
\begin{tabular}{l|p{5cm}|p{4cm}|p{4cm}}
\toprule
\rowcolor{gray!20}
\textbf{Name} & \textbf{Description} & \textbf{Input} & \textbf{Output} \\
\midrule

\texttt{OCR} & Extracts all visible text from an image along with spatial grounding when applicable. 
& Image 
& Text (optionally with bounding boxes) \\

\texttt{ImageDescription} & Generates a holistic natural language description summarizing the visual content. 
& Image 
& Image caption / summary \\

\texttt{RegionAttributeDescription} & Produces fine-grained descriptions of specified attributes within a given image region. 
& Image, bounding box, attribute (optional) 
& Region-level attribute description \\

\texttt{GoogleSearch} & Retrieves relevant external knowledge for a given query using a web search interface. 
& Query, top-$k$ (optional) 
& Ranked search results \\

\texttt{Calculator} & Evaluates mathematical expressions for symbolic and numerical reasoning (restricted execution environment). 
& Text expression 
& Computed result \\

\bottomrule
\end{tabular}
}
\end{table}

\section{Clinician Verification Protocol and Inter-Annotator Agreement}
\label{app:clinician_verification}

This appendix describes the clinician verification protocol used to construct and validate \ours. Because the benchmark evaluates medical tool agents, correctness is not limited to the final answer: the intermediate reasoning trajectory must also be clinically plausible, grounded in the available evidence, and consistent with real diagnostic workflows. We therefore use clinician review to verify both outcome correctness and process-level validity.

\subsection{Benchmark sample structure}
Each benchmark instance contains:
\begin{itemize}[leftmargin=1.2em, itemsep=2pt, topsep=2pt]
    \item a de-identified medical input, such as a radiology image, pathology image, clinical figure, report, or multimodal medical file;
    \item a clinically realistic, step-implicit task query;
    \item a structured executable trajectory consisting of tool calls, arguments, observations, and intermediate reasoning;
    \item a final clinically validated answer, with multiple acceptable answers when appropriate.
\end{itemize}
The goal of this design is to evaluate whether a model can solve a clinical task through a plausible tool-mediated reasoning process, rather than merely guessing a correct final response.

% \subsection{Clinician reviewers}
% The benchmark was reviewed by qualified clinician annotators with formal medical training and relevant experience in interpreting clinical images and medical evidence. To preserve anonymity during review, we report reviewer background only in aggregate: \texttt{[N]} clinician reviewers from \texttt{[specialties]}, with \texttt{[X]}--\texttt{[Y]} years of clinical experience. Clinicians acted as expert verifiers rather than automatic data generators. Their role was to assess whether each task, trajectory, and final answer reflected clinically meaningful and evidence-grounded reasoning.

Clinician reviewers were blinded to model predictions during dataset verification. This prevents model outputs from influencing the gold trajectory or final answer annotations.

\subsection{Verification protocol}
Each benchmark instance was reviewed along four dimensions.

\paragraph{Image and input validity.}
Clinicians verified that the input was medically interpretable, that the modality and anatomical region were correctly specified, and that image quality was sufficient for the intended task. Inputs with severe ambiguity, insufficient quality, or misleading artifacts were corrected or excluded.

\paragraph{Task goal correctness.}
Clinicians assessed whether the task query was clinically meaningful, realistic, and answerable from the provided input and available tool library. Queries that exposed the tool sequence explicitly, contained unsupported assumptions, or asked clinically unstable questions were revised.

\paragraph{Trajectory plausibility.}
Clinicians examined whether the reference tool trajectory followed a plausible diagnostic workflow. This included verifying that the trajectory used appropriate anatomical landmarks, visual findings, OCR evidence, external knowledge when needed, and localized region descriptions where clinically relevant. Trajectories containing hallucinated observations, unsupported shortcuts, redundant steps, or non-clinical reasoning were corrected.

\paragraph{Final answer accuracy.}
The final answer was checked against the visible evidence, tool observations, and accepted clinical interpretation. For tasks with non-unique valid descriptions, clinicians allowed multiple acceptable reference answers as long as they preserved the same clinical meaning.

Samples failing any verification criterion were either corrected through adjudication or removed from the benchmark.

\subsection{Agreement protocol}
To quantify annotation reliability, a stratified subset of samples was independently reviewed by multiple clinicians. The subset was selected to cover different modalities, anatomical regions, tool types, and trajectory lengths. Each clinician independently judged:
\begin{itemize}[leftmargin=1.2em, itemsep=2pt, topsep=2pt]
    \item whether the medical input was valid and interpretable;
    \item whether the task goal was clinically realistic and answerable;
    \item whether the reference trajectory was clinically plausible;
    \item whether the final answer was clinically correct.
\end{itemize}
Disagreements were resolved only after computing agreement statistics, so the reported agreement reflects independent judgments rather than post-hoc consensus. In addition, we perform a separate clinician audit of model rollouts in Appendix~\ref{app:human_eval}, which evaluates whether the automated clinical-reasoning judge aligns with human clinical assessment on $F_{\mathrm{acc}}$, $C_s$, and $S_{\mathrm{comp}}$.
% \subsection{Agreement metrics}
% We report inter-annotator agreement using standard chance-corrected statistics. For pairwise clinician agreement, we use Cohen's $\kappa$. When more than two clinicians annotate the same subset, we use Fleiss' $\kappa$. Binary judgments are used for input validity, task validity, and trajectory plausibility. Final-answer correctness is treated as binary when the answer is judged correct/incorrect, and categorical when multiple diagnostic categories are present.

% \subsection{Adjudication}
% After independent annotation and agreement computation, disagreements were adjudicated through discussion between clinician reviewers and the benchmark authors. Adjudication focused on whether the task was clinically answerable, whether the trajectory used sufficient evidence, and whether the final answer was medically justified. When disagreement reflected genuine clinical ambiguity rather than annotation error, the item was either revised to include multiple acceptable answers or excluded.

\subsection{Ethical considerations}
All medical inputs used in \ours are de-identified and contain no protected health information. The benchmark is intended for research evaluation of medical AI systems and is not intended for clinical deployment. The dataset should not be used to make patient-level decisions. Any institutional or ethics review requirements were handled according to local policy for de-identified retrospective or publicly available medical data.

\subsection{Human Annotation Effort}
\label{app:annotation_effort}

The construction of \ours required approximately \textbf{321 human annotation hours}, combining technical annotation, trajectory construction, and clinician verification. This effort was distributed across three stages.

\paragraph{Sample selection and query drafting (72 hours).}
Annotators first selected medically interpretable samples from public clinical and medical VQA sources, filtered ambiguous or low-quality cases, and converted perception-level questions into clinically grounded, step-implicit task goals. This stage also included removing tool-leaking phrasing and drafting initial task descriptions suitable for agentic evaluation.

\paragraph{Tool-chain construction and technical annotation (164 hours).}
Annotators then constructed executable tool trajectories for each task. This involved inspecting the medical inputs, selecting the required tools, specifying tool arguments, executing tool calls, recording observations, and revising trajectories for schema validity, minimality, and stability. LMM-generated drafts were used only as initialization; humans corrected redundant steps, invalid arguments, brittle localization, and unsupported reasoning.

\paragraph{Clinical verification and finalization (85 hours).}
Clinician reviewers verified the medical correctness of the final queries, reference trajectories, tool outputs, and final answers. This stage ensured that each trajectory followed a plausible clinical workflow, that intermediate evidence supported the conclusion, and that ambiguous or clinically unstable samples were corrected or removed. Finalized samples were then checked for formatting consistency and JSON executability.

\subsection{Clinician Audit of Model Rollouts}
\label{app:human_eval}

\paragraph{Motivation.}
MedCTA uses automated rubric-based evaluation to scale clinical reasoning assessment over many model rollouts. Because the benchmark is intended to diagnose clinically grounded tool use rather than only final-answer matching, we additionally performed a clinician audit on a subset of model trajectories. The goal of this audit is to test whether the automated clinical-reasoning metrics reflect clinically meaningful judgments made by a human reviewer.

\paragraph{Protocol.}
We evaluated 36 benchmark tasks for four representative models: GPT-5.4, Claude-opus-4-6, Gemini-3-flash, and Qwen3.5-9B. This yields 144 model--task rollouts. For each rollout, the clinician assigned scores for the same three clinical-reasoning dimensions used in the main evaluation:
$F_{\mathrm{acc}}$ for clinical faithfulness, $C_s$ for context integration, and $S_{\mathrm{comp}}$ for semantic completeness. Each dimension was scored on the same ordinal rubric $\{0, 0.1, 0.4, 0.7, 1.0\}$, where higher values indicate stronger clinical validity, evidence use, or completeness. We report all scores on a $0$--$100$ scale for consistency with the main paper.

\paragraph{Human audit results.}
Table~\ref{tab:human_clinical_eval} summarizes the clinician scores. The audit confirms that the clinical-reasoning bottleneck remains substantial: the overall mean score across all 144 evaluated rollouts is only 28.5/100. GPT-5.4 obtains the highest human clinical-reasoning score, but still reaches only 37.9/100 on average. Semantic completeness is the weakest dimension overall (25.1/100), suggesting that models often omit required clinical findings even when parts of the reasoning chain are plausible.

\begin{table}[t]
\centering
\small
\caption{
Clinician audit of model rollouts on 36 MedCTA tasks. Scores are percentages. 
``Mean'' is the average of $F_{\mathrm{acc}}$, $C_s$, and $S_{\mathrm{comp}}$ per rollout. 
The bracketed interval is a normal-approximation 95\% confidence interval for the per-task mean score. 
The last two columns report the percentage of rollouts whose mean score is at least 0.4 or 0.7, respectively.
}
\label{tab:human_clinical_eval}
\resizebox{\linewidth}{!}{
\begin{tabular}{lccccccc}
\toprule
Model & $N$ & $F_{\mathrm{acc}}$ & $C_s$ & $S_{\mathrm{comp}}$ & Mean & Mean $\geq 0.4$ & Mean $\geq 0.7$ \\
\midrule
GPT-5.4 & 36 & 38.3 & 39.7 & 35.6 & 37.9 [29.2, 46.5] & 41.7 & 11.1 \\
Claude-opus-4-6 & 36 & 32.2 & 30.3 & 23.1 & 28.5 [21.6, 35.4] & 33.3 & 2.8 \\
Gemini-3-flash & 36 & 29.2 & 21.9 & 26.7 & 25.9 [17.0, 34.9] & 25.0 & 11.1 \\
Qwen3.5-9B & 36 & 23.6 & 26.1 & 15.0 & 21.6 [16.2, 27.0] & 11.1 & 2.8 \\
\midrule
Average & 144 & 30.8 & 29.5 & 25.1 & 28.5 & 27.8 & 6.9 \\
\bottomrule
\end{tabular}
}
\end{table}

\paragraph{Agreement with the automated clinical judge.}
We next compare the clinician audit with the automated LLM-as-judge scores reported in the main benchmark table for the same four model families. Since the uploaded audit contains human scores but not per-sample automated scores for the same 36 examples, we report model-level agreement rather than per-instance inter-rater agreement. Table~\ref{tab:human_auto_agreement} shows that the aggregate clinical score has moderate model-level alignment between human and automated evaluation (Pearson $r=0.62$, Spearman $\rho=0.60$). Agreement is strongest for clinical faithfulness, where the human and automated judge induce the same model ranking (Spearman $\rho=1.00$). Context integration and semantic completeness show weaker alignment, indicating that evidence-use and missing-finding judgments are more sensitive to judge calibration and subset composition.

\begin{table}[t]
\centering
\small
\caption{
Model-level agreement between the clinician audit and the automated clinical-reasoning judge. 
MAE is reported in percentage points.
}
\label{tab:human_auto_agreement}
\begin{tabular}{lccc}
\toprule
Dimension & Pearson $r$ & Spearman $\rho$ & MAE \\
\midrule
Clinical Faithfulness ($F_{\mathrm{acc}}$) & 0.97 & 1.00 & 17.5 \\
Context Integration ($C_s$) & 0.40 & 0.20 & 15.8 \\
Semantic Completeness ($S_{\mathrm{comp}}$) & 0.33 & 0.40 & 7.5 \\
Aggregate clinical score & 0.62 & 0.60 & 13.5 \\
\bottomrule
\end{tabular}
\end{table}

\paragraph{Consistency among human clinical dimensions.}
The three human-scored dimensions are related but not redundant. Across all 144 rollouts, the three-dimensional clinical rubric has Cronbach's $\alpha=0.79$, indicating good internal consistency. Pairwise correlations show that clinical faithfulness and semantic completeness are strongly related (Pearson $r=0.78$, Spearman $\rho=0.74$), while context integration is less tightly coupled with completeness (Pearson $r=0.39$, Spearman $\rho=0.34$). This supports the design choice of reporting separate clinical-reasoning dimensions: a rollout can follow a plausible workflow while still failing to use multimodal evidence, or it can use some evidence while omitting required clinical findings.

\begin{table}[htbp]
\centering
\small
\caption{
Internal consistency of the clinician-scored clinical-reasoning dimensions across 144 audited rollouts.
``Within one step'' means that two scores differ by at most one adjacent rubric level.
}
\label{tab:human_metric_consistency}
\begin{tabular}{lcccc}
\toprule
Metric pair & Pearson $r$ & Spearman $\rho$ & Exact match & Within one step \\
\midrule
$F_{\mathrm{acc}}$ vs. $C_s$ & 0.52 & 0.42 & 44.4 & 89.6 \\
$F_{\mathrm{acc}}$ vs. $S_{\mathrm{comp}}$ & 0.78 & 0.74 & 27.8 & 91.7 \\
$C_s$ vs. $S_{\mathrm{comp}}$ & 0.39 & 0.34 & 34.7 & 68.8 \\
\bottomrule
\end{tabular}
\end{table}

\paragraph{Interpretation.}
The clinician audit strengthens the main finding of MedCTA. Even when judged by a human clinical reviewer, current tool agents rarely produce trajectories that are faithful, well-grounded, and complete. The best audited model remains far from saturation, and only 6.9\% of all audited rollouts reach a strong mean clinical score. The audit also shows that automated evaluation is directionally useful but imperfectly calibrated: it broadly agrees with human judgment at the model level, especially for clinical faithfulness, while context integration and semantic completeness remain more difficult to judge automatically. We therefore use the automated clinical-reasoning metrics as scalable diagnostics, not as a replacement for expert clinical review.

\newpage
\section{Instruction for Annotators}
\label{app: mcta_instruction}
\begin{figure}[htbp]
\centering
\begin{tcolorbox}[colframe=black, colback=white]

{\centering \textbf{Technical Annotation Guidelines for Executable Tool-Chain Verification} \par}

\vspace{0.3cm}

\textbf{Objective:}  
Ensure that each annotated reasoning trajectory is \emph{executable, valid, and minimal} under the deployed tool library, while preserving correctness with respect to the clinical objective.

\vspace{0.3cm}

\textbf{Verification Protocol:}

\begin{itemize}
    \item \textbf{Schema Compliance:}  
    Verify that every tool invocation strictly adheres to the predefined interface, including correct argument names, data types, and formatting consistent with the tool specification.

    \item \textbf{Tool Validity:}  
    Ensure that each selected tool belongs to the available tool set $\mathcal{D} = \{\texttt{OCR}, \texttt{ImageDescription}, \texttt{RegionAttributeDescription}, \texttt{GoogleSearch}, \\
    \texttt{Calculator}\}$ and is appropriate for the intended subgoal.

    \item \textbf{Argument Correctness:}  
    Check that all inputs are both syntactically valid and semantically grounded in the available evidence (e.g., image content, prior tool outputs, or retrieved knowledge).

    \item \textbf{Execution Robustness:}  
    Identify and eliminate brittle or failure-prone steps (e.g., unnecessary region specifications, ambiguous OCR dependencies, or redundant intermediate queries) that may reduce reproducibility.

    \item \textbf{Minimality and Efficiency:}  
    Remove redundant, circular, or over-decomposed steps. Each tool call should contribute directly to advancing the reasoning toward the final answer.

    \item \textbf{Trace Consistency:}  
    Ensure that the trajectory forms a coherent chain of reasoning, where each step $s_i$ logically follows from preceding outputs and maintains consistency of intermediate evidence.

    \item \textbf{Deterministic Executability:}  
    Confirm that the full trajectory can be executed sequentially without ambiguity, missing dependencies, or undefined intermediate variables.

\end{itemize}

\vspace{0.3cm}

\textbf{Constraints:}
\begin{itemize}
    \item Do not introduce new tools beyond the defined library.
    \item Do not alter the original clinical question or intended outcome.
    \item Avoid artificial expansion of reasoning steps that do not improve correctness or clarity.
\end{itemize}

\vspace{0.2cm}

\textbf{Deliverable:}  
Produce a corrected executable trajectory $\pi = \{s_i\}_{i=1}^m$, consisting of a minimal sequence of valid tool calls with well-formed arguments and consistent intermediate outputs.

\end{tcolorbox}
\caption{Annotation protocol for validating executable tool-based reasoning trajectories in \ours.}
\label{fig:technical_annotation}
\end{figure}
\begin{figure}[htbp]
\centering
\begin{tcolorbox}[colframe=black, colback=white]

{\centering \textbf{Clinical Annotation Guidelines for Medical Reasoning Validation} \par}

\vspace{0.3cm}

\textbf{Objective:}  
Ensure that each reasoning trajectory reflects clinically sound, evidence-based decision-making and adheres to realistic medical workflows.

\vspace{0.3cm}

\textbf{Validation Protocol:}

\begin{itemize}
    \item \textbf{Clinical Soundness:}  
    Verify that intermediate tool outputs correspond to medically meaningful and plausible findings (e.g., anatomically consistent observations, valid clinical interpretations).

    \item \textbf{Reasoning Completeness:}  
    Ensure that all critical diagnostic or interpretive steps required for the task are present, without omission of key clinical reasoning stages.

    \item \textbf{Logical Coherence:}  
    Confirm that the trajectory follows a coherent clinical reasoning pattern (e.g., observation $\rightarrow$ interpretation $\rightarrow$ conclusion), consistent with standard diagnostic workflows.

    \item \textbf{Evidence Attribution:}  
    Check that each conclusion is explicitly grounded in prior evidence, including visual findings, extracted text, or retrieved knowledge, and that the final answer $\mathcal{A}$ is fully supported by intermediate steps.

    \item \textbf{Safety and Reliability:}  
    Identify and eliminate unsupported assumptions, hallucinated findings, or clinically unsafe inferences that could lead to incorrect or misleading conclusions.

    \item \textbf{Uncertainty Handling:}  
    Ensure that ambiguous or insufficient evidence is appropriately reflected in the reasoning, avoiding overconfident conclusions when clinical certainty is not justified.

\end{itemize}

\vspace{0.3cm}

\textbf{Correction Guidelines:}
\begin{itemize}
    \item Revise incorrect or clinically implausible interpretations.
    \item Insert missing but necessary diagnostic reasoning steps.
    \item Remove steps that contradict established medical knowledge or evidence.
    \item Flag trajectories with unresolved ambiguity or potential clinical risk.
\end{itemize}

\vspace{0.2cm}

\textbf{Deliverable:}  
Produce a clinically validated trajectory and assign an outcome label indicating whether the final answer is (i) correct, (ii) partially correct, or (iii) requires revision, based on consistency with clinical evidence and reasoning.

\end{tcolorbox}
\caption{Clinical validation protocol for assessing medical correctness and reasoning quality in \ours trajectories.}
\label{fig:clinical_annotation}
\end{figure}

\clearpage
\newpage
\section{Case Studies: Stepwise Construction of Executable Clinical Tool Trajectories}
\label{mcta_pipeline}
To illustrate how \ours examples are constructed, corrected, and validated, Figures~\ref{fig:appendix_case_example1} and~\ref{fig:appendix_case_12915877_fig2} present two representative case studies that trace the full evolution of a benchmark sample from its original dataset form to its final executable trajectory. In both cases, the construction process is organized into five stages: (1) the original dataset instance and reference question-answer pair, (2) the initial GPT-generated reasoning and tool sequence, (3) technical verification of tool validity and execution logic, (4) clinical review and medical validation, and (5) the final executable trajectory retained in the benchmark.

\subsection{Trajectory for Histological Pattern Inference}
Figure~\ref{fig:appendix_case_example1} shows a histology example in which the original sample contained multiple questions, including one asking what type of cells are depicted across the panels. During review, the clinical team identified that the earlier answer introduced biological context that was not directly inferable from the image alone, such as proximal colon specificity, colonization status, and experimental reconstitution details. This motivated a revision of both the question-answer interpretation and the generated trajectory. The updated pipeline was redesigned to ground the response only in visible evidence from the figure itself, namely the panel labels \textit{Naive}, \textit{Th1}, and \textit{Th2}, together with the observed gut mucosal histology. Technically, the revised trajectory uses \texttt{OCR} to extract the panel labels, \texttt{ImageDescription} to characterize the overall histologic context, and \texttt{RegionAttributeDescription} to confirm the presence of structural and inflammatory differences across the labeled regions. After technical verification, the new GPT-generated trajectory was explicitly handed off for clinical review, where unsupported claims were removed and the final answer was approved as image-grounded. This example illustrates how the pipeline corrects over-interpretation by forcing both the reasoning trace and the final answer to remain tied to directly observable evidence.

\subsection{Trajectory for Sentinal Node Metastasis Detection}
Figure~\ref{fig:appendix_case_12915877_fig2} highlights a different but equally important failure mode: under-justified inference from figure labels to clinical methods. In this case, the original question asked which methods were used to detect metastases in sentinel nodes. The figure visibly contained assay labels such as \textit{H\&E}, \textit{CK-18}, \textit{CEA}, \textit{hTRT}, and \textit{MUC-1}, and the initial GPT-generated trajectory attempted to infer the underlying detection methods from these labels using only image-based tools. During technical and clinical review, it was determined that this inference could not be justified from the image alone, because mapping these labels to histopathology and molecular detection workflows requires external biomedical knowledge. To address this, the trajectory was revised by inserting an explicit \texttt{GoogleSearch} step after OCR extraction of the assay labels. This search was used to establish that H\&E corresponds to conventional histopathology, while CK-18, CEA, hTRT, and MUC-1 are biomarkers used in molecular assays such as RT-PCR and related nucleic acid-based methods. The placement of this search step was then further refined so that the trajectory first extracts visible evidence, then retrieves the necessary domain knowledge, and finally verifies that the figure indeed compares histologic and molecular detection approaches. The clinically validated final answer therefore reflects both what is shown in the chart and what must be supplied through justified external knowledge.

These examples illustrate the purpose of the staged \ours construction process. The initial GPT trajectory is treated as a draft hypothesis, not a final solution. It is audited from two perspectives: a technical review ensures tool validity, executable inputs, and coherent sequencing, while a clinical review checks that reasoning and answers are medically sound and grounded in visual evidence. When issues arise, the question, answer, and tool chain are revised and re-validated. This workflow ensures high-quality examples with not just correct answers, but executable, clinically grounded reasoning paths, allowing the benchmark to evaluate both task accuracy and the ability to produce valid multi-step tool trajectories.
\label{pipeline_detailed}
\begin{center}
\begin{tcolorbox}[
    enhanced,
    breakable,
    colframe=green!70!black,
    colback=green!2,
    coltitle=black,
    boxrule=1.1pt,
    arc=6pt,
    width=\textwidth,
    title=\textbf{\large Example 1: Anatomical Variant Identification},
    fonttitle=\bfseries,
    halign title=center
]
\small

% =========================
% Stage 1: Original sample
% =========================
\begin{stagebox}[colback=blue!2, colframe=blue!95!black]{Stage 1: Original Dataset Sample}
\begin{center}
\begin{minipage}[c]{0.42\textwidth}
    \centering
    \includegraphics[width=0.92\textwidth]{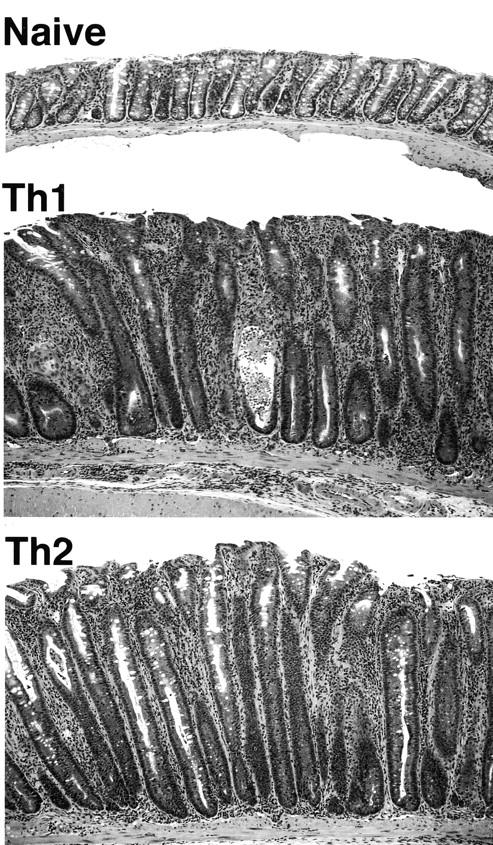}
    
    \vspace{3pt}
    {\small\ttfamily 11781367\_fig4.jpg}
\end{minipage}
\end{center}

\vspace{4pt}
\textbf{Question 1:} \textit{What type of cells are shown in the image?}

\textbf{Reference Answer:} \textbf{Proximal colon tissue with naïve T cells, Th1 cells, and Th2 cells}

\vspace{4pt}
\textbf{Question 2:} \textit{What is the magnification of the image?}

\textbf{Reference Answer:} \textbf{100$\times$}

\vspace{4pt}
\textbf{Question 3:} \textit{What are the characteristics of the colon tissue in the Th1 and Th2 groups?}

\textbf{Reference Answer:} \textbf{Inflammatory infiltrates, epithelial hyperplasia, mucin depletion}
\end{stagebox}

% ======================================
% Stage 2: Initial GPT-generated trace
% ======================================
\begin{stagebox}[colback=yellow!3, colframe=orange!90!black]{Stage 2: Initial GPT-Generated Trajectory}
\textbf{Agentic Query: } For the provided histology figure, determine the stated original magnification (e.g., 40×/100×/400×) by inspecting any scale/magnification annotation on the panel. Report the magnification exactly as written?

\textbf{Predicted Tools:} \texttt{ImageDescription, OCR, RegionAttributeDescription}

\textbf{Draft Reasoning Plan:}
\begin{enumerate}[leftmargin=1.4em, itemsep=2pt, topsep=2pt]
    \item Inspect the histology panel to identify any visible magnification or scale annotation.
    \item Extract text from the image to locate magnification information.
    \item Verify potential annotation regions if magnification is not directly visible.
\end{enumerate}

\textbf{Draft Tool Trace:}
\begin{itemize}[leftmargin=1.4em, itemsep=2pt, topsep=2pt]
    \item \texttt{ImageDescription(image)} $\rightarrow$ ``Histology figure with panels labeled Naive, Th1, and Th2; no clear magnification visible.''
    \item \texttt{OCR(image)} $\rightarrow$ ``Naive, Th1, Th2'' (no magnification text found)
    \item \texttt{RegionAttributeDescription(upper region)} $\rightarrow$ ``No magnification text present''
    \item \texttt{RegionAttributeDescription(lower region)} $\rightarrow$ ``No magnification text present''
\end{itemize}

\textbf{Draft Answer:} \textbf{100$\times$}
\end{stagebox}

% ======================================
% Stage 3: Technical verification
% ======================================
\begin{stagebox}[colback=cyan!3, colframe=cyan!60!black]{Stage 3: Technical Annotation and Tool-Chain Verification}
\textbf{Verification Focus:}
\begin{itemize}[leftmargin=1.4em, itemsep=2pt, topsep=2pt]
    \item Confirm that only tools from the deployed library are used.
    \item Ensure that tool calls are minimal and executable.
    \item Check whether any redundant or unsupported intermediate steps are present.
\end{itemize}

\textbf{Technical Assessment:}
\begin{itemize}[leftmargin=1.4em, itemsep=2pt, topsep=2pt]
    \item \textbf{Tool validity:} Valid. \texttt{OCR}, \texttt{ImageDescription}, and \texttt{RegionAttributeDescription} are part of the deployed tool set.
    \item \textbf{Schema compliance:} Valid. All tool inputs are well-formed and correctly specified.
    \item \textbf{Minimality:} Valid. OCR is required to extract panel labels, and additional tools support contextual verification.
    \item \textbf{Execution robustness:} Valid. The trajectory avoids unsupported steps and ensures reliable extraction and validation.
\end{itemize}

\textbf{Correction Applied:} 
The original question and answer were revised based on clinical feedback to remove unsupported contextual details. A new trajectory was generated that grounds the answer strictly in observable image evidence (panel labels and histological patterns), resulting in a corrected and executable pipeline.
\end{stagebox}

\begin{stagebox}[colback=purple!3, colframe=purple!1000!black]{Stage 3.5: Clinical Review Handoff}
\textbf{Post-Technical Step: Clinical Review Handoff}
\begin{itemize}[leftmargin=1.4em, itemsep=2pt, topsep=2pt]
    \item The updated GPT-generated trajectory is forwarded to clinical experts for validation.
    \item Clinical reviewers assess whether the reasoning and final answer are strictly grounded in visible image evidence.
    \item Any remaining inconsistencies or unsupported assumptions are flagged for correction.
    \item Upon successful validation, the trajectory is approved for final use.
\end{itemize}
\end{stagebox}
% ======================================
% Stage 4: Clinical validation
% ======================================
\begin{stagebox}[colback=red!2, colframe=red!1000!white]{Stage 4: Clinical Annotation and Medical Validation}
\textbf{Clinical Review Focus:}
\textbf{Clinical Review Focus:}
\begin{itemize}[leftmargin=1.4em, itemsep=2pt, topsep=2pt]
    \item Verify that the answer is grounded only in information directly visible in the histology figure.
    \item Confirm that the panel labels are sufficient to identify the depicted cell conditions.
    \item Ensure that no unsupported biological context is introduced beyond the image evidence.
\end{itemize}

\textbf{Clinical Assessment:}
\begin{itemize}[leftmargin=1.4em, itemsep=2pt, topsep=2pt]
    \item The panel labels clearly indicate the conditions \textbf{Naive}, \textbf{Th1}, and \textbf{Th2}.
    \item The histology image supports describing the tissue as gut mucosal tissue with these labeled conditions across panels.
    \item The earlier answer included unsupported details such as proximal colon, \textit{pnir15.OVA-E. coli} colonization, and reconstitution context, which cannot be directly inferred from the image alone.
    \item The corrected answer, \textbf{``The sections display the naive T cells, Th1 cells, and Th2 cells on the gut mucosal tissue.''}, is clinically appropriate and image-grounded.
\end{itemize}

\textbf{Clinical Verdict:} Correct after answer revision.
\end{stagebox}

% ======================================
% Stage 5: Final trajectory
% ======================================
\begin{stagebox}[colback=green!3, colframe=green!60!black]{Stage 5: Final Executable Trajectory}

\textbf{Final Query: } Using the provided histology figure, identify what type of cells/conditions are depicted across the panels?

\textbf{Final Tool Set:} \texttt{OCR, ImageDescription, RegionAttributeDescription}

\textbf{Executable Trajectory $\pi$:}
\begin{enumerate}[leftmargin=1.5em, itemsep=3pt, topsep=3pt]
    \item \textbf{Tool:} \texttt{OCR} \\
    \textbf{Input:} histology image \texttt{(image/image\_29.jpg)} \\
    \textbf{Output:} ``Naive, Th1, Th2''

    \item \textbf{Tool:} \texttt{ImageDescription} \\
    \textbf{Input:} histology image \texttt{(image/image\_29.jpg)} \\
    \textbf{Output:} ``Histology micrographs of gut mucosal tissue arranged in panels labeled Naive, Th1, and Th2, showing variations in cellular density.''

    \item \textbf{Tool:} \texttt{RegionAttributeDescription} \\
    \textbf{Input:} \texttt{bbox = (40, 360, 470, 610)} \\
    \textbf{Output:} ``Differences in cellular infiltration corresponding to Naive, Th1, and Th2 conditions.''
\end{enumerate}

\vspace{4pt}
\begin{center}
\fcolorbox{blue!50!black}{blue!5}{
    \parbox{0.84\textwidth}{
        \centering
        {\color{blue!80!black}\textbf{Final Answer}}\\[4pt]
        \textbf{Naive T cells, Th1 cells, and Th2 cells}
    }
}
\end{center}

\vspace{4pt}
\textbf{Final Justification:}
OCR identifies the panel labels (Naive, Th1, Th2), and visual inspection confirms corresponding histological differences in gut mucosal tissue, supporting that the sections depict these T cell conditions.
\end{stagebox}

\end{tcolorbox}
\captionof{figure}{Stepwise construction of an executable and clinically validated trajectory for anatomical variant identification.}
\label{fig:appendix_case_example1}
\end{center}

\begin{center}
\begin{tcolorbox}[
    enhanced,
    breakable,
    colframe=green!70!black,
    colback=green!2,
    coltitle=black,
    boxrule=1.1pt,
    arc=6pt,
    width=\textwidth,
    title=\textbf{\large Example 2: Sentinel Node Metastasis Detection Methods},
    fonttitle=\bfseries,
    halign title=center
]
\small

% =========================
% Stage 1: Original sample
% =========================
\begin{stagebox}[colback=blue!2, colframe=blue!995!black]{Stage 1: Original Dataset Sample}
\begin{center}
\begin{minipage}[c]{0.42\textwidth}
    \centering
    \includegraphics[width=0.92\textwidth]{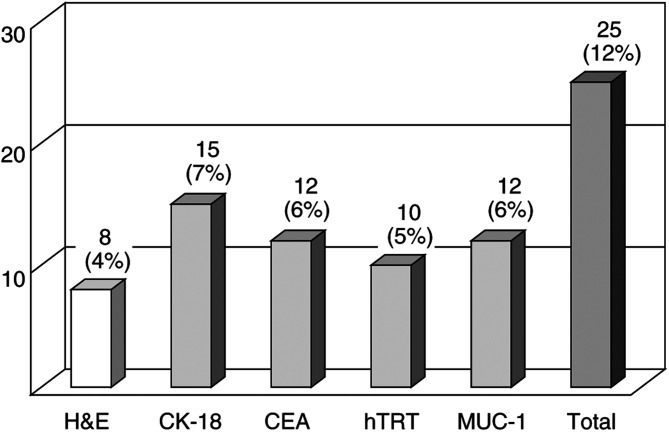}
    
    \vspace{3pt}
    {\small\ttfamily 12915877\_fig2.jpg}
\end{minipage}
\end{center}

\vspace{4pt}
\textbf{Question 1:} \textit{What are the methods used to detect metastases in the image?}

\textbf{Reference Answer:} \textbf{The image shows three methods used to detect metastases in sentinel nodes: conventional histology, hematoxylin and eosin (H\&E) staining, and RT-PCR and Southern blot assay.}

\vspace{4pt}
\textbf{Question 2:} \textit{What are the markers mentioned in the image?}

\textbf{Reference Answer:} \textbf{CK-18, CEA, and hTRT}

\vspace{4pt}
\textbf{Question 3:} \textit{How many nodes were harvested from the patients?}

\textbf{Reference Answer:} \textbf{25 nodes}
\end{stagebox}

% ======================================
% Stage 2: Initial GPT-generated trace
% ======================================
\begin{stagebox}[colback=yellow!3, colframe=orange!90!black]{Stage 2: Initial GPT-Generated Trajectory}
\textbf{Predicted Tools:} \texttt{ImageDescription, OCR, GoogleSearch, RegionAttributeDescription}

\textbf{Draft Reasoning Plan:}
\begin{enumerate}[leftmargin=1.4em, itemsep=2pt, topsep=2pt]
    \item Inspect the figure to determine whether it presents histologic and molecular detection methods.
    \item Extract the biomarker and assay labels shown in the bar chart.
    \item Use external medical knowledge to map the labels to the corresponding laboratory methods.
    \item Verify from the figure that the chart compares histologic and molecular detection approaches.
\end{enumerate}

\textbf{Draft Tool Trace:}
\begin{itemize}[leftmargin=1.4em, itemsep=2pt, topsep=2pt]
    \item \texttt{ImageDescription(image)} $\rightarrow$ ``A bar chart comparing detection rates using H\&E, CK-18, CEA, hTRT, and MUC-1 markers.''
    \item \texttt{OCR(image)} $\rightarrow$ ``H\&E, CK-18, CEA, hTRT, MUC-1''
    \item \texttt{GoogleSearch(``What methods are used to detect sentinel lymph node metastases using H\&E CK-18 CEA hTRT MUC-1'')} $\rightarrow$ ``H\&E is used for conventional histopathology. CK-18, CEA, hTRT, and MUC-1 are commonly detected using molecular techniques such as RT-PCR or related nucleic acid-based assays to identify micrometastases.''
    \item \texttt{RegionAttributeDescription(region)} $\rightarrow$ ``The figure compares H\&E with biomarker-based detection methods, indicating both histological and molecular approaches.''
\end{itemize}

\textbf{Draft Answer:} \textbf{The figure shows that metastases in sentinel lymph nodes are detected using conventional histopathology (hematoxylin and eosin, H\&E) and molecular techniques such as RT-PCR using biomarkers including CK-18, CEA, hTRT, and MUC-1.}
\end{stagebox}

% ======================================
% Stage 3: Technical verification
% ======================================
\begin{stagebox}[colback=cyan!3, colframe=cyan!60!black]{Stage 3: Technical Annotation and Tool-Chain Verification}
\textbf{Verification Focus:}
\begin{itemize}[leftmargin=1.4em, itemsep=2pt, topsep=2pt]
    \item Confirm that only tools from the deployed library are used.
    \item Ensure that tool calls are executable and logically ordered.
    \item Check whether external knowledge is required to infer laboratory methods from the biomarker labels.
    \item Remove unsupported assumptions and improve grounding between image evidence and final answer.
\end{itemize}

\textbf{Technical Assessment:}
\begin{itemize}[leftmargin=1.4em, itemsep=2pt, topsep=2pt]
    \item \textbf{Tool validity:} Valid. \texttt{ImageDescription}, \texttt{OCR}, \texttt{GoogleSearch}, and \texttt{RegionAttributeDescription} are all part of the deployed tool set.
    \item \textbf{Schema compliance:} Valid. All tool inputs are correctly formed and executable.
    \item \textbf{Minimality:} Revised. The original trajectory inferred laboratory methods directly from the chart labels, which was insufficiently grounded. \texttt{GoogleSearch} was added to support mapping H\&E and biomarker labels to detection methods.
    \item \textbf{Execution robustness:} Valid after revision. The updated trajectory first extracts visible labels, then uses external knowledge, and finally verifies that the figure compares histologic and molecular approaches.
\end{itemize}

\textbf{Correction Applied:}
Based on clinical feedback, a \texttt{GoogleSearch} step was inserted after OCR to justify the inference from biomarker labels to the underlying detection methods. The updated trajectory uses the clinician-provided query and revised reasoning step, and the placement of the search step was further corrected to maintain a logically grounded tool chain.
\end{stagebox}

% ======================================
% Stage 3.5: Clinical handoff
% ======================================
\begin{stagebox}[colback=purple!3, colframe=purple!80!black]{Stage 3.5: Clinical Review Handoff}
\textbf{Post-Technical Step: Clinical Review Handoff}
\begin{itemize}[leftmargin=1.4em, itemsep=2pt, topsep=2pt]
    \item The revised GPT-generated trajectory is forwarded to clinical experts for validation.
    \item Clinical reviewers assess whether the inferred detection methods are adequately supported by the visible chart labels and the external search evidence.
    \item The placement and necessity of the \texttt{GoogleSearch} step are specifically reviewed.
    \item Any remaining unsupported assumptions are flagged and corrected before final approval.
\end{itemize}
\end{stagebox}

% ======================================
% Stage 4: Clinical validation
% ======================================
\begin{stagebox}[colback=red!2, colframe=red!90!white]{Stage 4: Clinical Annotation and Medical Validation}
\textbf{Clinical Review Focus:}
\begin{itemize}[leftmargin=1.4em, itemsep=2pt, topsep=2pt]
    \item Verify that the figure alone shows assay labels rather than explicitly naming all laboratory methods.
    \item Confirm that external medical knowledge is needed to connect H\&E and the listed biomarkers to histopathology and molecular assays.
    \item Ensure that the final answer remains clinically accurate without overstating what is directly visible in the image.
\end{itemize}

\textbf{Clinical Assessment:}
\begin{itemize}[leftmargin=1.4em, itemsep=2pt, topsep=2pt]
    \item The figure visibly presents the labels \textbf{H\&E}, \textbf{CK-18}, \textbf{CEA}, \textbf{hTRT}, and \textbf{MUC-1}.
    \item These labels alone do not fully specify the laboratory detection methods without domain knowledge.
    \item Adding an external search step is clinically appropriate because H\&E corresponds to conventional histopathology, whereas CK-18, CEA, hTRT, and MUC-1 are biomarkers commonly used in molecular detection workflows such as RT-PCR.
    \item The revised answer is clinically acceptable because it distinguishes between histopathologic detection and molecular biomarker-based detection while remaining consistent with the figure content.
\end{itemize}

\textbf{Clinical Verdict:} \textbf{Correct after trajectory revision}
\end{stagebox}

% ======================================
% Stage 5: Final trajectory
% ======================================
\begin{stagebox}[colback=green!3, colframe=green!60!black]{Stage 5: Final Executable Trajectory}
\textbf{Final Tool Set:} \texttt{ImageDescription, OCR, GoogleSearch, RegionAttributeDescription}

\textbf{Executable Trajectory $\pi$:}
\begin{enumerate}[leftmargin=1.5em, itemsep=3pt, topsep=3pt]
    \item \textbf{Tool:} \texttt{ImageDescription} \\
    \textbf{Input:} figure image \texttt{(image/image\_55.jpg)} \\
    \textbf{Output:} ``A bar chart comparing detection rates using H\&E, CK-18, CEA, hTRT, and MUC-1 markers.''

    \item \textbf{Tool:} \texttt{OCR} \\
    \textbf{Input:} figure image \texttt{(image/image\_55.jpg)} \\
    \textbf{Output:} ``H\&E, CK-18, CEA, hTRT, MUC-1''

    \item \textbf{Tool:} \texttt{GoogleSearch} \\
    \textbf{Input:} ``What methods are used to detect sentinel lymph node metastases using H\&E CK-18 CEA hTRT MUC-1'' \\
    \textbf{Output:} ``H\&E is used for conventional histopathology. CK-18, CEA, hTRT, and MUC-1 are commonly detected using molecular techniques such as RT-PCR or related nucleic acid-based assays to identify micrometastases.''

    \item \textbf{Tool:} \texttt{RegionAttributeDescription} \\
    \textbf{Input:} \texttt{bbox = (40, 220, 580, 280)} \\
    \textbf{Output:} ``The figure compares H\&E with biomarker-based detection methods, indicating both histological and molecular approaches.''
\end{enumerate}

\vspace{4pt}
\begin{center}
\fcolorbox{blue!50!black}{blue!5}{
    \parbox{0.84\textwidth}{
        \centering
        {\color{blue!80!black}\textbf{Final Answer}}\\[4pt]
        \textbf{conventional histology, hematoxylin and eosin (H\&E) staining, and RT-PCR and Southern blot assay}
    }
}
\end{center}

\vspace{4pt}
\textbf{Final Justification:}
The chart explicitly shows H\&E and the biomarker labels CK-18, CEA, hTRT, and MUC-1. External medical knowledge is required to map these labels to their corresponding detection methods. Together, the figure and the search evidence support that the figure compares conventional histopathologic detection with molecular biomarker-based detection of sentinel node metastases.
\end{stagebox}

\end{tcolorbox}

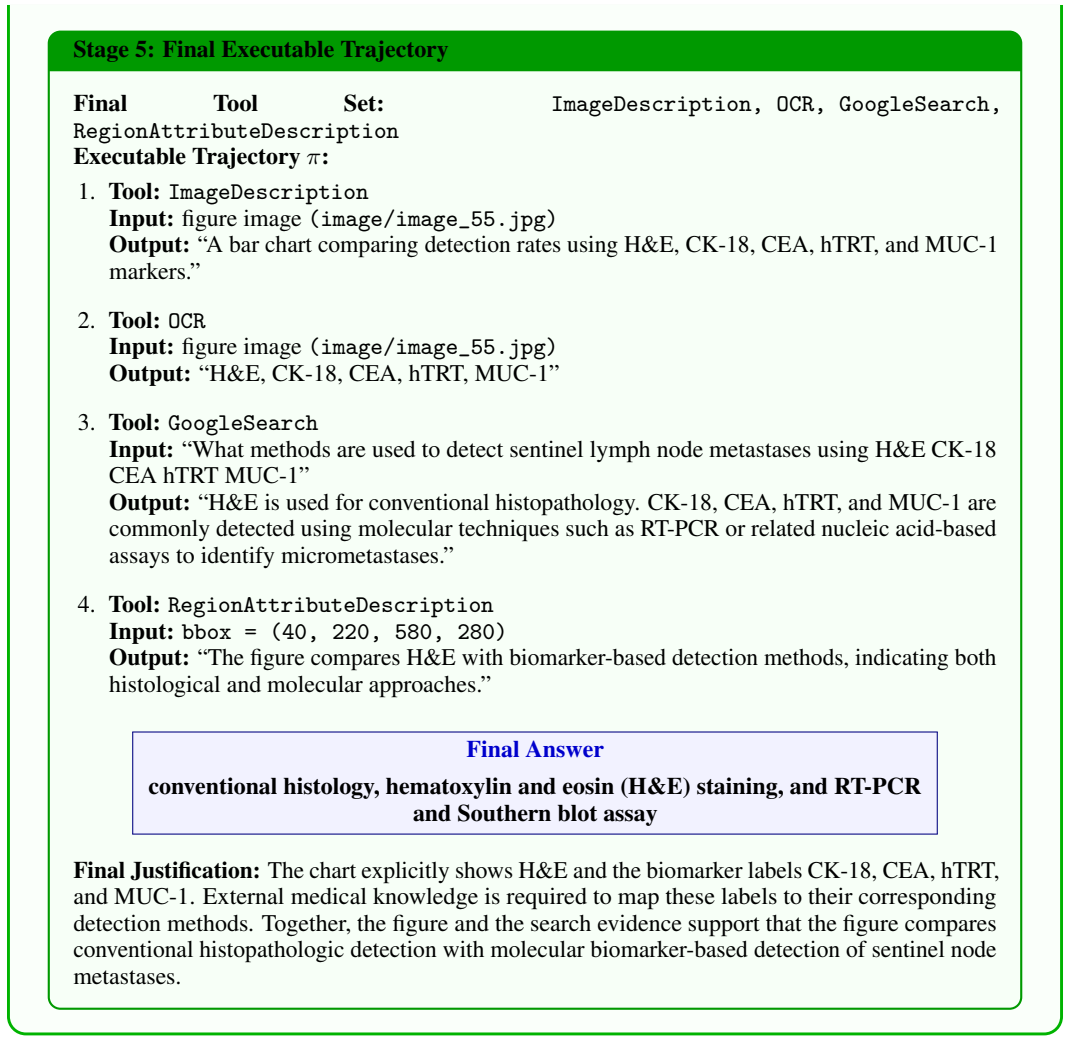
\captionof{figure}{Stepwise construction of an executable and clinically validated trajectory for identifying metastasis detection methods in sentinel nodes.}
\label{fig:appendix_case_12915877_fig2}
\end{center}

\clearpage
\section{Prompting Templates for Agent Trajectory Construction}
\label{mcta_prompts}
To ensure reproducible agent behavior, we employ two complementary prompting components: (1) a \textit{ReAct-style execution prompt} that governs how the agent reasons, invokes tools, and processes observations during inference, and (2) a \textit{trajectory generation prompt} used to synthesize a single executable ground-truth trajectory for each medical VQA sample. Following the GTA framework~\cite{GTA}, the agent operates in a structured reasoning-and-acting loop, interleaving concise reasoning steps, explicit tool calls, tool responses, and a final answer, ensuring that tool usage is integrated into a coherent and interpretable decision process. For benchmark construction, the generation prompt enforces strict constraints on output format, requiring a valid JSON trajectory with minimal and executable tool usage, realistic tool outputs, and exact preservation of the original answer, while encouraging clinically grounded strategies such as early OCR for labels, image-based contextualization, and region-level verification. Figures~\ref{fig:react_prompt_template} and~\ref{fig:gta_prompt_template} illustrate these two prompts, which together define both the runtime behavior of the agent and the standardized procedure for constructing clinically valid, executable trajectories.

\begin{figure*}[t]
\centering
\begin{tcolorbox}[
    enhanced,
    breakable,
    colback=gray!3,
    colframe=black!70,
    boxrule=0.8pt,
    arc=4pt,
    width=\textwidth,
    title=\textbf{ReAct-Style Execution Prompt Used in the Agent System},
    fonttitle=\bfseries
]
\small
\begin{verbatim}
CALL_PROTOCOL_EN = """You are a assistant who can utilize
external tools. {tool_description}
To use a tool, please use the following format:
‘‘‘
{thought}Think what you need to solve, do you need to use
tools?
{action}the tool name, should be one of [{action_names}]
{action_input}the input to the action
‘‘‘
The response after utilizing tools should using the following
format:
‘‘‘
{response}the results after call the tool.
‘‘‘
If you already know the answer, or you do not need to use
tools, please using the following format to reply:
‘‘‘
{thought}the thought process to get the final answer
{finish}final answer
‘‘‘
Begin!"""
\end{verbatim}
\end{tcolorbox}
\caption{
ReAct-style execution prompt used in the \ours agent system, adapted
from the GTA~\cite{GTA}-style tool-use protocol. The prompt specifies the action format,
tool input schema requirement, observation format, and final-answer format.
}
\label{fig:react_prompt_template}
\end{figure*}

The ReAct-style execution prompt governs how the agent interacts with external tools at inference time by structuring the process into explicit reasoning, tool selection, tool input, and observation steps, making the decision process modular, interpretable, and auditable, an essential requirement in clinical settings. In contrast, the trajectory generation prompt is used offline to construct benchmark examples, enforcing strict JSON structure, executable tool usage, realistic outputs, and exact preservation of the original answer, ensuring that each trajectory serves as a valid reference. Both prompts are necessary: the execution prompt enables consistent runtime reasoning and tool interaction, while the generation prompt standardizes benchmark construction and comparability across samples. This dual design is particularly important in medical applications, where prompt quality directly impacts trajectory reliability, encouraging grounded reasoning, justified tool usage, and clinically valid, verifiable outcomes.

\begin{figure*}[t]
\centering
\begin{tcolorbox}[
    enhanced,
    colback=gray!3,
    colframe=black!70,
    boxrule=0.8pt,
    arc=4pt,
    width=\textwidth,
    title=\textbf{Prompt for Generating Ground-Truth Agent Trajectories},
    fonttitle=\bfseries
]
\small
\begin{verbatim}
SYSTEM_PROMPT = r"""
You generate ONE agentic ground-truth trajectory for a medical VQA sample.

You will receive:
- toolmetadata (dict keyed by tool name),
- one medical image,
- original_question,
- original_answer,
- sample_key.

You MUST output ONE valid JSON object and nothing else.
The JSON must have EXACTLY ONE top-level key = sample_key.
That key maps to an object with EXACT keys:
- tools: list[tool_spec] (subset of provided toolmetadata; include ONLY tools 
actually used)
- files: list of file objects (must include the given image path)
- dialogs: list of messages (role=user/assistant/tool)
- gt_answer: {"whitelist": [[original_answer_exact]], "blacklist": null}

HARD FORMAT RULES
A) Tool calls MUST look exactly like:
  {"role":"assistant","tool_calls":[{"type":"function","function":
  {"name":"ToolName","arguments":{...}}}, ...],"thought":"..."}
B) After each tool call, add a tool message:
  {"role":"tool","name":"ToolName","content":{"type":"text","content":"
  <string>"}}
   - tool content.type MUST be "text" always
   - tool content.content MUST be a string (even for ints)
C) The FINAL message in dialogs MUST be:
  {"role":"assistant","content":"<EXACT original_answer>"}  
  (character-for-character)

CRITICAL CONSTRAINTS
- Minimum tool calls = 2 (use more only if truly needed; 2–6 typical).
- Keep a clinician/radiologist/pathologist style of reasoning: brief, 
grounded, stepwise.
- The rewritten question should be answerable with the tools you use 
(agentic, but not silly).
- Tool outputs must be realistic:
   - OCR: "(x1, y1, x2, y2) TEXT" lines separated by "\n" (empty allowed
   if no text)
   - RegionAttributeDescription: short attribute grounded on provided bbox 
   (numeric coordinates)
   - ImageDescription: brief factual description
   - GoogleSearch: 3–6 short result lines
   - Calculator: numeric string only
   - exact preservation of the original answer (answer whitelist)

ADAPTIVE BEST-CLINICAL STRATEGY (choose per sample)
1) If there might be labels/markers/measurements: use OCR early.
2) Use ImageDescription to orient (modality, anatomy, figure type).
3) Use RegionAttributeDescription with a plausible numeric bbox to verify a 
key structure/finding.
4) Optionally add a sanity-check step (e.g., another 
RegionAttributeDescription or GoogleSearch if the question is definitional).
5) End with the exact original answer string.

Return valid JSON only.
"""
\end{verbatim}
\end{tcolorbox}
\caption{Prompt template used to synthesize a single executable ground-truth trajectory for each medical VQA sample. The prompt enforces strict JSON structure, executable tool usage, realistic tool outputs, and exact preservation of the reference answer.}
\label{fig:gta_prompt_template}
\end{figure*}

\clearpage

\section{Metrics and Implementation Details}
\label{app:metrics_details}

\paragraph{Hardware and Setup.}
All experiments were conducted on NVIDIA A100 GPUs (80GB), ensuring consistent hardware conditions across models. Each evaluated \lmm agent receives the clinical context $\mathcal{X}$ and query $\mathcal{Q}$, and interacts with the executable tool library $\mathcal{D}$ under a standardized agent framework. Tool invocation follows a structured ReAct-style prompting protocol, enabling step-wise reasoning and deterministic execution.

\paragraph{Tool Call Execution.}
All tools are implemented as callable Python functions with strictly defined input–output schemas. Model-generated reasoning traces are parsed as structured JSON, and tools are executed sequentially according to the predicted actions. A tool call is considered \emph{valid} if (i) the selected tool exists in $\mathcal{D}$, (ii) the arguments conform to the required schema, and (iii) execution produces a non-empty output. Invalid formatting, incorrect tool selection, or argument mismatches are recorded as execution failures and incorporated into evaluation statistics.

\begin{figure}[t]
    \centering
    \includegraphics[width=\linewidth]{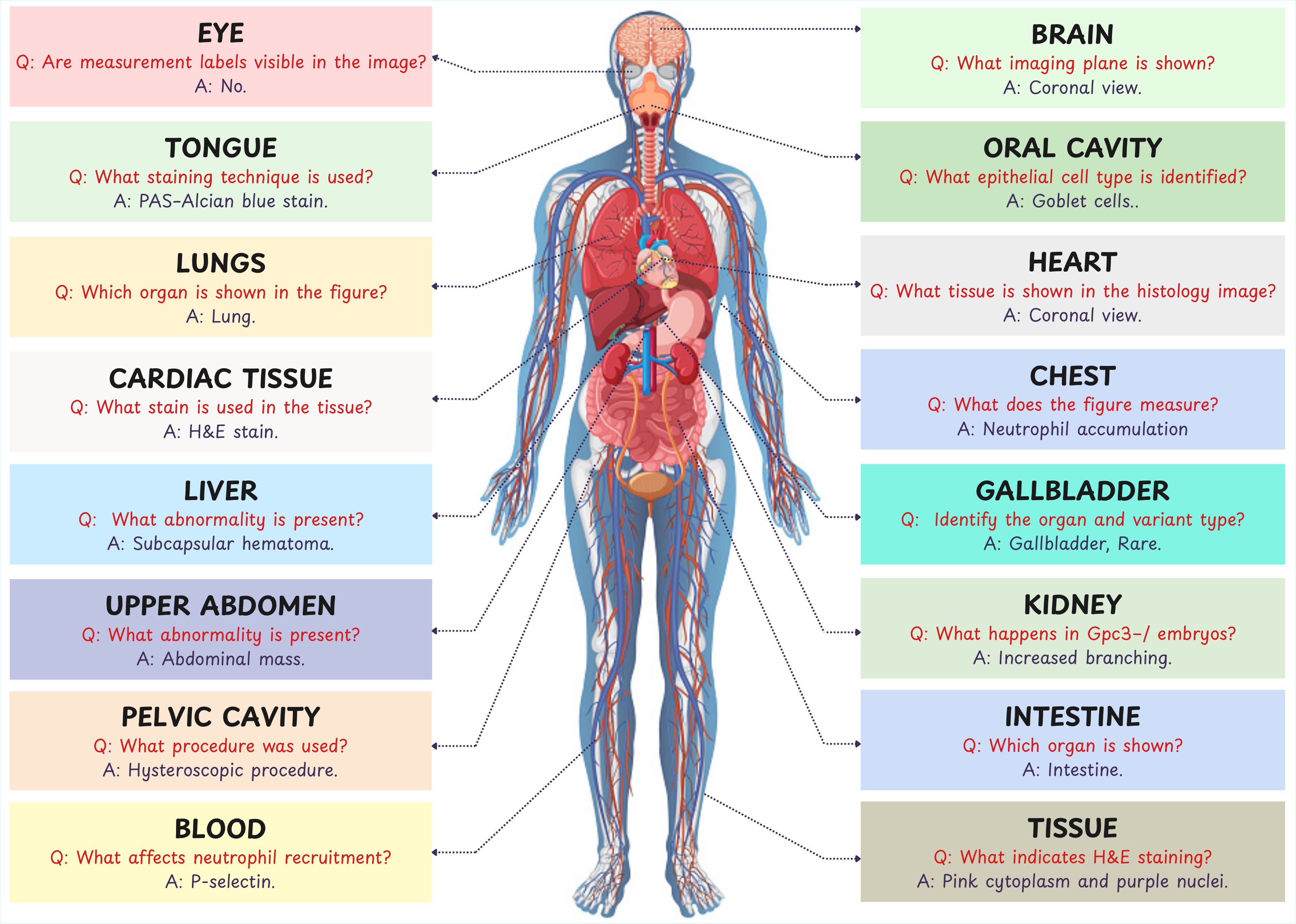}
    \caption{\textbf{Anatomical and modality coverage of MedCTA.} The benchmark spans diverse body regions and clinical inputs, including eye, brain, oral cavity, lungs, heart, liver, kidney, intestine, gallbladder, pelvis, blood, and tissue-level pathology. Each example shows a step-implicit clinical query and its target answer, illustrating that MedCTA evaluates broad cross-specialty medical tool use rather than a single organ system or imaging modality.}
    \label{fig:anatomical_coverage}
\end{figure}

Figure~\ref{fig:anatomical_coverage} summarizes the clinical breadth of \ours. The benchmark covers whole-body anatomy and tissue-level pathology, with tasks drawn from radiology, histology, gross pathology, ophthalmology, and report-based clinical reasoning. Importantly, the queries are not simple recognition prompts: they ask agents to identify findings, interpret stains, recognize variants, reason over measurements, and connect localized evidence to clinically meaningful answers. This diversity ensures that \ours evaluates general medical tool-agent behavior rather than overfitting to one modality, organ system, or question type.

\paragraph{Metrics Overview.}
As summarized in Table~\ref{tab:metrics}, MedCTA employs a multi-level evaluation framework consisting of three modes:
\begin{itemize}
    \item \textbf{Step-by-Step Mode:} Evaluates intermediate tool-use behavior using \textit{InstAcc}, \textit{ToolAcc}, \textit{ArgAcc}, and \textit{SummAcc}, measuring protocol adherence, tool selection correctness, argument validity, and step-level summarization quality.
    
    \item \textbf{Clinical Reasoning Mode:} Assesses trajectory-level coherence and medical grounding via \textit{Clinical Faithfulness} ($F_{acc}$), \textit{Context Integration Score} ($C_s$), and \textit{Semantic Completeness} ($S_{comp}$).
    
    \item \textbf{Outcome Mode:} Measures end-to-end correctness using \textit{Goal Accuracy} ($G_{acc}$).
\end{itemize}

\paragraph{Metric scaling and aggregation.}
All reported metrics are normalized to the range $[0,100]$, where higher is better. Scores are macro-averaged over tasks unless otherwise stated.  Premature final answers before the reference stop point are
counted as under-calls and receive zero credit for the missing remaining
step-level decisions.

\paragraph{Step-by-step metrics.}
Let $s_i=(t_i,\alpha_i,\rho_i)$ be the $i$-th reference step and let
$\hat{s}_i=(\hat{t}_i,\hat{\alpha}_i,\hat{\rho}_i)$ be the model-predicted
step.
\begin{itemize}
    \item \textbf{InstAcc} measures whether the model emits a parseable action
    or final-answer message under the required protocol.
    \item \textbf{ToolAcc} is one if $\hat{t}_i=t_i$ and zero otherwise. If
    multiple tools are clinically equivalent for a task, the accepted set is
    specified in the task metadata.
    \item \textbf{ArgAcc} measures whether $\hat{\alpha}_i$ is executable and
    clinically appropriate. We assign full credit when all required keys, data
    types, and clinically relevant values match after canonicalization. We assign
    half credit when the call is executable and semantically usable but contains
    harmless extra keys, minor formatting drift, or spatial arguments that
    overlap the reference region sufficiently. Non-parseable, non-executable,
    wrong-file, or clinically inconsistent arguments receive zero credit.
    \item \textbf{SummAcc} measures whether the model correctly integrates the
    current observation into the expected intermediate conclusion. It is graded
    using the same rubric as clinical reasoning but restricted to the current
    step.
\end{itemize}

\paragraph{Clinical reasoning metrics.}
Reasoning metrics are graded from the executed trajectory only using
a rubric calibrated with clinician annotations:
\begin{itemize}
    \item \textbf{Clinical Faithfulness} ($F_{\mathrm{acc}}$): whether reasoning
    claims are supported by tool observations and do not contradict the evidence.
    \item \textbf{Context Integration} ($C_s$): whether the trajectory integrates all
    relevant modalities and files rather than relying on a single cue or prior.
    \item \textbf{Semantic Completeness} ($S_{\mathrm{comp}}$): whether all
    clinically necessary findings for the answer are covered.
\end{itemize}
Each dimension is scored on a 0--1 rubric and linearly mapped to $[0,100]$.
Scores are clipped to $[0,100]$; negative values are not permitted.

\paragraph{Outcome metric.}
Goal accuracy ($G_{\mathrm{acc}}$) measures final answer correctness against
the clinician-validated answer whitelist. For closed-form answers, we use exact
or canonicalized matching. For open-ended clinical answers, we use rubric-based
semantic equivalence with clinician-calibrated adjudication.

\section{Additional Experimental Results and Diagnostic Analyses}
\label{app:diagnostics}

This appendix keeps only the supplementary diagnostics referenced in the main text. We focus on the views that are most useful for interpreting benchmark behavior: a backbone-only zero-shot sanity check, the gold-standard tool routing gap, rollout failure dynamics, descriptive conditional diagnostics, horizon effects, and tool-level bottlenecks.

\subsection{gold-standard tool routing gap and controller/reader separation}
\label{app:oracle}

\begin{figure}[t]
    \centering
    
    \begin{subfigure}[t]{0.44\linewidth}
        \centering
        \includegraphics[width=\linewidth]{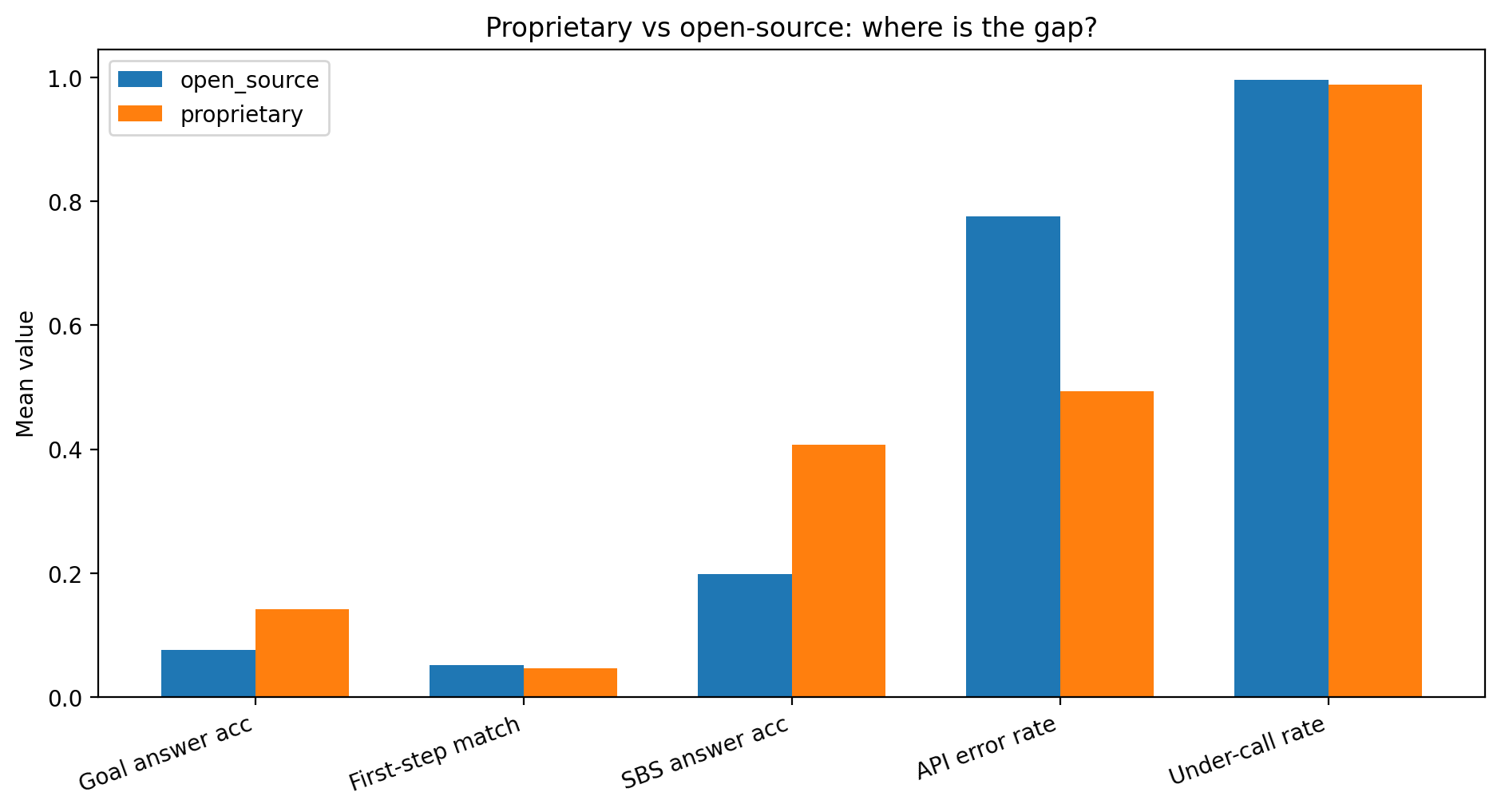}
        \caption{Open-source vs.\ proprietary models under rollout diagnostics.}
        \label{fig:closed_vs_open}
    \end{subfigure}
    \hfill
    \begin{subfigure}[t]{0.55\linewidth}
        \centering
        \includegraphics[width=\linewidth]{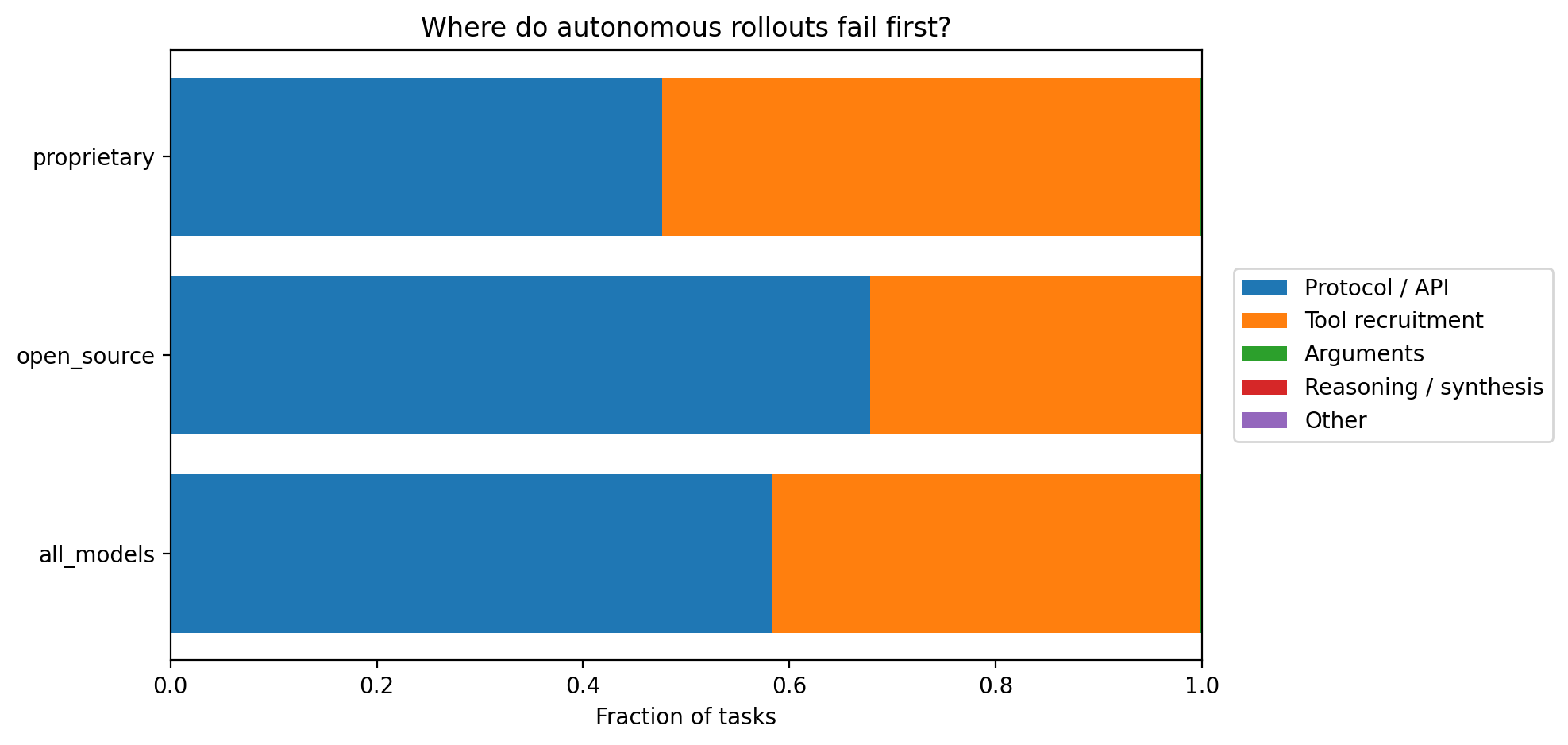}
        \caption{Where autonomous rollouts fail first.}
        \label{fig:failure_bucket_focus}
    \end{subfigure}
    
    \caption{\textbf{Rollout diagnostics across models.} 
    (Left) Proprietary models show lower API/protocol failures and stronger oracle answerability. 
    (Right) Failures are dominated by protocol/API handling and tool recruitment.}
    
    \label{fig:combined_rollout}
\end{figure}

% \input{tables/sbs_goal_gap}

% The clearest controlled analysis is the teacher-forced gold-standard tool routing comparison in Table~\ref{tab:sbs_goal_gap}. For most models, answer accuracy rises sharply once the gold next tool is supplied, confirming that a large fraction of current loss comes from controller instability rather than lack of latent clinical reasoning. The gap is especially large for Claude-haiku-4-5 (\textbf{+51.4}), Claude-opus-4-6 (\textbf{+47.0}), GPT-4o (\textbf{+46.7}), Claude-sonnet-4-6 (\textbf{+43.9}), and Qwen3.5-9B (\textbf{+32.7}).

% At the same time, exact tool-sequence match remains near zero even for the strongest models, indicating that high oracle gains do not come from already-faithful trajectories. Instead, they reflect substantial \emph{conditional answerability}: once the correct evidence stream is externally routed, much more of the task becomes solvable. A few models even show negative oracle gaps, suggesting that part of their autonomous success comes from shortcutting or direct guessing rather than faithful evidence use.

Figure~\ref{fig:closed_vs_open} provides the same story at group level. Proprietary models outperform open-source models not mainly because of dramatically better first actions, but because they are more stable under the interaction protocol and better at exploiting the evidence once correct routing is provided.

\subsection{Autonomous rollout dynamics and early-failure behavior}
\label{app:failures}

\begin{table}[t]
\centering
\small
\caption{Autonomous call-count behavior (\%). Only informative deviations from the dominant pattern are shown.}
\label{tab:call_count_behavior}

\rowcolors{2}{stripeA}{stripeB}

\begin{tabular}{lccc}
\rowcolor{headerBlue}
\color{white}{\textbf{Model}} &
\color{white}{\textbf{Under-call}} &
\color{white}{\textbf{Exact match}} &
\color{white}{\textbf{Over-call}} \\
\midrule

\rowcolor{highlightRow}
\textbf{GPT\_4o} & 91.6 & \textbf{3.74} & \textcolor{gainGreen}{4.67} \\

Qwen3.5-9B & 95.3 & \textbf{4.67} & 0.00 \\

claude-haiku-4-5 & \textcolor{lossRed}{99.1} & 0.00 & \textcolor{gainGreen}{0.93} \\

gpt-5.4-mini & \textcolor{lossRed}{99.1} & 0.00 & \textcolor{gainGreen}{0.93} \\

gemini-3.1-flash-lite-preview & \textcolor{lossRed}{99.1} & 0.00 & \textcolor{gainGreen}{0.93} \\

\midrule

\rowcolor{stripeA}
\textit{all remaining models} & \textcolor{lossRed}{100.0} & 0.00 & 0.00 \\

\bottomrule
\end{tabular}
\end{table}

Table~\ref{tab:call_count_behavior} shows that autonomous rollouts are overwhelmingly dominated by \emph{under-calling}. The only informative deviations from the default pattern come from a very small subset of models: GPT-4o and Qwen3.5-9B are the only models with non-trivial exact tool-count match, and over-calling is confined to GPT-4o and a few models at negligible rates. All remaining models under-call on essentially every task.

Figure~\ref{fig:failure_bucket_focus} sharpens the diagnosis. First failures are concentrated almost entirely in protocol/API handling or tool recruitment, while downstream reasoning and synthesis contribute very little as \emph{first} failure modes. This is the key reason the benchmark remains difficult: many agents fail before stable evidence-conditioned reasoning can even begin.

\subsection{Conditional diagnostics and descriptive partial-oracle views}
\label{app:conditions}

% =========================================================
% Beautiful Condition Uplift Table (consistent styling)
% =========================================================

\begin{table*}[t]
\centering
\small
\caption{Observed (descriptive) answer uplift under different rollout conditions. These are associative (not causal).}
\label{tab:condition_uplift_pooled}

\rowcolors{2}{stripeA}{stripeB}

\begin{adjustbox}{width=\textwidth}
\begin{tabular}{lccccc}
\rowcolor{headerBlue}
\color{white}{\textbf{Condition}} &
\color{white}{\textbf{Support}} &
\color{white}{\textbf{Rate}} &
\color{white}{\textbf{Baseline}} &
\color{white}{\textbf{If condition}} &
\color{white}{\textbf{Uplift}} \\
\midrule

% ================= ALL MODELS =================
\rowcolor{headerBlue!15}
\multicolumn{6}{l}{\textbf{All models}} \\

All matched arguments schema-clean & 145 & 7.13 & 10.7 & 13.1 & \textcolor{gainGreen}{\underline{\textbf{+2.4}}} \\

No API / format error & 728 & 35.81 & 10.7 & 12.1 & \textcolor{gainGreen}{\underline{\textbf{+1.4}}} \\

Exact tool-step count match & 9 & 0.44 & 10.7 & 11.1 & \textcolor{gainGreen}{+0.4} \\

First tool correct & 128 & 6.30 & 10.7 & 8.6 & \textcolor{lossRed}{\underline{\textbf{-2.1}}} \\

Matched gold prefix $\geq 2$ steps & 36 & 1.77 & 10.7 & 2.8 & \textcolor{lossRed}{\underline{\textbf{-7.9}}} \\

Exact tool-sequence match & 5 & 0.25 & 10.7 & 0.0 & \textcolor{lossRed}{\underline{\textbf{-10.7}}} \\

\midrule

% ================= OPEN =================
\rowcolor{headerBlue!15}
\multicolumn{6}{l}{\textbf{Open-source models}} \\

\rowcolor{highlightRow}
All matched arguments schema-clean & 70 & 6.54 & 7.6 & \textbf{14.3} & \textcolor{gainGreen}{\underline{\textbf{+6.7}}} \\

First tool correct & 69 & 6.45 & 7.6 & 10.1 & \textcolor{gainGreen}{\underline{\textbf{+2.6}}} \\

No API / format error & 240 & 22.43 & 7.6 & 7.5 & \textcolor{lossRed}{-0.1} \\

Matched gold prefix $\geq 2$ steps & 13 & 1.21 & 7.6 & 0.0 & \textcolor{lossRed}{\underline{\textbf{-7.6}}} \\

Exact tool-sequence match & 3 & 0.28 & 7.6 & 0.0 & \textcolor{lossRed}{\underline{\textbf{-7.6}}} \\

\midrule

% ================= PROPRIETARY =================
\rowcolor{headerBlue!15}
\multicolumn{6}{l}{\textbf{Proprietary models}} \\

\rowcolor{highlightRow}
Exact tool-step count match & 4 & 0.42 & 14.1 & \textbf{25.0} & \textcolor{gainGreen}{\underline{\textbf{+10.9}}} \\

No API / format error & 488 & 50.67 & 14.1 & 14.3 & \textcolor{gainGreen}{+0.2} \\

All matched arguments schema-clean & 75 & 7.79 & 14.1 & 12.0 & \textcolor{lossRed}{\underline{\textbf{-2.1}}} \\

First tool correct & 59 & 6.13 & 14.1 & 6.8 & \textcolor{lossRed}{\underline{\textbf{-7.3}}} \\

Matched gold prefix $\geq 2$ steps & 23 & 2.39 & 14.1 & 4.3 & \textcolor{lossRed}{\underline{\textbf{-9.8}}} \\

Exact tool-sequence match & 2 & 0.21 & 14.1 & 0.0 & \textcolor{lossRed}{\underline{\textbf{-14.1}}} \\

\bottomrule
\end{tabular}
\end{adjustbox}
\end{table*}

\begin{figure}[t]
    \centering
    \includegraphics[width=0.7\linewidth]{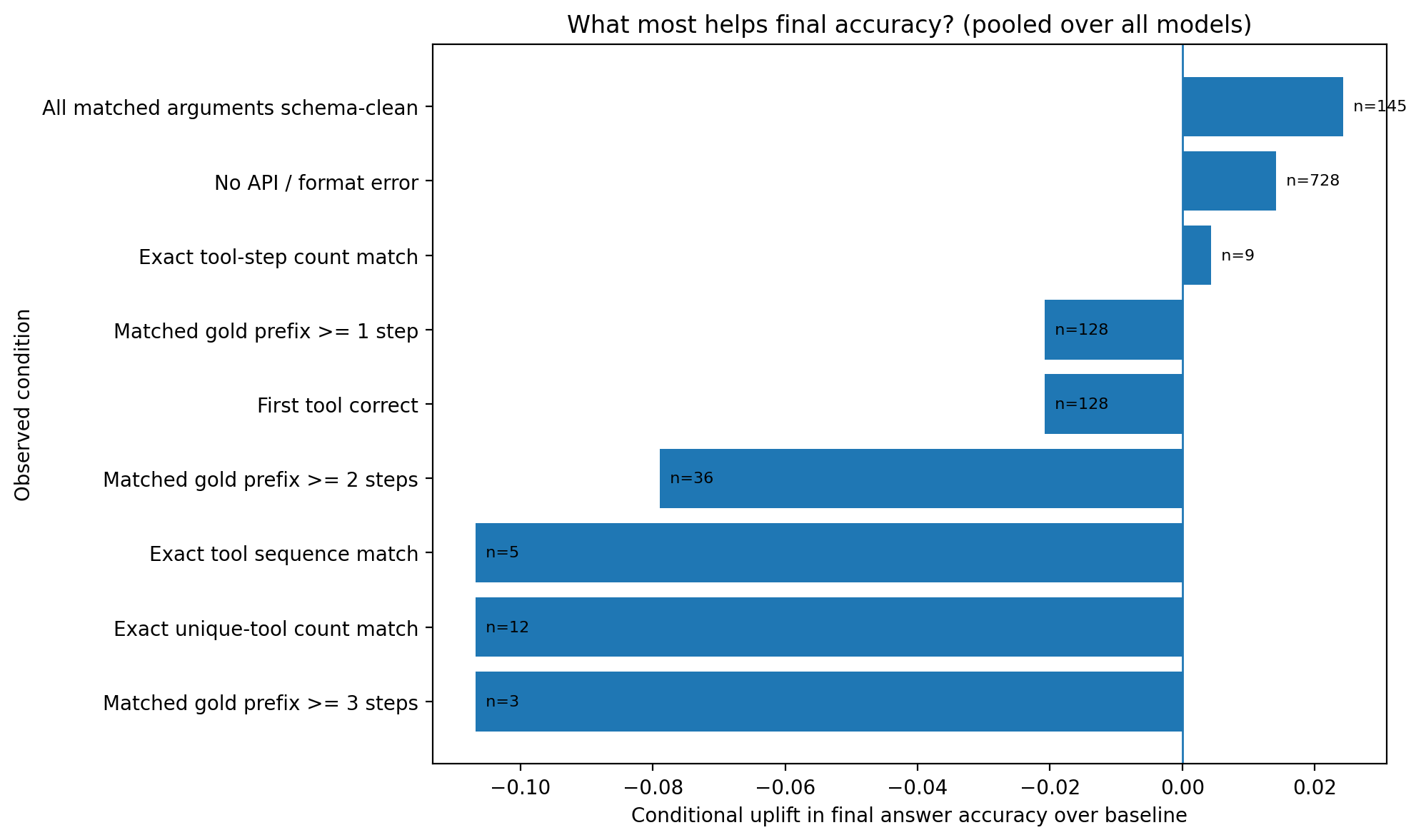}
    \caption{\textbf{Pooled conditional advantage analysis.} The strongest positive descriptive signals come from schema-clean matched arguments and the absence of API/format errors. These are descriptive associations, not causal interventions.}
    \label{fig:component_advantage}
\end{figure}
% =========================================================
% Beautiful Stepwise Survival Table (consistent styling)
% =========================================================

\begin{table}[t]
\centering
\small
\caption{Gold-prefix survival under autonomous rollout (\%).}
\label{tab:stepwise_mode_breakdown}

\rowcolors{2}{stripeA}{stripeB}

\begin{tabular}{lccc}
\rowcolor{headerBlue}
\color{white}{\textbf{Model}} &
\color{white}{\textbf{Step 1}} &
\color{white}{\textbf{Step 2}} &
\color{white}{\textbf{Step 3}} \\
\midrule

\rowcolor{highlightRow}
\textbf{Qwen3.5-9B} & \textbf{33.64} & 9.35 & 0.97 \\

\rowcolor{highlightRow}
\textbf{GPT\_4o} & 32.71 & \textbf{17.76} & \textbf{1.94} \\

Qwen3-8B & 24.30 & 2.80 & 0.00 \\

gpt-5.4-mini & 10.28 & 1.87 & 0.00 \\

gpt-5.4-nano & 7.48 & 0.93 & 0.00 \\

gemini-3.1-flash-lite-preview & 2.80 & 0.00 & 0.00 \\

claude-haiku-4-5 & 0.93 & 0.00 & 0.00 \\

\textcolor{lossRed}{claude-opus-4-6} & 0.00 & 0.00 & 0.00 \\

\bottomrule
\end{tabular}
\end{table}

\begin{figure*}[t]
\centering
\begin{minipage}{0.32\textwidth}
    \centering
    \includegraphics[width=\linewidth]{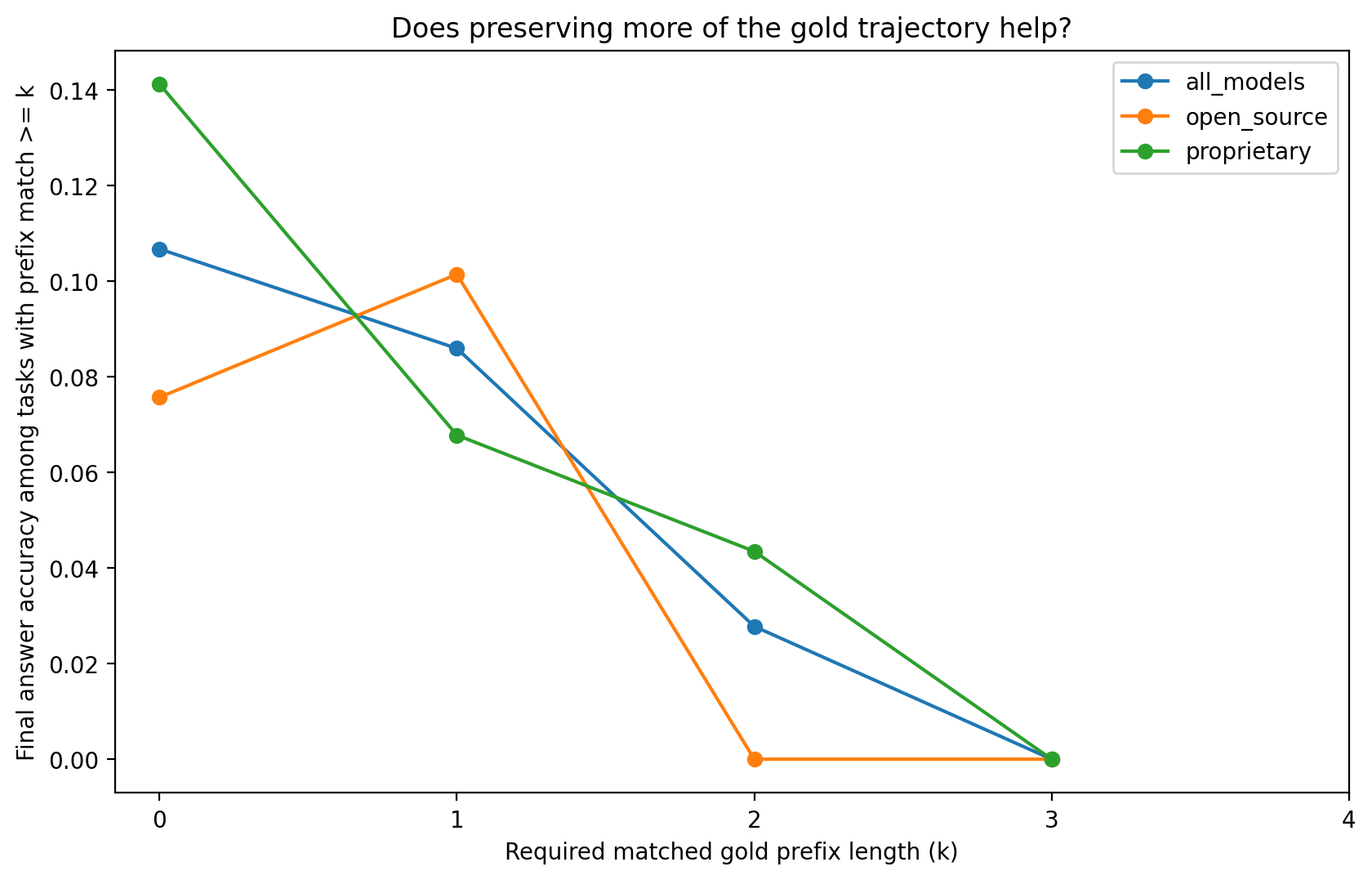}
\end{minipage}
\hfill
\begin{minipage}{0.32\textwidth}
    \centering
    \includegraphics[width=\linewidth]{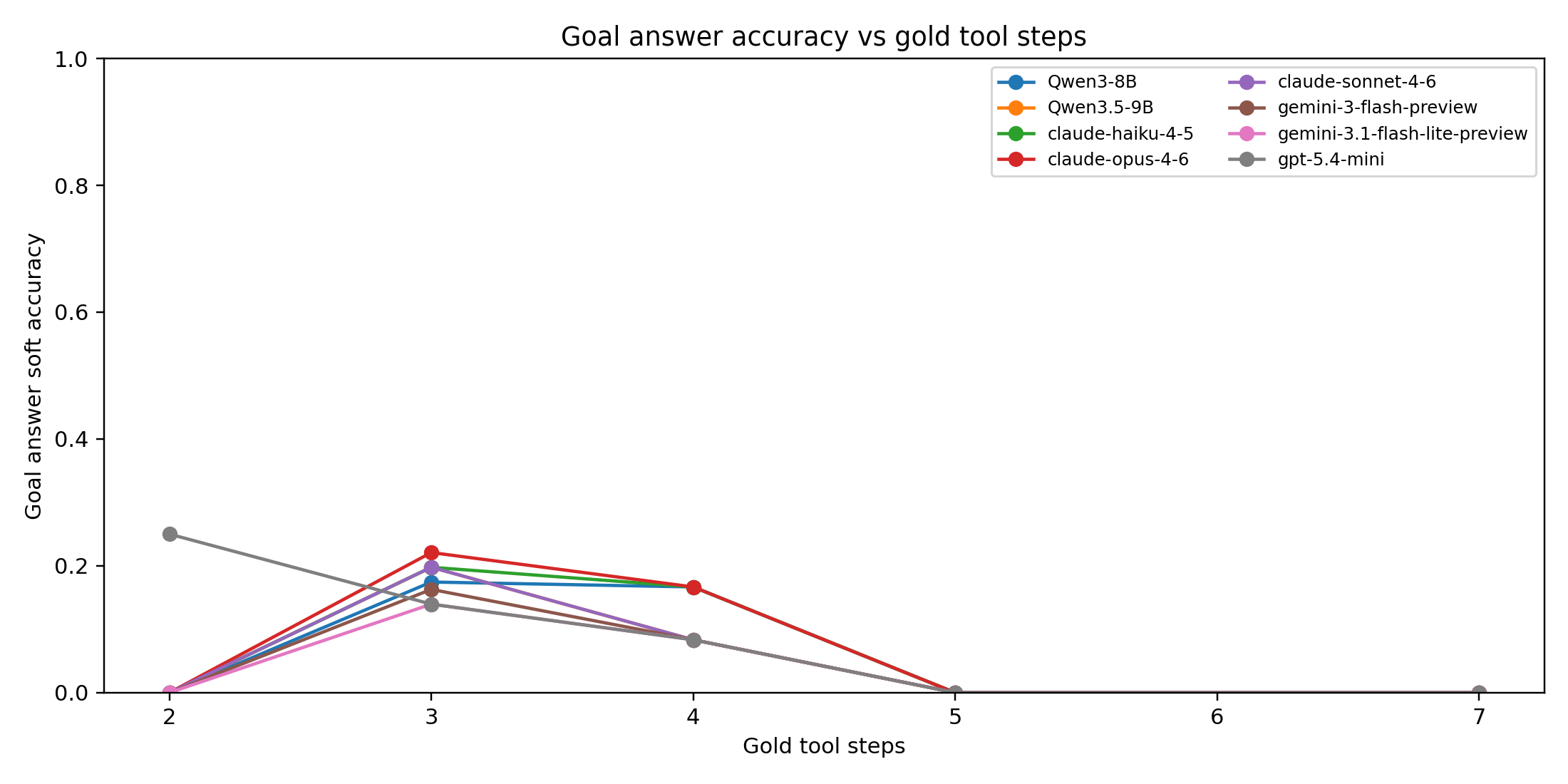}
\end{minipage}
\hfill
\begin{minipage}{0.32\textwidth}
    \centering
    \includegraphics[width=\linewidth]{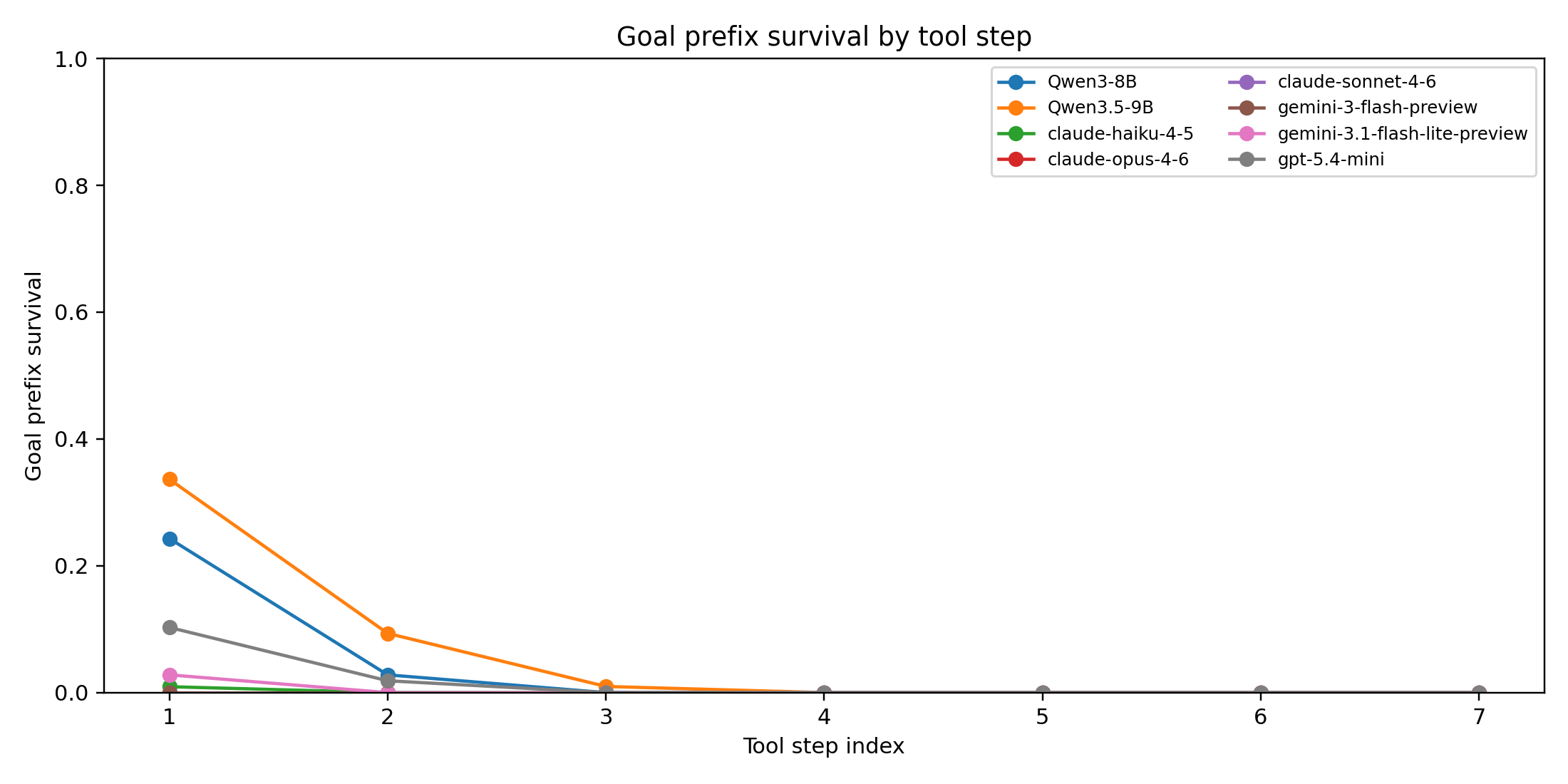}
\end{minipage}
\caption{\textbf{Trajectory-length diagnostics.} \textbf{Left:} prefix-conditioned answer accuracy. \textbf{Middle:} answer accuracy as a function of gold tool-step count. \textbf{Right:} prefix survival by step index. Together these plots show that performance drops with horizon and collapses rapidly after the first action.}
\label{fig:combined_diagnostics}
\end{figure*}
% =========================================================
% Argument pathology: aggregate by tool, not by repeated model rows
% =========================================================
\begin{table}[t]
\centering
\tightarray
\caption{Structural argument pathology aggregated by gold tool, conditioned on \emph{correct-tool calls only}. Schema errors are rare for \texttt{ImageDescription} and \texttt{OCR} and concentrate in \texttt{RegionAttributeDescription}.}
\label{tab:arg_schema_pathology}
\rowcolors{2}{mctaStripe}{white}
\begin{tabular}{lccc}
\rowcolor{mctaHeader}
\theadc{Tool} & \theadc{$n$} & \theadc{Schema OK} & \theadc{Any issue} \\
\midrule
GoogleSearch & 1 & 100.0 & 0.0 \\
ImageDescription & 118 & 100.0 & 0.8 \\
OCR & 32 & 100.0 & 3.1 \\
RegionAttributeDescription & 10 & 90.0 & 40.0 \\
\bottomrule
\end{tabular}
\end{table}

\begin{figure}[htbp]
    \centering
    \includegraphics[width=0.82\linewidth]{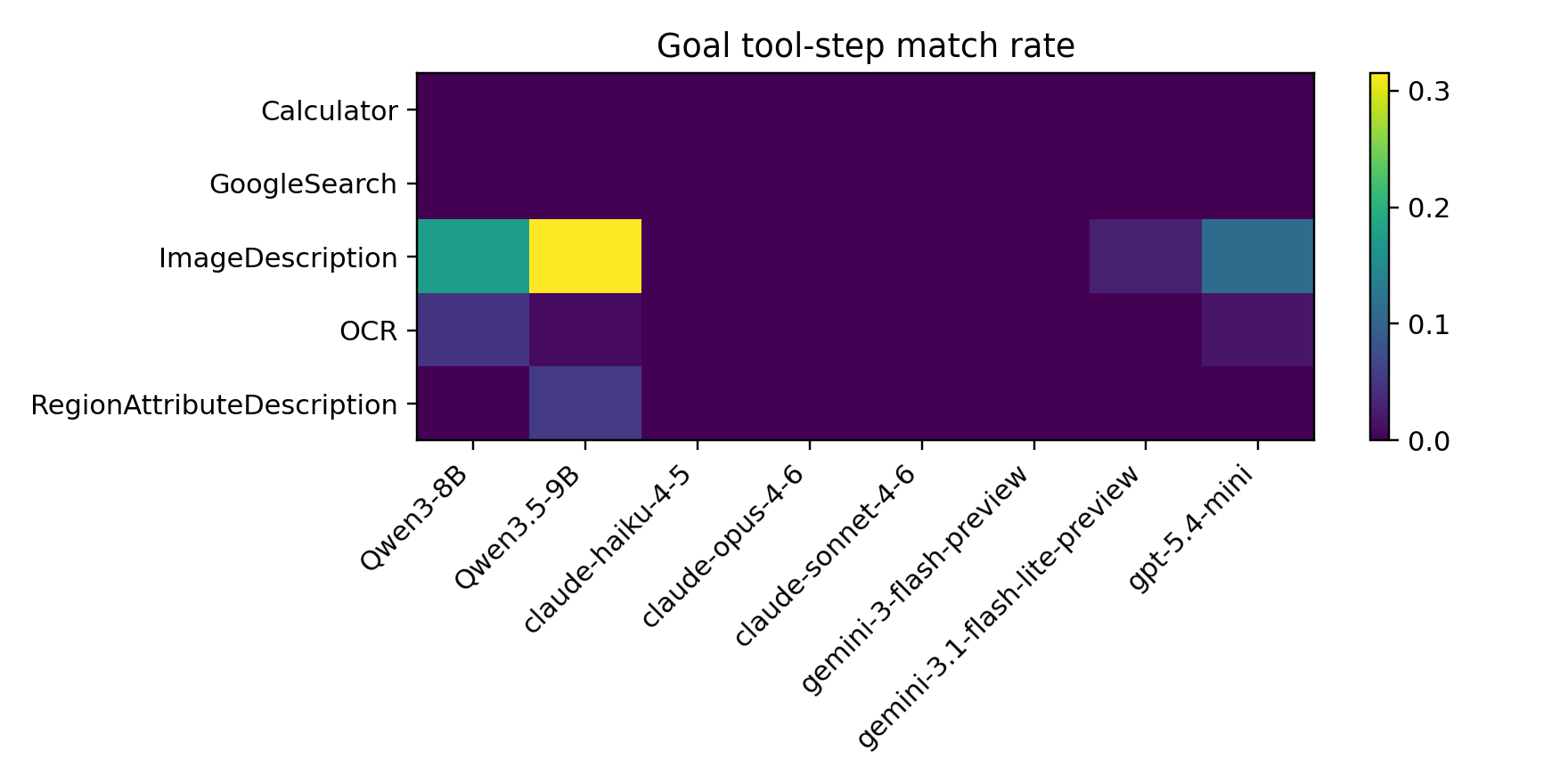}
    \caption{\textbf{Autonomous tool-step match rate by gold tool.} Successful matching is concentrated on \texttt{ImageDescription}, with much weaker alignment on \texttt{OCR} and near-zero matching on more demanding tools such as \texttt{RegionAttributeDescription}.}
    \label{fig:tool_bottleneck}
\end{figure}
Table~\ref{tab:condition_uplift_pooled} and Figure~\ref{fig:component_advantage} summarize which properties of autonomous rollouts tend to co-occur with higher final accuracy. The strongest positive descriptive signal is the subset of rollouts with schema-clean matched arguments, followed by the subset with no API/format error. By contrast, exact-sequence and long-prefix conditions have very small support and do not provide stable evidence of improvement.

These results should be read as \emph{descriptive diagnostics}, not causal claims. They show which rollout properties are associated with success, and they motivate future partial-oracle interventions that provide only limited gold supervision instead of full teacher forcing.

\subsection{Trajectory length, prefix survival, and horizon effects}
\label{app:length}

Table~\ref{tab:stepwise_mode_breakdown} shows that gold-prefix survival collapses rapidly after the first action. Even the strongest autonomous controllers only survive the first step on roughly one-third of tasks, drop to single digits or low teens by step~2, and are essentially at zero by step~3. This explains why strict trajectory success is effectively absent: the rollout rarely stays aligned long enough for full end-to-end faithfulness.

Figure~\ref{fig:combined_diagnostics} gives the complementary picture. The middle panel shows that answer accuracy degrades with gold trajectory length, while the right panel shows the same collapse in prefix survival. The left panel should be interpreted descriptively: tasks with longer matched prefixes are also the hardest long-horizon cases, so prefix-conditioned accuracy is not a causal oracle result.

\subsection{Tool-level bottlenecks and localized grounding}
\label{app:tools}

Figure~\ref{fig:tool_bottleneck} shows that autonomous tool matching is concentrated on relatively easy global-perception tools, especially \texttt{ImageDescription}. The weakest alignment is on \texttt{RegionAttributeDescription}, which requires localized visual grounding and is rarely reached correctly.

Table~\ref{tab:arg_schema_pathology} complements this by aggregating structural argument pathology by tool, conditioned on \emph{correct tool calls}. Once the correct tool is reached, schema problems are very rare for \texttt{ImageDescription} and \texttt{OCR}, but remain much more frequent for \texttt{RegionAttributeDescription}. This supports the main-text conclusion that current limitations are not merely generic formatting failures: after controller repair, the residual difficulty is accurate long-horizon local evidence use.

\subsection{Cross-model qualitative failure cases}
Figure~\ref{fig:qual_examples} complements the aggregate metrics by showing how MedCTA failures unfold at trajectory level. The first case is a \emph{shared} failure: both GPT-5.4 and Qwen3.5-9B miss the correct portal/superior-mesenteric venous thrombosis, but one fails through an immediate argument-format break while the other degrades through repeated API-error loops. The second case shows \emph{semantic drift}: both models ignore the evidence chain needed to map chart labels to assay types and instead answer from broad medical priors. The third case highlights \emph{long-horizon evidence failure}: GPT-5.4 produces a fluent but sign-reversed developmental comparison without using tools, whereas Qwen3.5-9B never recovers after an incorrect initial visual grounding. These examples align closely with the quantitative findings that MedCTA is primarily limited by controller instability, wrong evidence recruitment, and poor obedience to localized or multi-step evidence.
\label{app:qual_examples}

\begin{center}
\begin{tcolorbox}[
    enhanced,
    breakable,
    colframe=black!10,
    colback=gray!1,
    coltitle=black,
    boxrule=1.1pt,
    arc=6pt,
    width=\textwidth,
    title=\textbf{\large Cross-model qualitative examples on MedCTA},
    fonttitle=\bfseries,
    halign title=center
]
\small

% =========================================================
% Case 1
% =========================================================
\begin{tcolorbox}[colback=blue!2, colframe=blue!70!black, title=\textbf{Case 1: Abdominal CT venous thrombosis --- shared failure across both models}]
\textbf{Query:} \emph{Based on the CT image, what type of venous thrombosis is present?}

\vspace{4pt}
\begin{minipage}[t]{0.31\textwidth}
\textbf{Gold trajectory}
\begin{itemize}[leftmargin=1.2em, itemsep=2pt, topsep=2pt]
    \item \texttt{ImageDescription}: identifies an upper-abdominal CT at the porta hepatis.
    \item \texttt{Region..Description}: localizes low-attenuation material in the portal venous system extending into the superior mesenteric vein.
    \item \textbf{Gold answer:} portal vein thrombosis with superior mesenteric vein thrombosis.
\end{itemize}
\end{minipage}\hfill
\begin{minipage}[t]{0.31\textwidth}
\textbf{GPT-5.4}
\begin{itemize}[leftmargin=1.2em, itemsep=2pt, topsep=2pt]
    \item First tool call fails with \texttt{ARGS\_ERROR}: the model appends a free-text interpretation into the JSON arguments.
    \item That inserted text already reframes the image as a \emph{chest} CT with pulmonary arterial filling defects.
    \item \textbf{Predicted answer:} pulmonary embolism.
    \item \textbf{Failure type:} protocol break followed by a wrong anatomic prior.
\end{itemize}
\end{minipage}\hfill
\begin{minipage}[t]{0.31\textwidth}
\textbf{Qwen3.5-9B}
\begin{itemize}[leftmargin=1.2em, itemsep=2pt, topsep=2pt]
    \item Starts with a valid \texttt{ImageDescription}, but only obtains a generic liver/vessel summary.
    \item Then enters repeated \texttt{NoAction}/API-error loops before issuing a coarse full-image region query.
    \item \textbf{Predicted answer:} deep vein thrombosis (DVT).
    \item \textbf{Failure type:} unstable controller followed by a generic thrombosis shortcut.
\end{itemize}
\end{minipage}

\vspace{3pt}
\textbf{Takeaway.} Both models fail on the same case, but in different ways: GPT-5.4 breaks the tool protocol immediately, whereas Qwen3.5-9B stays in the loop longer but never stabilizes enough to gather the decisive local evidence.
\end{tcolorbox}

% =========================================================
% Case 2
% =========================================================
\begin{tcolorbox}[colback=orange!2, colframe=orange!80!black, title=\textbf{Case 2: Sentinel-node metastasis detection methods --- semantic drift from figure evidence}]
\textbf{Query:} \emph{Using the provided figure, identify the methods used to detect metastases in sentinel nodes.}

\vspace{4pt}
\begin{minipage}[t]{0.31\textwidth}
\textbf{Gold trajectory}
\begin{itemize}[leftmargin=1.2em, itemsep=2pt, topsep=2pt]
    \item \texttt{ImageDescription}: recognizes a bar chart comparing H\&E, CK-18, CEA, hTRT, and MUC-1.
    \item \texttt{OCR}: extracts the assay labels.
    \item \texttt{GoogleSearch} $+$ \texttt{Region..Description}: maps the labels to histologic vs.\ molecular detection.
    \item \textbf{Gold answer:} conventional histology, H\&E staining, and RT-PCR/Southern blot assay.
\end{itemize}
\end{minipage}\hfill
\begin{minipage}[t]{0.31\textwidth}
\textbf{GPT-5.4}
\begin{itemize}[leftmargin=1.2em, itemsep=2pt, topsep=2pt]
    \item Skips the evidence chain entirely and answers directly without calling any tool.
    \item Compresses the figure into a generic ``IHC + RT-PCR'' story.
    \item \textbf{Predicted answer:} immunohistochemistry and RT-PCR.
    \item \textbf{Failure type:} premature answer with partial but incomplete method mapping.
\end{itemize}
\end{minipage}\hfill
\begin{minipage}[t]{0.31\textwidth}
\textbf{Qwen3.5-9B}
\begin{itemize}[leftmargin=1.2em, itemsep=2pt, topsep=2pt]
    \item Begins with a vague chart description and then accumulates repeated API errors.
    \item Later OCR recovers the labels (\texttt{H\&E, CK-18, CEA, hTRT, MUC-1}), but the final answer ignores them.
    \item \textbf{Predicted answer:} blue dye injection, radioisotope lymphoscintigraphy, combined method, surgical excision.
    \item \textbf{Failure type:} semantic drift from chart evidence to unrelated clinical prior knowledge.
\end{itemize}
\end{minipage}

\vspace{3pt}
\textbf{Takeaway.} This case shows that failure is not only about syntax. Even when relevant labels are available, the models may abandon the figure-specific evidence path and answer from broad medical priors instead.
\end{tcolorbox}

% =========================================================
% Case 3
% =========================================================
\begin{tcolorbox}[colback=green!2, colframe=green!60!black, title=\textbf{Case 3: Gpc3 embryonic kidney comparison --- evidence-free inversion vs.\ rollout collapse}]
\textbf{Query:} \emph{Compare Gpc3$+$ and Gpc3$-$ embryonic kidneys across E12.0, E13.5, and E16.5, focusing on ureteric bud branching, kidney size, and cortical--medullary architecture.}

\vspace{4pt}
\begin{minipage}[t]{0.31\textwidth}
\textbf{Gold trajectory}
\begin{itemize}[leftmargin=1.2em, itemsep=2pt, topsep=2pt]
    \item \texttt{GoogleSearch}: retrieves known Gpc3 developmental phenotypes.
    \item \texttt{ImageDescription} $+$ \texttt{OCR}: identifies stage labels and genotype layout.
    \item \texttt{Region..Description}: confirms enhanced branching at E12.0, larger mutant kidney at E13.5, and abnormal cortical organization/cysts by E16.5.
    \item \textbf{Gold answer:} mutant kidneys show enhanced branching, enlargement, and later structural disorganization.
\end{itemize}
\end{minipage}\hfill
\begin{minipage}[t]{0.31\textwidth}
\textbf{GPT-5.4}
\begin{itemize}[leftmargin=1.2em, itemsep=2pt, topsep=2pt]
    \item Produces a fluent direct answer without calling any tool.
    \item Reverses the main biological trend: it says the mutant kidneys are \emph{smaller}, \emph{less branched}, and \emph{hypoplastic}.
    \item \textbf{Predicted answer:} reduced branching, smaller kidneys, delayed maturation.
    \item \textbf{Failure type:} evidence-free but linguistically confident inversion of the correct comparison.
\end{itemize}
\end{minipage}\hfill
\begin{minipage}[t]{0.31\textwidth}
\textbf{Qwen3.5-9B}
\begin{itemize}[leftmargin=1.2em, itemsep=2pt, topsep=2pt]
    \item First visual grounding step already fails: the figure is misdescribed as a purple tissue sample or brain/tumor image.
    \item OCR later extracts noisy fragments of the stage labels, but the rollout never recovers and ends in repeated API-error loops.
    \item \textbf{Predicted answer:} no final answer produced.
    \item \textbf{Failure type:} incorrect global grounding followed by controller collapse.
\end{itemize}
\end{minipage}

\vspace{3pt}
\textbf{Takeaway.} The two models expose different failure modes on the same long-horizon comparison task: GPT-5.4 remains fluent but ungrounded, while Qwen3.5-9B loses the task much earlier at the perception/controller interface.
\end{tcolorbox}

\end{tcolorbox}
\captionof{figure}{\textbf{Cross-model qualitative failure cases on MedCTA.} The examples are chosen to span three recurrent benchmark failure modes: \textbf{(i)} early controller/protocol failure with shortcut answering, \textbf{(ii)} semantic drift from figure evidence to broad prior knowledge, and \textbf{(iii)} long-horizon comparison failure caused by either evidence-free reasoning or rollout collapse. Together they make the aggregate diagnostic story concrete: MedCTA errors are not only wrong final answers, but failures of protocol stability, evidence acquisition, and evidence obedience.}
\label{fig:qual_examples}
\end{center}

\subsection{Responsible Use and Ethical Considerations}
\label{sec:ethics}

\ours is an evaluation benchmark for research on clinical tool-agent
reliability. It is not a medical device, not a diagnostic system, and not
intended for patient-facing deployment. Models evaluated on \ours may
produce plausible but incorrect clinical conclusions, may fail to call necessary
tools, and may rely on unsupported priors. We therefore release the benchmark
with research-use terms that prohibit direct clinical deployment, patient triage,
or automated decision making without independent clinical validation and
appropriate regulatory review.

All source assets are public and de-identified to the best of our knowledge.
We include an asset-provenance table and require users to respect upstream
licenses. To reduce privacy and memorization risk, released annotations do not
include identifiable patient information. To reduce misuse risk, benchmark
documentation emphasizes that high leaderboard performance should not be
interpreted as clinical readiness.

\paragraph{Representativeness.}
The current benchmark contains 107 tasks and therefore cannot represent
the full distribution of clinical workflows, patient demographics, institutions,
scanner protocols, or disease prevalence. We report modality and body-system
coverage, but demographic attributes are unavailable for many public assets.
Consequently, \ours should be used to diagnose tool-use and reasoning
failure modes, not to estimate real-world clinical safety or fairness.

\paragraph{LLM-assisted construction and leakage control.}
LLMs are used only to draft candidate query rewrites and candidate trajectories.
Final queries, reference trajectories, and final answers are rewritten and
validated by human annotators and clinician reviewers. The final task statement
is tool-agnostic: it does not expose the reference tool subset, tool order, or
intermediate observations. Reference trajectories are not included in model
prompts during autonomous evaluation. To reduce family-specific construction
bias, we report construction-assistant models separately from the official
leaderboard when applicable, and we perform all evaluation after freezing the
validated benchmark. We also release construction prompts and validation
guidelines so that future benchmark versions can be audited for LLM-induced
artifacts. Because GPT-4o was used during drafting, GPT-4o results are reported as a
construction-family diagnostic rather than as part of the primary leaderboard.

\paragraph{Frozen retrieval.}
To make external retrieval reproducible, \textsc{GoogleSearch} does not query
the live web during benchmark evaluation. During dataset construction we cache
the retrieved snippets, source identifiers, and access dates used by the
reference trajectory. During evaluation, the tool server returns the frozen
top-$k$ snippets associated with the generated query after canonicalization.
This prevents temporal drift in search results and ensures that future runs are
comparable.

\newpage
\clearpage
\section{MedCTA Examples}
\label{mcta_examples}
\begin{figure}[htbp]
    \centering
    \begin{tcolorbox}[
        enhanced,
        colframe=green!80!black,
        colback=green!2,
        coltitle=black,
        boxrule=1.2pt,
        arc=8pt,
        shadow={0.8mm}{-0.8mm}{0mm}{black!20},
        width=\textwidth,
        title=\textbf{\large Example 1},
        fonttitle=\bfseries,
        halign title=center
    ]
    \small

    \begin{center}
    \begin{minipage}[c]{0.42\textwidth}
        \centering
        \includegraphics[width=0.92\textwidth]{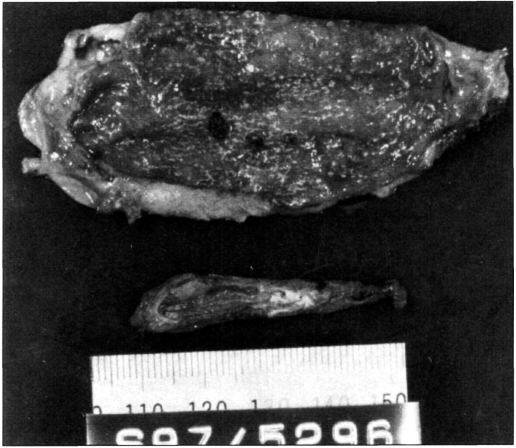}
        \vspace{3pt}\\
        {\small\ttfamily 10444014\_F1.jpg}
    \end{minipage}
    \end{center}

    \vspace{6pt}

    {\color{blue!70!black}\textbf{Query:}}\\
    {\itshape Identify the organ shown and state whether this anatomical variant is common or rare.}

    \vspace{5pt}

    {\color{blue!70!black}\textbf{Involved Tools:}} 
    {\ttfamily ImageDescription, GoogleSearch}

    \vspace{5pt}

    {\color{blue!70!black}\textbf{Steps:}}
    \begin{itemize}[leftmargin=1.2em, itemsep=2pt, topsep=2pt]
        \item Inspect the gross specimen to identify the organ and any abnormal accessory structures.
        \item Use external medical knowledge to determine whether the observed variant is common or rare.
    \end{itemize}

    \vspace{3pt}

    {\color{blue!70!black}\textbf{Reasoning Trace:}}
    \begin{itemize}[leftmargin=1.2em, itemsep=2pt, topsep=2pt]
        \item \textbf{ImageDescription:} The specimen is consistent with a gallbladder and includes an additional smaller sac-like structure, suggesting an accessory or duplicated biliary structure.
        \item \textbf{GoogleSearch:} Accessory or duplicated gallbladder is a rare congenital biliary anomaly, typically identified incidentally during imaging or surgery.
    \end{itemize}

    \vspace{5pt}

    \begin{center}
        \fcolorbox{blue!50!black}{blue!5}{
            \parbox{0.82\textwidth}{
                \centering
                {\color{blue!80!black}\textbf{Answer:}}\\[4pt]
                \textbf{Gallbladder, Rare}
            }
        }
    \end{center}

    \vspace{4pt}

    {\color{blue!70!black}\textbf{Justification:}} 
    The gross specimen shows a gallbladder with an additional accessory sac-like structure, consistent with an accessory or duplicated gallbladder. This is a rare congenital anatomical variant rather than a common finding.

    \end{tcolorbox}
    \caption{\textbf{Agentic reasoning example for anatomical variant identification}}
    \label{fig:agentic-task-3}
\end{figure}

\begin{figure}[htbp]
    \centering
    \begin{tcolorbox}[
        enhanced,
        colframe=green!80!black,
        colback=green!2,
        coltitle=black,
        boxrule=1.2pt,
        arc=8pt,
        shadow={0.8mm}{-0.8mm}{0mm}{black!20},
        width=\textwidth,
        title=\textbf{\large Example 2},
        fonttitle=\bfseries,
        halign title=center
    ]

    \begin{center}
    \begin{minipage}[c]{0.42\textwidth}
        \centering
        \includegraphics[width=0.92\textwidth]{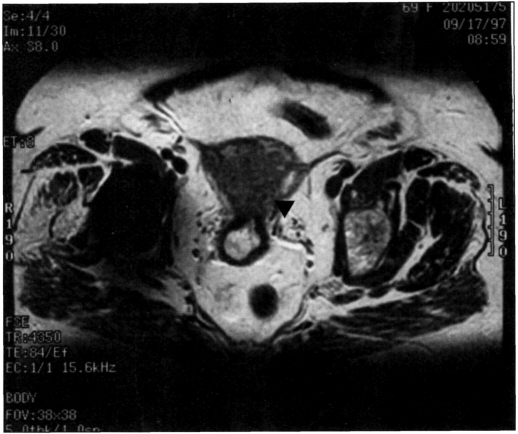}
        \vspace{3pt}\\
        {\small\ttfamily 10444016\_F2.jpg}
    \end{minipage}
    \end{center}

    \vspace{6pt}

    {\color{blue!70!black}\textbf{Query:}}\\
    {\itshape Determine the imaging plane or view of the image.}

    \vspace{5pt}

    {\color{blue!70!black}\textbf{Involved Tools:}} \\
    {\ttfamily OCR, ImageDescription, GoogleSearch, RegionAttributeDescription}

    \vspace{5pt}

    {\color{blue!70!black}\textbf{Steps:}}
    \begin{itemize}[leftmargin=1.2em, itemsep=2pt, topsep=2pt]
        \item Use OCR to extract orientation markers and scan annotations.
        \item Use global image description to infer the anatomical layout.
        \item Use external radiology knowledge to distinguish standard MRI planes.
        \item Verify local attributes such as airway appearance and left--right symmetry.
    \end{itemize}

    \vspace{3pt}

    {\color{blue!70!black}\textbf{Reasoning Trace:}}
    \begin{itemize}[leftmargin=1.2em, itemsep=2pt, topsep=2pt]
        \item \textbf{OCR:} The image contains markers such as \texttt{R} and \texttt{L}, along with MRI metadata.
        \item \textbf{ImageDescription:} The slice shows symmetric left and right soft tissues with a central airway-like structure.
        \item \textbf{GoogleSearch:} Standard radiology references note that bilateral symmetry and cross-sectional anatomy are consistent with common MRI plane cues.
        \item \textbf{RegionAttributeDescription:} The image contains a central round airway-like lumen and symmetric lateral neck soft tissues.
    \end{itemize}

    \vspace{5pt}

    \begin{center}
        \fcolorbox{blue!50!black}{blue!5}{
            \parbox{0.82\textwidth}{
                \centering
                {\color{blue!80!black}\textbf{Answer:}}\\[4pt]
                \textbf{Coronal MRI view}
            }
        }
    \end{center}

    \vspace{4pt}

    {\color{blue!70!black}\textbf{Justification:}} 
    The agent combines OCR-based orientation cues, global anatomical layout, and regional attributes. The observed bilateral symmetry of the cervical soft tissues, together with the central airway-like structure and supporting radiology knowledge, leads to the prediction that the image is a \textbf{coronal MRI view}.

    \end{tcolorbox}
    \caption{\textbf{Agentic reasoning example for imaging plane identification}}
    \label{fig:agentic-task-4}
\end{figure}

\begin{figure}[htbp]
    \centering
    \begin{tcolorbox}[
        enhanced,
        colframe=green!80!black,
        colback=green!2,
        coltitle=black,
        boxrule=1.2pt,
        arc=8pt,
        shadow={0.8mm}{-0.8mm}{0mm}{black!20},
        width=\textwidth,
        title=\textbf{\large Example 3},
        fonttitle=\bfseries,
        halign title=center
    ]
    \small

    \begin{center}
    \begin{minipage}[c]{0.42\textwidth}
        \centering
        \includegraphics[width=0.92\textwidth]{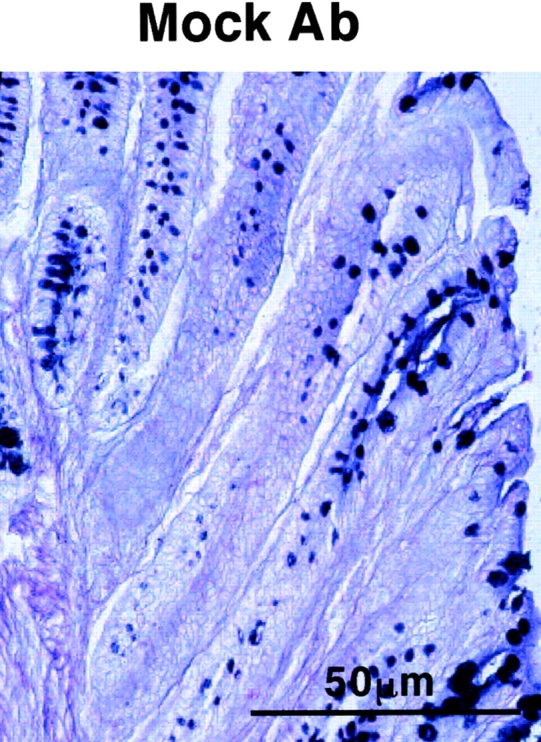}
        \vspace{3pt}\\
        {\small\ttfamily 10477546\_F8.jpg}
    \end{minipage}
    \end{center}

    \vspace{6pt}

    {\color{blue!70!black}\textbf{Query:}}\\
    {\itshape Specify the epithelial cell type of interest, and staining technique used.}

    \vspace{5pt}

    {\color{blue!70!black}\textbf{Involved Tools:}} 
    {\ttfamily OCR, ImageDescription, GoogleSearch}

    \vspace{5pt}

    {\color{blue!70!black}\textbf{Steps:}}
    \begin{itemize}[leftmargin=1.2em, itemsep=2pt, topsep=2pt]
        \item Extract any textual cues (e.g., labels, scale bars) using OCR.
        \item Analyze tissue morphology and staining pattern from the image.
        \item Use domain knowledge to identify cell type and staining technique.
    \end{itemize}

    \vspace{3pt}

    {\color{blue!70!black}\textbf{Reasoning Trace:}}
    \begin{itemize}[leftmargin=1.2em, itemsep=2pt, topsep=2pt]
        \item \textbf{OCR:} Extracted scale bar (\texttt{50 µm}), confirming histological context.
        \item \textbf{ImageDescription:} The image shows intestinal crypt--villus architecture with columnar epithelium and numerous mucin-filled cells.
        \item \textbf{GoogleSearch:} Goblet cells are mucin-secreting epithelial cells; PAS--Alcian blue staining highlights mucin in blue.
    \end{itemize}

    \vspace{5pt}

    \begin{center}
        \fcolorbox{blue!50!black}{blue!5}{
            \parbox{0.82\textwidth}{
                \centering
                {\color{blue!80!black}\textbf{Answer:}}\\[4pt]
                \textbf{Goblet cells, PAS--Alcian blue stain}
            }
        }
    \end{center}

    \vspace{4pt}

    {\color{blue!70!black}\textbf{Justification:}} 
    The intestinal epithelium shows characteristic mucin-filled goblet cells. The strong blue staining of mucin, along with dark nuclear counterstaining and absence of eosinophilic cytoplasm, is consistent with PAS--Alcian blue staining commonly used to highlight mucin-producing cells.

    \end{tcolorbox}
    \caption{\textbf{Agentic reasoning example for epithelial cell and stain identification}}
    \label{fig:agentic-task-5}
\end{figure}

\begin{figure}[htbp]
    \centering
    \begin{tcolorbox}[
        enhanced,
        colframe=green!80!black,
        colback=green!2,
        coltitle=black,
        boxrule=1.2pt,
        arc=8pt,
        shadow={0.8mm}{-0.8mm}{0mm}{black!20},
        width=\textwidth,
        title=\textbf{\large Example 4},
        fonttitle=\bfseries,
        halign title=center
    ]
    \small

    \begin{center}
    \begin{minipage}[c]{0.42\textwidth}
        \centering
        \includegraphics[width=0.92\textwidth]{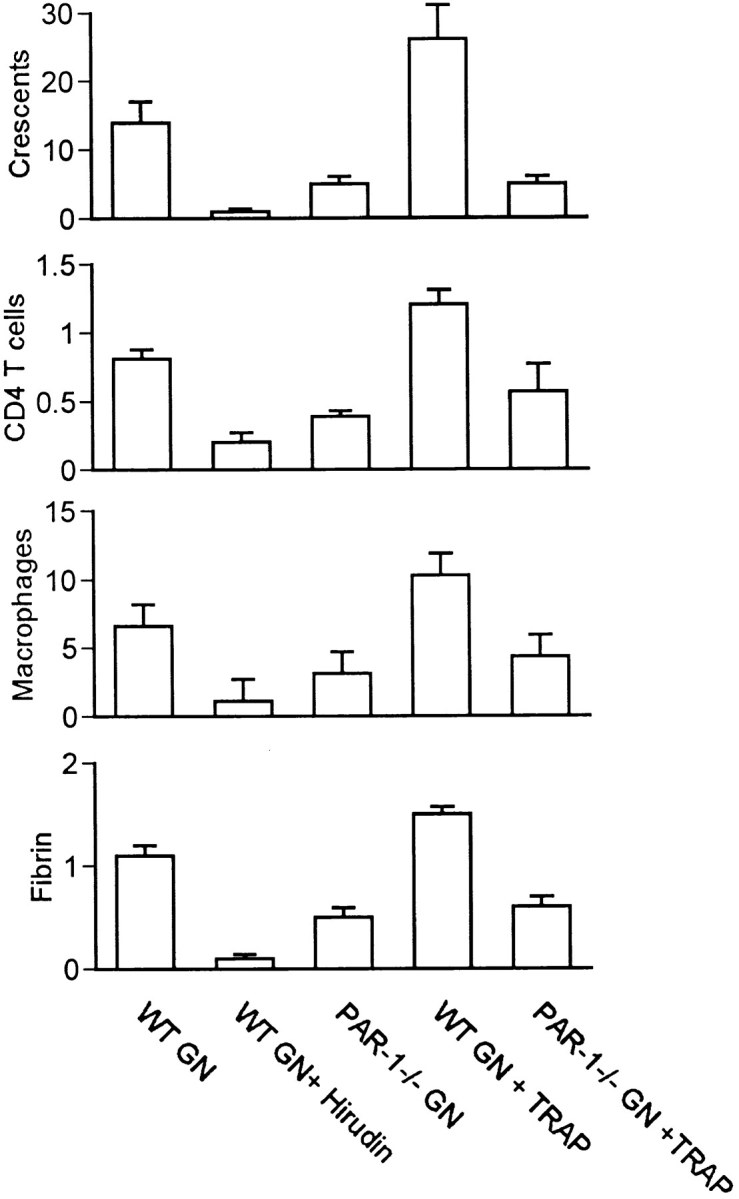}
        \vspace{3pt}\\
        {\small\ttfamily 10662791\_F2.jpg}
    \end{minipage}
    \end{center}

    \vspace{6pt}

    {\color{blue!70!black}\textbf{Query:}}\\
    {\itshape Extract the numerical values of the bars in the top pane and compute the difference between the highest and lowest values.}

    \vspace{5pt}

    {\color{blue!70!black}\textbf{Involved Tools:}} 
    {\ttfamily OCR, ImageDescription, Calculator}

    \vspace{5pt}

    {\color{blue!70!black}\textbf{Steps:}}
    \begin{itemize}[leftmargin=1.2em, itemsep=2pt, topsep=2pt]
        \item Identify the structure of the figure and locate the top panel.
        \item Extract axis values and scale using OCR.
        \item Estimate bar heights relative to the axis.
        \item Compute the difference between maximum and minimum values.
    \end{itemize}

    \vspace{3pt}

    {\color{blue!70!black}\textbf{Reasoning Trace:}}
    \begin{itemize}[leftmargin=1.2em, itemsep=2pt, topsep=2pt]
        \item \textbf{ImageDescription:} The figure contains multiple panels with bar charts; the top panel ranges approximately from 0 to 30.
        \item \textbf{OCR:} Extracted axis ticks \texttt{0, 10, 20, 30}, enabling estimation of bar values.
        \item \textbf{Estimation:} The highest bar is approximately 26, and the lowest is approximately 1.
        \item \textbf{Calculator:} Computes the difference as $26 - 1 = 25$.
    \end{itemize}

    \vspace{5pt}

    \begin{center}
        \fcolorbox{blue!50!black}{blue!5}{
            \parbox{0.82\textwidth}{
                \centering
                {\color{blue!80!black}\textbf{Answer:}}\\[4pt]
                \textbf{25}
            }
        }
    \end{center}

    \vspace{4pt}

    {\color{blue!70!black}\textbf{Justification:}} 
    Using the extracted axis scale and visual estimation of bar heights, the highest value is approximately 26 and the lowest is approximately 1. Their difference is therefore 25.

    \end{tcolorbox}
    \caption{\textbf{Agentic reasoning example for numerical extraction and computation}}
    \label{fig:agentic-task-6}
\end{figure}

\begin{figure}[htbp]
    \centering
    \begin{tcolorbox}[
        enhanced,
        colframe=green!80!black,
        colback=green!2,
        coltitle=black,
        boxrule=1.2pt,
        arc=8pt,
        shadow={0.8mm}{-0.8mm}{0mm}{black!20},
        width=\textwidth,
        title=\textbf{\large Example 5},
        fonttitle=\bfseries,
        halign title=center
    ]
    \small

    \begin{center}
    \begin{minipage}[c]{0.42\textwidth}
        \centering
        \includegraphics[width=0.92\textwidth]{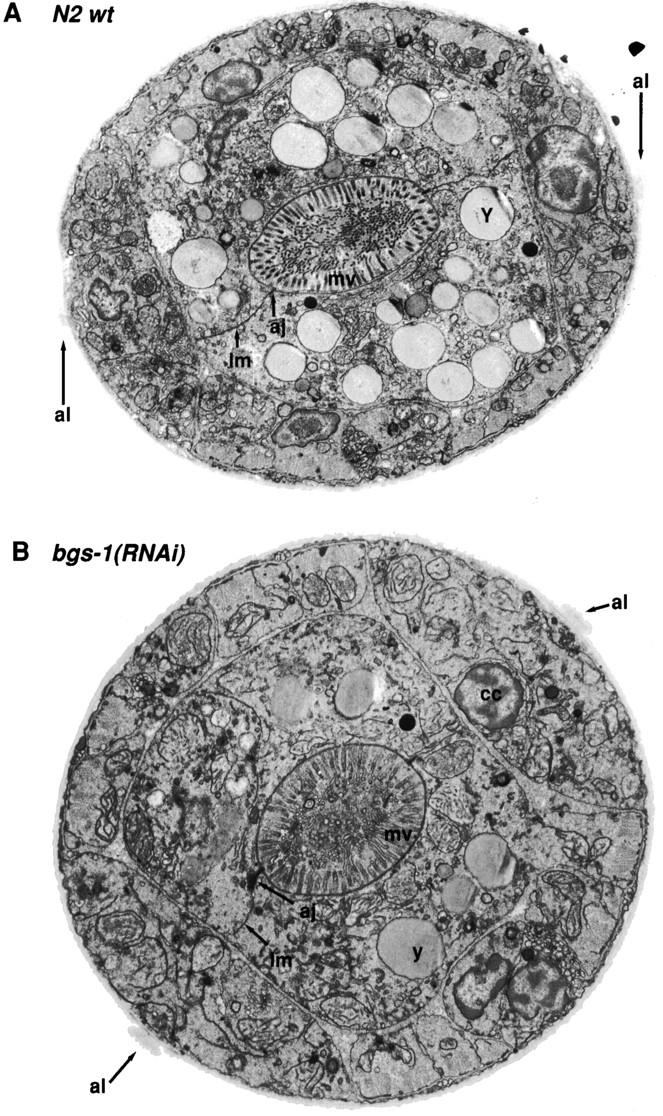}
        \vspace{3pt}\\
        {\small\ttfamily 10811831\_F8.jpg}
    \end{minipage}
    \end{center}

    \vspace{6pt}

    {\color{blue!70!black}\textbf{Query:}}\\
    {\itshape What is the cell type in the image?}

    \vspace{5pt}

    {\color{blue!70!black}\textbf{Involved Tools:}} 
    {\ttfamily ImageDescription, OCR, RegionAttributeDescription}

    \vspace{5pt}

    {\color{blue!70!black}\textbf{Steps:}}
    \begin{itemize}[leftmargin=1.2em, itemsep=2pt, topsep=2pt]
        \item Analyze the global ultrastructure using image description.
        \item Extract and verify structural labels using OCR.
        \item Examine regional attributes to identify specific cell types.
    \end{itemize}

    \vspace{3pt}

    {\color{blue!70!black}\textbf{Reasoning Trace:}}
    \begin{itemize}[leftmargin=1.2em, itemsep=2pt, topsep=2pt]
        \item \textbf{ImageDescription:} Electron micrograph shows intestinal lumen lined by epithelial cells, with labeled features including microvilli (mv), adherens junctions (aj), and seam cells.
        \item \textbf{OCR:} Extracted labels \texttt{mv, aj, lm, y, cc}, confirming annotated ultrastructural features.
        \item \textbf{RegionAttributeDescription:} Peripheral cells correspond to seam cells, distinct from intestinal epithelial cells and coelomocytes.
    \end{itemize}

    \vspace{5pt}

    \begin{center}
        \fcolorbox{blue!50!black}{blue!5}{
            \parbox{0.82\textwidth}{
                \centering
                {\color{blue!80!black}\textbf{Answer:}}\\[4pt]
                \textbf{Seam cells}
            }
        }
    \end{center}

    \vspace{4pt}

    {\color{blue!70!black}\textbf{Justification:}} 
    The peripheral cells identified in the electron micrograph correspond to seam cells, distinct from the central intestinal epithelial cells lining the lumen and the coelomocyte in the pseudocoelomic space.

    \end{tcolorbox}
    \caption{\textbf{Agentic reasoning example for ultrastructural cell type identification}}
    \label{fig:agentic-task-7}
\end{figure}

\begin{figure}[htbp]
    \centering
    \begin{tcolorbox}[
        enhanced,
        colframe=green!80!black,
        colback=green!2,
        coltitle=black,
        boxrule=1.2pt,
        arc=8pt,
        shadow={0.8mm}{-0.8mm}{0mm}{black!20},
        width=\textwidth,
        title=\textbf{\large Example 6},
        fonttitle=\bfseries,
        halign title=center
    ]
    \small

    \begin{center}
    \begin{minipage}[c]{0.42\textwidth}
        \centering
        \includegraphics[width=0.92\textwidth]{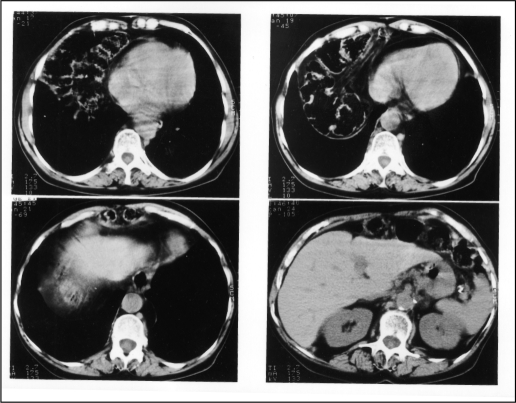}
        \vspace{3pt}\\
        {\small\ttfamily 10917128\_F1.jpg}
    \end{minipage}
    \end{center}

    \vspace{6pt}

    {\color{blue!70!black}\textbf{Query:}}\\
    {\itshape You are given axial CT slices of the lower thorax/upper abdomen. What does the CT scan show?}

    \vspace{5pt}

    {\color{blue!70!black}\textbf{Involved Tools:}} 
    {\ttfamily OCR, ImageDescription, RegionAttributeDescription}

    \vspace{5pt}

    {\color{blue!70!black}\textbf{Steps:}}
    \begin{itemize}[leftmargin=1.2em, itemsep=2pt, topsep=2pt]
        \item Extract any textual or positional cues from the image.
        \item Identify anatomical structures and abnormalities across slices.
        \item Verify regional findings to confirm suspected pathology.
    \end{itemize}

    \vspace{3pt}

    {\color{blue!70!black}\textbf{Reasoning Trace:}}
    \begin{itemize}[leftmargin=1.2em, itemsep=2pt, topsep=2pt]
        \item \textbf{ImageDescription:} Axial CT slices show thoracic and upper abdominal anatomy with abnormal gas-filled structures in the right anterior mediastinum.
        \item \textbf{OCR:} No clinically informative labels detected.
        \item \textbf{RegionAttributeDescription:} Presence of gas-containing bowel loops within the mediastinum confirms abnormal herniation.
    \end{itemize}

    \vspace{5pt}

    \begin{center}
        \fcolorbox{blue!50!black}{blue!5}{
            \parbox{0.82\textwidth}{
                \centering
                {\color{blue!80!black}\textbf{Answer:}}\\[4pt]
                \textbf{Bowel herniation}
            }
        }
    \end{center}

    \vspace{4pt}

    {\color{blue!70!black}\textbf{Justification:}} 
    The CT images show gas-filled bowel loops abnormally located within the right anterior mediastinum, consistent with bowel herniation into the thoracic cavity.

    \end{tcolorbox}
    \caption{\textbf{Agentic reasoning example for thoracic CT abnormality detection}}
    \label{fig:agentic-task-8}
\end{figure}

\begin{figure}[htbp]
    \centering
    \begin{tcolorbox}[
        enhanced,
        colframe=green!80!black,
        colback=green!2,
        coltitle=black,
        boxrule=1.2pt,
        arc=8pt,
        shadow={0.8mm}{-0.8mm}{0mm}{black!20},
        width=\textwidth,
        title=\textbf{\large Example 7},
        fonttitle=\bfseries,
        halign title=center
    ]
    \small

    \begin{center}
    \begin{minipage}[c]{0.42\textwidth}
        \centering
        \includegraphics[width=0.92\textwidth]{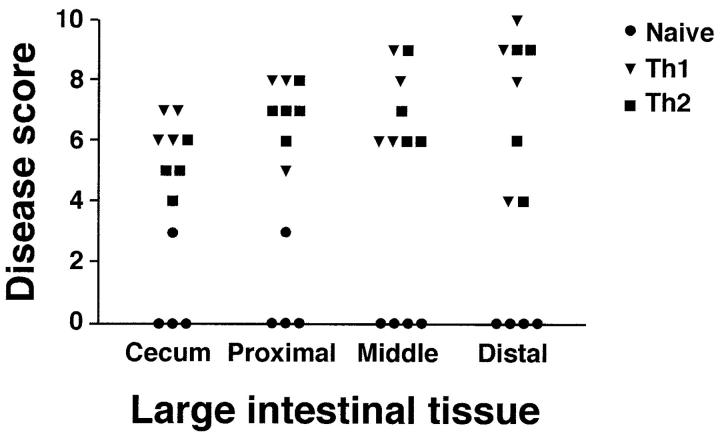}
        \vspace{3pt}\\
        {\small\ttfamily 11781367\_fig5.jpg}
    \end{minipage}
    \end{center}

    \vspace{6pt}

    {\color{blue!70!black}\textbf{Query:}}\\
    {\itshape How many cells are discussed in the figure?}

    \vspace{5pt}

    {\color{blue!70!black}\textbf{Involved Tools:}} 
    {\ttfamily ImageDescription, OCR, RegionAttributeDescription}

    \vspace{5pt}

    {\color{blue!70!black}\textbf{Steps:}}
    \begin{itemize}[leftmargin=1.2em, itemsep=2pt, topsep=2pt]
        \item Identify the overall figure type and legend entries.
        \item Extract exact labels and group names using OCR.
        \item Verify which cell groups are being compared in the plot.
    \end{itemize}

    \vspace{3pt}

    {\color{blue!70!black}\textbf{Reasoning Trace:}}
    \begin{itemize}[leftmargin=1.2em, itemsep=2pt, topsep=2pt]
        \item \textbf{ImageDescription:} The figure is a strip/scatter plot with legend entries \texttt{Naive}, \texttt{Th1}, and \texttt{Th2}.
        \item \textbf{OCR:} Confirms the large intestinal regions and the three groups shown in the legend.
        \item \textbf{RegionAttributeDescription:} The biologically relevant compared cell groups are \textbf{Th1} and \textbf{Th2}, while \texttt{Naive} is the control condition.
    \end{itemize}

    \vspace{5pt}

    \begin{center}
        \fcolorbox{blue!50!black}{blue!5}{
            \parbox{0.82\textwidth}{
                \centering
                {\color{blue!80!black}\textbf{Answer:}}\\[4pt]
                \textbf{Two cells: Th1 and Th2}
            }
        }
    \end{center}

    \vspace{4pt}

    {\color{blue!70!black}\textbf{Justification:}} 
    The figure discusses disease patterns associated with two cell types, \textbf{Th1} and \textbf{Th2}. The \texttt{Naive} group serves as a control rather than a cell type of interest.

    \end{tcolorbox}
    \caption{\textbf{Agentic reasoning example for identifying cell groups in a disease plot}}
    \label{fig:agentic-task-9}
\end{figure}
\clearpage
\section{LLM-as-Judge Prompts}
\vspace{-8 cm}
\label{app:judge_prompts}

\begin{center}
\begin{tcolorbox}[
    enhanced,
    breakable,
    colframe=gray!70!black,
    colback=gray!2,
    coltitle=black,
    boxrule=1pt,
    arc=5pt,
    width=\textwidth,
    title=\textbf{\large LLM-as-Judge Prompts}
]
\small

% =========================
% F_acc
% =========================
\begin{tcolorbox}[colback=gray!2, colframe=gray!70!black, title=\textbf{(a) Clinical Faithfulness ($F_{acc}$)}]
\begin{verbatim}
You are a medical trajectory evaluator.

Evaluate only Clinical Faithfulness (F_acc).

Definition:
Clinical Faithfulness measures whether the predicted reasoning process
follows a clinically valid step-by-step workflow.

Focus ONLY on:
- sequence of reasoning
- whether later conclusions are supported by earlier steps
- whether the workflow is clinically sensible
- whether there are logical jumps or contradictions in the reasoning path

Ignore:
- minor factual wording issues
- missing details unless they break the reasoning chain
- whether all findings are fully covered
- the final answer itself

Important:
- This metric is about LOGICAL WORKFLOW, not completeness.
- A trajectory can be factually incomplete but still faithful.
- A trajectory can have a good-looking final answer but low faithfulness 
if the reasoning path is poor.
- Evaluate only from the trajectory content provided.

Scoring guide:
- 1.0 = clinically coherent stepwise reasoning, no major logic flaws
- 0.7 = mostly reasonable workflow with some weak jumps
- 0.4 = partially logical but important workflow issues
- 0.1 = mostly illogical or unsupported reasoning
- 0.0 = contradictory or clinically nonsensical reasoning

Return JSON only:
{
  "score": number
}
\end{verbatim}
\end{tcolorbox}

\vspace{4pt}

% =========================
% C_s
% =========================
\begin{tcolorbox}[colback=gray!2, colframe=gray!70!black, title=\textbf{(b) Context Integration Score ($C_s$)}]
\begin{verbatim}
You are a medical trajectory evaluator.

Evaluate only Context Integration Score (C_s).

Definition:
Context Integration Score measures how well the predicted trajectory 
uses the available multimodal evidence.

Focus ONLY on:
- whether tool outputs are actually used
- whether image findings, OCR outputs, region descriptions, and evidence 
are integrated into the reasoning
- whether the trajectory is grounded in available context rather than 
generic guessing

Ignore:
- whether the reasoning order is ideal
- whether every medical fact is correct
- whether the final answer is complete
- the final answer itself

Important:
- This metric is about EVIDENCE USAGE, not logic or correctness alone.
- A trajectory can be logically organized but still have low context 
integration if it ignores tool evidence.
- A trajectory can partially use evidence and should get partial credit.
- Evaluate only from the trajectory content provided.

Scoring guide:
- 1.0 = directly and effectively integrates relevant context
- 0.7 = uses some important context but not all
- 0.4 = weak or superficial use of context
- 0.1 = almost no meaningful evidence use
- 0.0 = ignores available context entirely or is unrelated

Return JSON only:
{
  "score": number
}
\end{verbatim}
\end{tcolorbox}

\vspace{4pt}

% =========================
% F_p
% =========================
% \begin{tcolorbox}[colback=gray!2, colframe=gray!70!black, title=\textbf{(c) Clinical Factual Precision ($F_p$)}]
% \begin{verbatim}
% You are a medical trajectory evaluator.

% Evaluate only Clinical Factual Precision (F_p).

% Definition:
% Clinical Factual Precision measures whether the medical claims in the
% trajectory are factually correct and non-hallucinated.

% Focus ONLY on:
% - correctness of diagnoses, anatomy, modality, measurements, locations, 
% and medical claims inside the trajectory
% - hallucinated findings
% - fabricated tool interpretations
% - clinically unsafe or false statements

% Ignore:
% - reasoning order
% - whether every required finding is covered
% - whether evidence integration is elegant
% - the final answer itself

% Important:
% - This metric is about MEDICAL FACTUAL CORRECTNESS only.
% - Penalize hallucinations strongly.
% - If the trajectory contains correct facts plus extra wrong facts, reduce
% the score.
% - Evaluate only from the trajectory content provided.

% Scoring guide:
% - 1.0 = medically precise and factually correct
% - 0.7 = mostly correct with minor inaccuracies
% - 0.4 = mix of correct and incorrect facts
% - 0.1 = mostly false or hallucinated
% - 0.0 = dangerously incorrect, fabricated, or unrelated

% Return JSON only:
% {
%   "score": number
% }
% \end{verbatim}
% \end{tcolorbox}

% \vspace{4pt}

% =========================
% S_comp
% =========================
\begin{tcolorbox}[colback=gray!2, colframe=gray!70!black, title=\textbf{(c) Semantic Completeness ($S_{comp}$)}]
\begin{verbatim}
You are a medical trajectory evaluator.

Evaluate only Semantic Completeness (S_comp).

Definition:
Semantic Completeness measures whether the trajectory covers all 
clinically necessary findings required by the task.

Focus ONLY on:
- whether required reasoning components and findings are present in the 
trajectory
- whether important findings are omitted
- whether the trajectory fully covers the task requirements

Ignore:
- reasoning order
- elegance of evidence usage
- small factual imprecision unless it removes a required component
- the final answer itself

Important:
- This metric is about COVERAGE and MISSING INFORMATION.
- A concise trajectory can score 1.0 if it includes all required findings.
- A logically good trajectory can still score low if key clinical findings
are missing.
- Evaluate only from the trajectory content provided.

Scoring guide:
- 1.0 = all required clinical content is covered
- 0.7 = most important content covered, some omissions
- 0.4 = only partial coverage
- 0.1 = very incomplete
- 0.0 = missing essentially all required content or unrelated

Return JSON only:
{
  "score": number
}
\end{verbatim}
\end{tcolorbox}

\end{tcolorbox}
\end{center}

\end{document}